\documentclass[journal]{IEEEtran}

\usepackage{amsmath}
\usepackage{amssymb}
\usepackage{amsthm}
\usepackage{multirow}
\usepackage{mathtools}
\usepackage{subcaption}
\usepackage[pdftex]{graphicx}
\usepackage[numbers,sort]{natbib}
\usepackage{url}
\usepackage{rotating}
\usepackage{pdflscape}
\usepackage{afterpage}
\usepackage{color}
\usepackage{tablefootnote}
\usepackage[multiple]{footmisc}

\newtheorem{theorem}{Theorem}
\newtheorem{lemma}{Lemma}
\newtheorem{definition}{Definition}

\newcommand{\reals}{\mathbb R}

\def\bP{\boldsymbol{P}}
\def\bPi{\boldsymbol{\Pi}}
\def\bI{\boldsymbol{I}}
\def\bY{\boldsymbol{Y}}
\def\bX{\boldsymbol{X}}

\def\bN{\boldsymbol{N}}
\def\bR{\boldsymbol{R}}
\def\bU{\boldsymbol{U}}
\def\bW{\boldsymbol{W}}

\def\bS{\boldsymbol{S}}
\def\bQ{\boldsymbol{Q}}
\def\bL{\boldsymbol{L}}
\def\bS{\boldsymbol{S}}

\def\bC{\boldsymbol{C}}

\def\bV{\boldsymbol{V}}

\def\bx{\boldsymbol{x}}

\def\by{\boldsymbol{y}}

\def\bz{\boldsymbol{z}}

\def\br{\boldsymbol{r}}
\def\be{\boldsymbol{e}}
\def\bb{\boldsymbol{b}}

\def\bv{\boldsymbol{v}}

\def\bn{\boldsymbol{n}}

\def\cS{\mathcal{S}}

\def\cX{\mathcal{X}}

\def\cP{\mathcal{P}}
\def\cA{\mathcal{A}}
\def\cR{\mathcal{R}}

\def\bzero{\boldsymbol{0}}

\def\bmu{\boldsymbol{\mu}}

\def\bSigma{\boldsymbol{\Sigma}}

\def\nnz{\mathtt{nnz}}
\def\rank{\text{rank}}

\def\poly{\text{poly}}

\def\argmin{\text{argmin}}
\def\argmax{\text{argmax}}

\DeclareMathOperator{\Tr}{\text{Tr}}
\DeclareMathOperator{\Sp}{\text{Sp}}
\DeclareMathOperator{\tr}{tr}

\title{An Overview of Robust Subspace Recovery}
\author{Gilad Lerman and Tyler Maunu}
\date{}

\begin{document}
	
\maketitle	
	
\begin{abstract}
    This paper will serve as an introduction to the body of work on robust subspace recovery. Robust subspace recovery involves finding an underlying low-dimensional subspace in a dataset that is possibly corrupted with outliers. While this problem is easy to state, it has been difficult to develop optimal algorithms due to its underlying nonconvexity. This work emphasizes advantages and disadvantages of proposed approaches and unsolved problems in the area.	
\end{abstract}


\section{Introduction: What is Robust Subspace Recovery?}

The purpose of this work is to survey and discuss the existing literature related to the problem of \emph{robust subspace recovery} (RSR). By ``robust'', we mean that the methods we consider should not be too sensitive to corruptions in a dataset. These ideas trace their roots back quite far in the statistical literature~\citep{huber_book,maronna2006robust}. The basic motivation behind the development of robust procedures is that real data often does not subscribe to the clean assumptions required by many classical statistical procedures. Quoting Huber~\cite{huber_book}, ``robustness signifies insensitivity to small deviations from the assumptions''. The body of work considered in this survey tackles the question of robustness in a certain challenging and nonconvex statistical problem.

RSR involves finding a low-dimensional subspace structure in a corrupted, potentially high-dimensional dataset. Since the set of all subspaces of a fixed dimension is nonconvex, the RSR problem itself is inherently nonconvex. This has made the problem challenging to solve and has, in part, led to the variety of works outlined here.

At this point, it is essential that we clearly specify the problem, since there are many works in related but different areas. Indeed, the literature is confusing to navigate because this problem has also been coined robust principal component analysis (RPCA). As a classical statistical method, principal component analysis (PCA) attempts to model data by a subspace that captures the directions of maximum variance, but it is notoriously sensitive to corrupted data. Many researchers have proposed robust estimators, but the estimators mostly fall into two camps: \emph{outlier-robust} methods and \emph{sparse-corruption} methods. We hope to make this distinction clear, so as to avoid confusion between the two competing bodies of literature. The RSR problem is related to the former, while it has become common to use RPCA to refer to the latter.

For this discussion, assume we are given a dataset $\cX = \{\bx_1,\dots,
\bx_N\}$, with corresponding data matrix $\bX = [\bx_1,\dots,\bx_N]$. In the
literature, RPCA or \emph{sparse-corruption} methods have focused on
decomposition of a matrix $\bX$ into low-rank and sparse components, $\bX = \bL
+ \bS$, where $\bL$ is low-rank and $\bS$ is sparse
(elementwise)~\citep{candes2011robust,zhou2010stable}. {
Here, the goal is to recover the full low rank matrix $\bL$ from the corrupted observations. A comprehensive review
of this topic is given in~\citep{vaswani2018static}.}

\begin{figure}\centering
    \includegraphics[width = .45\textwidth]{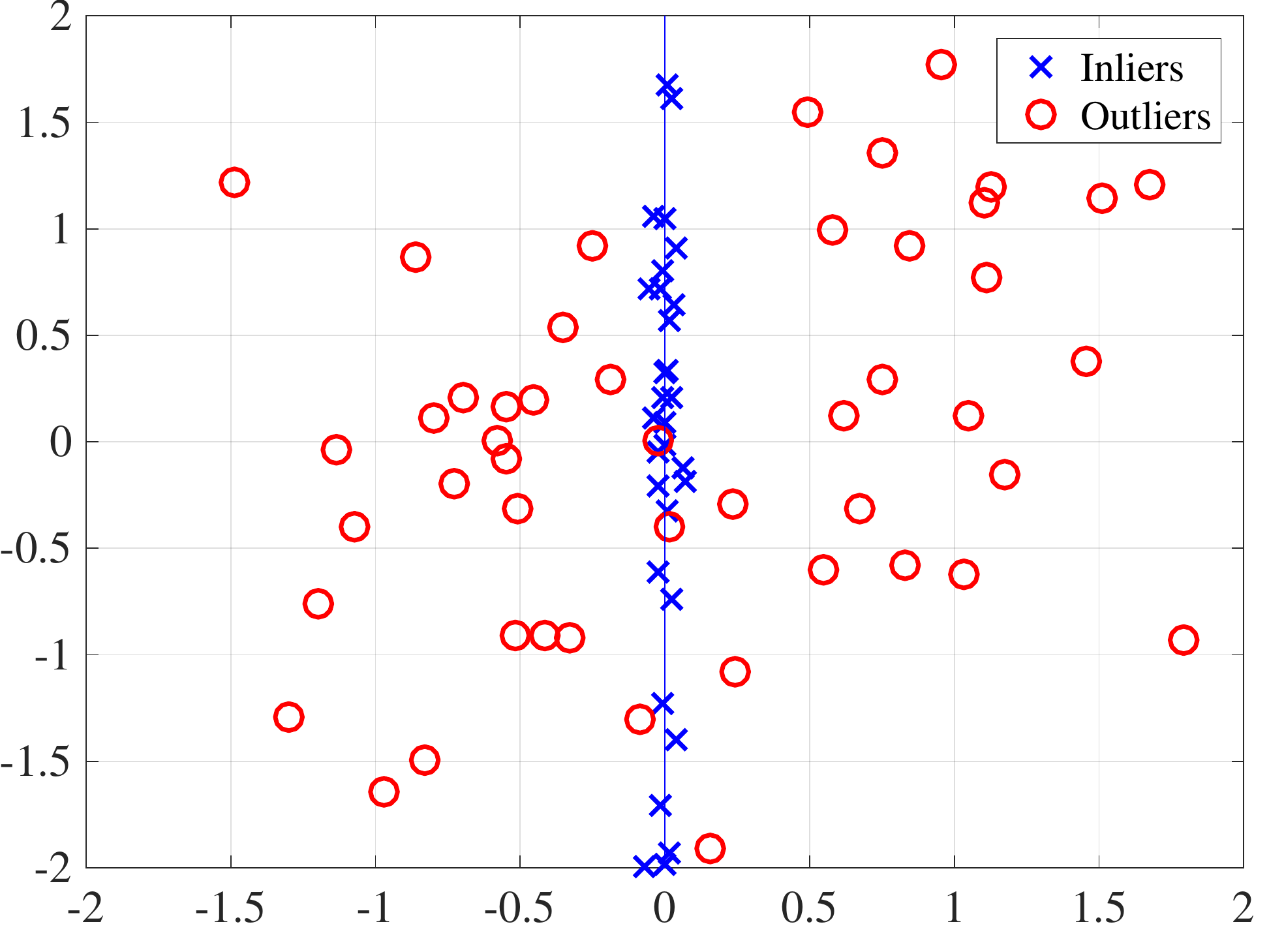}
    \caption{Demonstration of an inlier-outlier dataset in $\reals^2$ with an underlying one-dimensional subspace (the $y$-axis). The inliers are denoted by ``x'' and the outliers are denoted by ``o''.}\label{fig:inout}
\end{figure}

On the other hand, the best way of thinking about RSR datasets is through partitioning $\cX$ into inlier and outlier components, $\cX = \cX_{\mathrm{in}} \cup \cX_{\mathrm{out}}$, where the inliers lie on or near a low-dimensional subspace, and the outliers are somehow distributed in the ambient space. We call such a dataset an inlier-outlier dataset. For clarity and so the reader may visualize the case we are talking about, we have displayed an artificial inlier-outlier dataset in Figure~\ref{fig:inout}.
The RSR problem asks to recover the underlying low-dimensional subspace.
This problem is sometimes written as $\bX = \bL + \bC$, where the columns of $\bL$ span the underlying subspace and the non-zero columns of $\bC$ correspond to outliers.
Similar to the formulations of RPCA, some works have enforced column-sparsity of $\bC$. However, calling $\bC$ column-sparse in general is misleading, since many works on RSR consider very high percentages of outliers, in which case most of the columns of $\bC$ are non-zero.
This notation is also somewhat problematic, since the actual goal of RSR is to recover the underlying low-dimensional subspace, rather than the full low-rank matrix $\bL$. Estimation of the subspace itself gives a more flexible output, while there is some freedom in choosing low-rank matrices $\bL$ and corruption matrices $\bC$ corresponding to a given subspace.

It is also important to note that the second case (column-sparse-corruption) is not just a special case of the first (elementwise-sparse-corruption). First of all, as mentioned above, many works on outlier-robust methods have considered cases with high percentages of outliers and, in some cases, have considered models where algorithms can tolerate arbitrary percentages of outliers. In this case, the corruption matrix can become quite dense. Second of all, the theoretical results for most sparse-corruption based methods have assumed that the corruptions are uniformly distributed across the elements of the data matrix. A matrix with column-sparse corruptions would have positions that are highly correlated and thus none of the current theoretical results for RPCA apply to RSR.

\subsection{Roadmap}

Here we briefly give an overview of the structure of this survey paper. We first give the basic formulations and algorithmic approaches for robust subspace recovery in \S\ref{sec:basics}. Then, in \S\ref{sec:theory}, we discuss and compare the various recovery guarantees for RSR algorithms, and we include a detailed discussion on well-defined data models. We display the computational complexity and memory requirements for the competing RSR algorithms in \S\ref{sec:comp}. Empirical comparisons of the various RSR algorithms are discussed in \S\ref{sec:exp}, where we consider how one should measure the performance of an RSR algorithm, give comprehensive comparisons using various simulated datasets, discuss experiments that have been done on real data, and propose the creation of a substantial database for testing the applicability of RSR algorithms. The influence of RSR methods on other areas is discussed in \S\ref{sec:rsr_influence}. Finally, in \S\ref{sec:conclusion}, we finish with an outline of what remains to be done for RSR algorithms, and where we believe the field should go next.

\subsection{Notation}
\label{sec:notation}

In general, bold capital letters denote matrices and bold lower case letters denote vectors. For two sets $A$ and $B$, $A \setminus B$ denotes the relative complement of $B$ in $A$. The $(D-1)$-dimensional unit sphere in $\reals^{D}$ is denoted by $S^{D-1}$. The Grassmannian $G(D,d)$ is the set of $d$-dimensional linear subspaces in $\reals^D$, which we also refer to as $d$-subspaces. For a subspace $L \in G(D,d)$, its orthogonal complement is denoted by $L^{\perp} \in G(D,D-d)$. The matrix $\bI_d$ denotes the $d \times d$ identity matrix, and, where it is not ambiguous, we just write $\bI$. The set of semi-orthogonal matrices $O(D,d)$ is defined as $O(D,d) = \{ \bU \in \reals^{D \times d} : \bU^T \bU = \bI_d\}$. The norm $\| \cdot \|$ is used to refer to the Euclidean norm, and $\#(\cdot)$ denotes the number of elements in a set. The matrix $\bP_L$ denotes the orthoprojector onto the subspace $L$, while $\bQ_L$ is the orthoprojector onto $L^\perp$: $\bQ_L = \bI - \bP_L$. Throughout the paper, we assume an inliers-oultiers dataset $\cX = \cX_{\mathrm{in}} \cup \cX_{\mathrm{out}}$ with $N$ points and define $N_{\mathrm{in}} = \#(\cX_{\mathrm{in}})$ and $N_{\mathrm{out}} = \#(\cX_{\mathrm{out}})=N-N_{\mathrm{in}}$.
As mentioned earlier, we denote the data points of $\cX$ by $\bx_1,\dots,\bx_N \in \reals^D$ and their corresponding data matrix by $\bX = [\bx_1, \dots, \bx_N] \in  R^{D\times N}$.
The data matrices for $\cX_{\mathrm{in}}$ and $\cX_{\mathrm{out}}$ are $\bX_{\mathrm{in}} \in \reals^{D \times N_{\mathrm{in}}}$ and $\bX_{\mathrm{out}} \in \reals^{D \times N_{\mathrm{out}}}$, respectively.
We use ``w.h.p.'' to denote ``with high probability'', which refers to probabilities that have orders $1-O(N^{-a})$, for some absolute constant $a>0$. Similarly, we use ``w.o.p.'' to denote ``with overwhelming probability'', which refers to probabilities that scale at least like $1-O(e^{-bN^a})$, for an absolute constant $a>0$, and a constant $b > 0$ that is independent of $N$, but may depend on $D$, $d$, and the fraction of outliers.
In many of the nonconvex optimization problems considered here, the minimizer or maximizer may not be unique in general. Thus, we write ``$\in \argmax$'' or ``$\in \argmin$'' to denote that the estimator is contained in the set of maximizers of minimizers, respectively.

\section{Basic Formulations for Robust Subspace Recovery}
\label{sec:basics}

In this section, we hope to motivate a few basic strategies for subspace recovery in order to give a better understanding of the problem.
For the rest of this survey, we assume a linear subspace setting.
That is, the subspace on or around which the inliers lie is linear. 
Here, we have an inlier-outlier data matrix, $\bX \in \reals^{D \times N}$, 
and we wish to recover a linear subspace $L \in G(D,d)$. We may interchangeably search for a matrix $\bU \in O(D,d)$ whose columns span $L \in G(D,d)$. The case of affine subspaces is discussed in \S\ref{sec:conclusion}.
After briefly reviewing PCA in~\S\ref{subsec:pcarev} and discussing the difficulties of developing an outlier-robust version of PCA in~\S\ref{subsec:nontriv}, we discuss the various approaches of RSR algorithms in the following categories
\begin{itemize}
	\item \S\ref{subsubsec:projpur} Projection Pursuit
	\item \S\ref{subsec:lad} Least Absolute Deviations
    \item \S\ref{subsubsec:l1pca}  $L_1$-PCA
	\item \S\ref{subsec:robcov} Robust Covariances
	\item \S\ref{subsec:otheren} Other Energy Minimizers
	\item \S\ref{subsec:filter} Filtering Outliers
	\item \S\ref{subsec:exhaust} Exhaustive Subspace Search
\end{itemize}
At last, in \S\ref{sec:parallel} we discuss some related parallel works to RSR.

\subsection{Review of Subspace Modeling by PCA}
\label{subsec:pcarev}

Classically, subspace modeling has been formulated using principal component analysis (PCA), which finds the orthogonal directions of maximum variance.
Using the notation in \S\ref{sec:notation}, the PCA $d$-subspace of the dataset $\{\bx_i\}_{i=1}^N \subset \reals^D$ is defined as
\begin{equation}
L_{PCA} \in \argmax_{L \in G(D,d)} \sum_{i=1}^N \| \bP_L \bx_i \|^2.
\label{eq:pcamax}
\end{equation}
This subspace has a direct and simple numerical solution. Indeed, it is the span of the top $d$ eigenvectors of the scaled sample covariance, $\bX \bX^T$, or equivalently, the top $d$ left singular vectors of $\bX$.
This solution is unique when the $d$th and $(d+1)$st eigenvalues of $\bX \bX^T$ are not equal. Otherwise, all $d$-subspaces of a larger subspace of $\reals^D$ are the global minimizers, and there are no other local minimizers.
The PCA minimization is very nice compared to many other nonconvex optimization formulations due to this direct solution.

The equivalent formulation for this problem over $O(D,d)$ is
\begin{equation}
\bU_{PCA} \in \argmax_{\bU \in O(D,d)} \sum_{i=1}^N \| \bU \bU^T \bx_i \|^2.
\label{eq:pcamax2}
\end{equation}
Another equivalent formulation of~\eqref{eq:pcamax} immediately follows from the identity $\|\bx_i\|^2 = \| \bP_L \bx_i \|^2 + \| \bQ_L \bx_i \|^2$:
\begin{equation}
L_{PCA} \in \argmin_{L \in G(D,d)} \sum_{i=1}^N \| \bQ_L \bx_i \|^2.
\label{eq:pcamin}
\end{equation}
This formulation can be interpreted as minimizing the variance orthogonal to a subspace. In simple geometric terms, it minimizes the sum of squared orthogonal distances between the data points and the subspace $L$. Indeed, the function $\| \bQ_L \bx_i \|$ in \eqref{eq:pcamin} is just the orthogonal distance between the point, $\bx_i$, and the subspace $L$.
Notice that the choice of the squared Euclidean norm can be motivated by maximum likelihood estimation of the PCA subspace under a Gaussian generative model, analogous to the least squares estimator in ordinary least squares regression.

\subsection{Difficulties of Developing Outlier-Robust PCA}
\label{subsec:nontriv}

Beyond PCA, which has a direct solution, the problem of robustly estimating a subspace becomes hard. Indeed, issues range from the proper definition of a robust estimator to the actual calculation of these estimators.

As an example, consider the following program to robustly find an underlying subspace. In a noiseless inlier-outlier dataset, one may replace the least squares formulation of PCA in~\eqref{eq:pcamin} with the following $\ell_0$-type formulation:
\begin{equation}
\hat{L} \in \argmin_{L \in G(D,d)} \#(\cX \setminus L).
\label{eq:l0}
\end{equation}
In the case of noisy inliers, one may try to find
\begin{equation}
\hat{L} \in \argmin_{L \in G(D,d)} \#\{\bx \in \cX : \|\bQ_L \bx\| >\epsilon \},
\label{eq:l0nmin}
\end{equation}
where $\epsilon>0$ is somehow tied to the magnitude of the noise.
There is no easy way of even approximating the solution to~\eqref{eq:l0} or~\eqref{eq:l0nmin} in general. Further, when real data is noisy, there is no obvious way to choose the parameter $\epsilon$ in~\eqref{eq:l0nmin}. As we will discuss later, relaxing~\eqref{eq:l0} to an $\ell_1$ formulation still results in an NP-hard problem. This stands in contrast to the $\ell_0$ to $\ell_1$ relaxation in settings like regression or compressed sensing, where one gets a convex program that can be solved using a variety of methods.
Also, the solutions of~\eqref{eq:l0} and~\eqref{eq:l0nmin} may not be unique, whereas our initial formulation of the RSR problem assumes a unique underlying subspace. This issue, which is evident in non-convex programs for RSR, will be later addressed in \S\ref{subsec:statmod}.

It is also unclear that the formulations in~\eqref{eq:l0} and~\eqref{eq:l0nmin} are the most natural ones. Indeed, in real situations, data is quite messy and never lies exactly on a subspace, and so one must consider~\eqref{eq:l0nmin} in general. However, there are various scenarios where~\eqref{eq:l0nmin} may not give a useful estimate. For example,~\eqref{eq:l0nmin} may not perform well when the noise is not uniform around the subspace or when the outliers lie around a union of nearby subspaces and $\epsilon$ is overestimated, as we demonstrate later in Figure~\ref{subfig:mssissue}.

\subsection{Projection Pursuit}
\label{subsubsec:projpur}

A body of works on robust subspace recovery includes projection pursuit based methods~\citep{friedman1974projection,huber_book,Li_85,Ammann1993,maronna2005principal,choulakian06,kwak08,mccoy2011two}, which can be motivated in the following way. One can attempt to find a direction (component) maximizing a robust scale function $\rho: \reals^N \to [0,\infty)$ with respect to the data as follows:
\begin{equation}\label{eq:projpur}
    \bv_1 \in \argmax_{\bv \in S^{D-1}} \rho (\bv^T \bX).
\end{equation}
One typically finds all $d$ components in a sequential manner, which we explain after discussing the notion of a robust scale function and attempts to solve \eqref{eq:projpur}.

When using the non-robust scale function $\rho(\by) = \| \by \|_2^2$, $\bv_1$ is the top principal component, which is also expressed by~\eqref{eq:pcamax2} when $d=1$. A robust version of the top principal component can be developed by choosing a proper scale function, such as a trimmed variance, $\rho(\by) = \| \by \|_1$, or a Huber-type scale function. When $d = 1$, using $\rho(\by) = \| \by \|_1$ results in the maximization variants of both least absolute deviations and $L_1$-PCA, which will be presented later in \eqref{eq:lad_new} and \eqref{eq:l1max} respectively. One can attempt to optimize the nonconvex objective \eqref{eq:projpur} in many ways.
In general, exhaustively searching for this maximizer results in a non-polynomial time algorithm. Instead, most algorithms resort to finding a local maximum of~\eqref{eq:projpur} or some sort of approximate global maximum.
Past works have used iterative reweighting schemes~\cite{maronna2005principal}, bit-flipping~\citep{kwak08}, and convex relaxation~\citep{mccoy2011two}.

One can estimate a set of components in a sequential manner in the following way. After finding $\bv_1$ by \eqref{eq:projpur}, each sequential component $\bv_j$, $1<j \leq d$, is found by solving the same problem with the added constraint of orthogonality with the previously found vectors $\bv_1, \dots , \bv_{j-1}$.
That is, $\bv_j$ is found as
\begin{equation}\label{eq:projpur2}
\bv_j \in \underset{{\bv \in S^{D-1}, \bv \perp \bv_1,\dots,\bv_{j-1}}}{\argmax} \rho(\bv^T \bX).
\end{equation}
Note that this is equivalent to solving~\eqref{eq:projpur} after the columns of $\bX$ are projected onto the orthogonal complement of $\Sp(\bv_1,\dots,\bv_{i-1})$. One can also try to find a maximizer of the joint energy $\sum_j \rho(\bv_j^T \bX)$ such that the set of components, $\bv_j \in S^{D-1}$, $j=1,\dots,d$, are pairwise orthogonal~\citep{mccoy2011two}. \citet{mccoy2011two} develop the Maximum Mean Absolute Deviation Rounding (MDR) algorithm, which finds an approximate global maximizer for the joint problem
\begin{equation}\label{eq:projpurjoint}
    \underset{\substack{\bv_1,\dots,\bv_d \in S^{D-1} \\ \bv_j \perp \bv_k, j \neq k}}{\argmax} \sum_{j=1}^d \|\bv_j^T \bX\|_1 .
\end{equation}
We note that~\eqref{eq:projpurjoint} is also known as the maximization variant or $L_1$-PCA, which we discuss further in \S\ref{subsubsec:l1pca}.

\subsection{Least Absolute Deviations}
\label{subsec:lad}

A popular approach to RSR is to replace the least squares formulation in~\eqref{eq:pcamin} with least absolute deviations:
\begin{equation}
\hat{L} \in \argmin_{L \in G(D,d)} \sum_{i=1}^N \| \bQ_L \bx_i \|.
\label{eq:lad}
\end{equation}
This problem has been considered for many reasons, such as its nice interpretation as a geometric median subspace. Indeed, the minimizer of~\eqref{eq:lad} can heuristically be motivated by the geometric median, which solves the least absolute deviations analog for estimating the center of a dataset~\citep{lopuhaa1991breakdown}. Despite being an appealing formulation,~\eqref{eq:lad} is NP-hard to even approximately minimize to an error of order $\Omega(1/\poly(D))$~\citep{clarkson2015input}.

One of the attractive features of using the least absolute deviations formulation is that it is rotationally invariant with respect to choice of basis~\citep{Ding+06}. We clarify this notion of invariance as follows. A subspace in $G(D,d)$ can be represented by an orthonormal system of vectors spanning this subspace. The latter vectors can be identified with the columns of an element of $O(D,d)$. Right multiplication of this element in $O(D,d)$ by an element of $O(d,d)$ results in another semi-orthogonal matrix whose columns still span the same subspace. Therefore, $G(D,d)$ is identified with equivalence classes of $O(D,d)$, and the equivalence relation is obtained by the right action of $O(d,d)$. The cost in~\eqref{eq:lad} is the same for any choice of coordinates within an equivalence class. Indeed, if $L = \Sp(\bU_1) = \Sp(\bU_2)$ for two different matrices $\bU_1, \bU_2 \in O(D,d)$, where $\bU_1 = \bU_2 \bR$ for some $\bR \in O(d,d)$, then
\begin{align}
    \sum_{i=1}^N \| \bQ_L \bx_i \| &= \sum_{i=1}^N \| (\bI - \bU_1 \bU_1^T) \bx_i \| \\ \nonumber
                                   &= \sum_{i=1}^N \| (\bI - \bU_2 \bU_2^T)  \bx_i \|.
\end{align}
This rotational invariance is an essential feature of estimation over the Grassmannian, and not all problem formulations have this (see, e.g., the later formulation for $L_1$-PCA in \eqref{eq:l1max}, which is not rotationally invariant).

Some motivation for this formulation of RSR can also come from relaxing an $\ell_0$ problem to an $\ell_1$ problem, mirroring ideas in compressed sensing. This involes rewriting the function $\#(\cdot)$ in~\eqref{eq:l0} as the $\ell_0$-norm of the vector of distances between the data points and a subspace. One can then relax the $\ell_0$-norm to an $\ell_1$-norm and arrive at~\eqref{eq:lad}.

A recent work that was motivated by the sparse formulation in~\eqref{eq:l0} was originally discussed by~\citep{spath_watson1987} and further analyzed by~\citet{tsakiris2015dual}. However, their formulation is really just least absolute deviations in disguise. Indeed, they iteratively try to find a hyperplane that approximately contains as many points as possible by solving for its normal vector, $\bb$, as follows:
\begin{equation}\label{eq:dpcp}
	\min_{\bb \in S^{D-1}} \| \bX^T \bb \|_1.
\end{equation}
We note that this is equivalent to~\eqref{eq:lad} with $d=D-1$ because $| \bx_i^T \bb | = \| \bQ_{\Sp(\bb)^\perp} \bx_i \|_2$, where $\Sp(\bb)^\perp \in G(D,D-1)$.

We observe that a ``least absolute deviation'' formulation of~\eqref{eq:pcamax} is
\begin{equation}
\hat{L} \in \argmax_{L \in G(D,d)} \sum_{i=1}^N \| \bP_L \bx_i \|.
\label{eq:lad_new}
\end{equation}
Even though the solutions of \eqref{eq:lad} and \eqref{eq:lad_new} may not necessarily be the same, many of the methods developed for \eqref{eq:lad} can be adapted to \eqref{eq:lad_new}. When $d=1$, the projection pursuit procedure in \eqref{eq:projpur} with $\rho(\by) = \| \by \|_1$ coincides with~\eqref{eq:lad_new}. An approximate polynomial-time solution of \eqref{eq:lad_new} for any fixed $d$, within a large absolute factor, was suggested in \cite{naor2013efficient}.

We claim that the least absolute deviations formulation is very amenable to the use of iteratively reweighted least squares (IRLS). It is easiest to explain this claim with the following straightforward argument of \citet{lerman2017fast} for approximating~\eqref{eq:lad} (this can also be adapted to~\eqref{eq:lad_new}). They suggest the iterative procedure
\begin{equation}
    L^{k+1} \in \argmin_{L \in G(D,d)} \sum_{i=1}^N w_i^k \| \bQ_L \bx_i \|^2,
\label{eq:ladirls}
\end{equation}
where $w_i^k = 1/\| \bQ_{L^k} \bx_i\|$. The formulation~\eqref{eq:ladirls} is a weighted PCA problem, which has a direct solution via the SVD of the matrix whose columns are $\{\sqrt{w_i^k} \bx_i\}_{i=1}^N$. Other IRLS approaches for the least absolute deviations problem appear in~\citep{zhang2014novel,lerman2015robust}.

Many methods have been developed to approximate the solution of~\eqref{eq:lad}. We distinguish below between convex relaxations and direct nonconvex strategies.

\subsubsection{Convex Relaxations}

The first relaxation of the least absolute deviations problem was concurrently considered by~\citep{mccoy2011two} and~\citep{xu2012robust}. These works propose the following optimization problem for noiseless RSR
\begin{equation}\label{eq:op}
    \min_{\bL, \bC \in \reals^{D \times N}} \| \bL \|_* + \lambda \| \bC \|_{1,2}, \ \mathrm{s.t.} \ \bL + \bC = \bX.
\end{equation}
Here, $\|\cdot\|_*$ denotes the nuclear norm of a matrix and $\|\cdot\|_{1,2}$ denotes the sum of the column norms of a matrix. The reason why~\eqref{eq:op} relaxes~\eqref{eq:lad} is discussed in the next paragraph. For the noisy case, they consider the problem
\begin{equation}\label{eq:opn}
\min_{\bL, \bC \in \reals^{D \times N}} \| \bL \|_* + \lambda\| \bC \|_{1,2}, \ \mathrm{s.t.} \ \|\bX -(\bL + \bC)\|_F \leq \epsilon,
\end{equation}
where $\epsilon$ is an estimated small noise level. In both algorithms, the parameter $\lambda$ is chosen to be $3/(7\sqrt{N_{\mathrm{out}}})$~\citep{xu2012robust}. Since $N_{\mathrm{out}}$ is not known, one must guess an upper bound on the number of outliers. In practice, with sufficiently small percentages of outliers, the authors argue that one can overestimate $N_{\mathrm{out}}$ and still have good performance, because the algorithm will first remove all outliers and then remove some inliers. The resulting set of inliers can then still recover the correct column space of $\bL$. We have found that choosing $\lambda = 3/(7\sqrt{N_{\mathrm{out}}})$ does not perform well in the settings we test. We instead choose $\lambda = 0.8\sqrt{D/N}$, which seems to work better (this choice was also used in~\citep{zhang2014novel}).

To show that this is a convex relaxation of~\eqref{eq:lad}, one can replace $\|\bL\|_*$ with $\rank(\bL)$. Then, a simple geometric argument shows that $\bC = \bQ_{\Sp(\bL)} \bX$, and so $\|\bC\|_{1,2}$ then measures the deviation of the columns of $\bX$ from the column span of $\bL$. In other words, $\|\bC\|_{1,2}$ is the sum of orthogonal distances between points and the span of $\bL$.

Since~\eqref{eq:op} and~\eqref{eq:opn} form a convex programs, it is possible to optimize them using a range of algorithms. \citet{xu2012robust} advocate using a proximal gradient algorithm. They refer to the problems in~\eqref{eq:op} and~\eqref{eq:opn} as ``outlier pursuit'', for which we use the acronym OP.

Two other algorithms, the Geometric Median Subspace (GMS)~\citep{zhang2014novel} and REAPER~\cite{lerman2015robust}, are also convex relaxations of the robust energy in~\eqref{eq:lad}.
GMS seeks a relaxed orthoprojection onto the orthogonal complement of an underlying subspace through robustly estimating the inverse covariance matrix. The GMS estimator is constructed through the convex relaxation
\begin{align}\label{eq:gms}
&\hat{\bQ} = \argmin_{\bQ \in \mathbb{H}} \sum_{i=1}^N \| \bQ \bx_i\|,  \\ \nonumber
& \ \ \ \mathbb{H} = \{ \bQ \in \reals^{D \times D} : \bQ = \bQ^T, \Tr(\bQ) = 1\}.
\end{align}
The underlying subspace is then estimated from the bottom eigenvectors of $\hat{\bQ}$.
On the other hand, REAPER solves a tighter convex relaxation designed to robustly estimate the orthoprojector onto the underlying subspace $L_*$. The convex program is
\begin{align}
\label{eq:reaper}
	&\hat{\bP} = \argmin_{\bP \in \mathbb{G}} \sum_{i=1}^N \| (\bI - \bP) \bx_i\|,  \\ \nonumber
	&\ \ \ \mathbb{G} = \{ \bP \in \reals^{D \times D} : \bzero \preceq \bP \preceq \bI , \Tr(\bP) = d\}.
\end{align}
Here, the estimated subspace is calculated as the top eigenvectors of $\hat{\bP}$.
Note that $\mathbb{G}$ is the convex hull of the set of orthoprojectors of rank $d$~\citep{lerman2015robust}. Thus, by identifying subspaces with their orthoprojectors, we see that~\eqref{eq:reaper} is the tightest convex relaxation of~\eqref{eq:lad}. We remark that the minimizer of~\eqref{eq:reaper} does not change if the constraint $\preceq \bI$ is removed (see proof of Lemma 14 in~\citep{zhang2014novel}).
Therefore, one may note that~\eqref{eq:gms} is obtained from~\eqref{eq:reaper} by setting $\bP=\bI-\bQ$ and dropping the constraint $\bP\succeq \bzero$. Indeed, after doing this, any fixed value of $\Tr(\bQ)$ yields the same subspace.
Both of these algorithms employ IRLS procedures to efficiently solve their respective optimization problems.

\subsubsection{Nonconvex Optimization}
\label{subsec:nonconvlad}

An alternative to convex relaxation of~\eqref{eq:lad} is to attempt to directly minimize this energy function. The advantage of doing this is that one can obtain faster algorithms for special settings. However, these algorithms are typically hard to theoretically justify, despite their impressive practical performance. Only recently have theoretical results shown the strength of these methods in certain regimes.

\citet{Ding+06} considered direct optimization of the nonconvex program in~\eqref{eq:lad}, which they incorrectly assumed was convex. To do this, they use a form of the power method (see the method of orthogonal iteration in \S8.2.1 of~\citep{golub1996matrix}). This algorithm is referred to as Rotational Invariant $L_1$-norm PCA (R1PCA). This method is somewhat problematic since the optimization technique they use is tied to convex methods and may lead to poor solutions in the nonconvex case.

Direct optimization of~\eqref{eq:lad} on $G(D,d)$ was later considered in the sequence of works by~\citet{lerman2017fast} and~\citet{maunu2017well}.
In~\citep{lerman2017fast}, the authors directly use IRLS on~\eqref{eq:lad}. The resulting method is called the Fast Median Subspace algorithm (FMS). In the next work~\citep{maunu2017well}, they use a geodesic gradient descent method to minimize~\eqref{eq:lad} over $G(D,d)$ by drawing on ideas from~\citep{Edelman98thegeometry}. In practice, FMS seems to perform better than GGD, but existing theoretical guarantees for GGD are stronger.

Another work that attempts to approximately minimize the least absolute deviations energy is given in~\citep{clarkson2015input}. Their algorithm, called ConstApprox, also accounts for sparse inputs, which yields reduced computational complexity for sparse matrices.
The approximation method can return a $(1+\epsilon)$ approximation to the minimum value of the program in~\eqref{eq:lad} for sufficiently large $\epsilon$. On the other hand, they show that the approximation problem becomes NP-hard when $\epsilon = \Omega(1/\poly(D))$.

Another nonconvex optimization method closely tied to~\eqref{eq:lad} and the outlier pursuit relaxation came in~\citet{cherapanamjeri2017thresholding}, where the authors coin their algorithm Thresholding based Outlier Robust PCA (TORP).
The authors use a nonconvex thresholding based algorithm, which iterates between fitting a PCA subspace and filtering points that are either far from the subspace or highly incoherent. The definition of incoherence is later given in \S\ref{subsec:optheory}. A disadvantage of this method is that it requires the user to input the percentage of outliers, which is not known in practice. As in OP, one can overestimate the percentage of outliers and still have accurate recovery when the percentage of outliers is sufficiently small.

\citet{tsakiris2015dual} proposed the Dual Principal Component Pursuit (DPCP) algorithm that sequentially fits nested hyperplanes by finding stationary points in the program~\eqref{eq:dpcp}.
For solving~\eqref{eq:dpcp}, they follow an algorithm of~\citep{spath_watson1987}, which uses an alternating sequence of convex relaxation followed by a nonconvex projection. More precisely, the sequence $(\tilde{\bb}^k)_{k \geq 1}$ is defined by the following program:
\begin{equation}\label{eq:dpcpseq}
    \bb^{k+1} \in \argmin_{\bn^T \tilde{\bb}^k = 1 } \|\bX \bn \|_1, \   \tilde{\bb}^{k+1} = \frac{\bb^{k+1}}{\| \bb^{k+1} \|}.
\end{equation}
Notice that the minimization in~\eqref{eq:dpcpseq} just involves solving a linear program at each iteration.
After one hyperplane is found (i.e., the $(D-1)$-subspace perpendicular to the limit of this sequence), the DPCP procedure searches for a hyperplane of this hyperplane, which results in a $(D-2)$-subspace. This procedure is repeated until one is left with a $d$-subspace.

\subsection{$L_1$-PCA}
\label{subsubsec:l1pca}

There are two different variants of $L_1$-PCA that we discuss here: the minimization and maximization based formulations~\citep{brooks2013pure,markopoulos2018outlier}. It seems that the minimization based variant is more closely tied to the RPCA problem reviewed in~\citep{vaswani2018static}, while the maximization variant seems to be tied to joint projection pursuit and thus is more closely related to RSR.

The minimization formulation of $L_1$-PCA forms the following analog of \eqref{eq:pcamin}:
\begin{equation}\label{eq:l1min}
    \bU_{L_1-\min} \in \argmin_{\substack{\bU \in O(D,d) \\ \{\by_i\}_{i=1}^N \subset \reals^d}} \sum_{i=1}^N \| \bx_i - \bU \by_i \|_1.
\end{equation}
 One can also write~\eqref{eq:l1min} as
\begin{equation}\label{eq:l1min2}
    \bU_{L_1-\min} \in \argmin_{\substack{\bU \in O(D,d) \\ \bY \in \reals^{d \times N}}}  \| \bX - \bU \bY \|_{1,1},
\end{equation}
where the $\| \cdot \|_{1,1}$-norm sums the absolute values of the matrix elements. This formulation is equivalent to PCA when one uses the squared $L_2$-norm instead of the $L_1$-norm. Indeed, one can write the PCA minimization as
\begin{equation}
\min_{\bU \in O(D,d),\bY \in \reals^{d \times N}} \|\bX - \bU \bY\|_{2,2},
\end{equation}
where $\| \cdot \|_{2,2}$ corresponds to the Frobenius norm. Notice that the minimization in~\eqref{eq:l1min} is rotationally invariant in the sense of \S\ref{subsec:lad} since we can write
\begin{equation}\label{eq:l1min3}
    \bU_{L_1-\min} \in \argmin_{\substack{\bU \in O(D,d) \\ \{\by_i\}_{i=1}^N \subset \reals^d}} \sum_{i=1}^N \| \bx_i - \bU\bR \bR^T\by_i \|_1,
\end{equation}
and cast the optimization over the variables $\bU' = \bU\bR$ and $\bz_i = \bR^T\by_i$.

A perhaps simpler equivalent formulation of the minimization in~\eqref{eq:l1min} is given by
\begin{equation}\label{eq:l1min4}
    \hat{\bL} \in \argmin_{\rank(\bL) \leq d} \| \bX - \bL \|_{1,1}.
\end{equation}
The subspace estimate can then be found from the span of $\bL$.

We remark that \eqref{eq:l1min4} can be viewed a non-convex relaxation of the following problem
\begin{equation}\label{eq:rpca1}
\hat{\bL} \in \argmin_{\rank(\bL) \leq d} \| \bX - \bL \|_{0,0},
\end{equation}
where the $\| \cdot \|_{0,0}$ is just the number of non-zero entries of a matrix.
The later problem is in fact the RPCA problem, where one seeks low-rank approximation to a matrix with sparse corruptions. Attempts to find approximate solutions for this problem are discussed in \cite{vaswani2018static}.

The nonconvex and nonsmooth minimization problem in~\eqref{eq:l1min} was originally considered in~\citet{Baccini96}, where the authors show that this choice of norm is equivalent to finding the maximum likelihood estimate (MLE) subspace under a Laplacian noise assumption (rather than Gaussian for PCA). Further convex relaxation algorithms were developed by~\citep{ke2005robust} and later by a more recent surge of work (see~\citet{yu2012efficient} and~\citet{brooks2013pure} for some examples). \citet{brooks2013pure} give a nonconvex, polynomial time algorithm for the special case of $d=D-1$. \citet{gillis2015complexity} showed that this minimization problem is NP-hard for $d<D-1$. \citet{song2017low} study approximate minimization of this quantity, where they derive a polynomial time algorithm to approximate the minimizer up to a given threshold.

We emphasize that while the minimization variant of $L_1$-PCA is a natural robust extension of PCA, it may not be ideal for solving the RSR problem discussed in this paper. Indeed, the formulation in \eqref{eq:l1min4} and its MLE interpretation seem to be more robust to elementwise corruption than to outliers.

Unlike least absolute deviations, the minimization variant of $L_1$-PCA does not have a simple IRLS formulation to take advantage of. Indeed, the elementwise weighting procedure presents some issues.
For example, similar to the idea summarized in \eqref{eq:ladirls}, one could try to apply the following IRLS procedure to approximate~\eqref{eq:l1min4}:
\begin{equation}\label{eq:l1irls}
        \bL^{k+1} \in \argmin_{\rank(\bL) \leq d} \sum_{i,j} w^k_{ij} (\bX_{ij} - \bL_{ij})^2,
    \end{equation}
where $w^k_{ij} = 1/|\bX_{ij} - \bL^k_{ij}|$. However, this least squares problem has no straightforward solution at each iteration~\citep{srebro2003weighted}.
One could use a strategy like the alternating least squares algorithm presented by~\citet{Torre:03} for solving~\eqref{eq:l1irls} with different robust weights $w_{ij}^k$. However, there would be no guarantee of globally minimizing the least squares problem at each iteration.

The maximization formulation of $L_1$-PCA is given by
\begin{equation}\label{eq:l1max}
\bU_{L_1-\max} \in \argmax_{\bU \in O(D,d)} \sum_{i=1}^N \| \bU^T \bx_i \|_1.
\end{equation}
Note that~\eqref{eq:l1min} and~\eqref{eq:l1max} are the $L_1$-PCA versions of~\eqref{eq:pcamin} and~\eqref{eq:pcamax}, respectively. However, while~\eqref{eq:pcamin} and~\eqref{eq:pcamax} are equivalent,~\eqref{eq:l1min} and~\eqref{eq:l1max} are not.
The $L_1$-PCA version in~\eqref{eq:l1max} is actually a special case of joint energy projection pursuit. If one considers the joint projection pursuit energy from \S\ref{subsubsec:projpur}, $\sum_{j=1}^d \rho(\bv_j^T \bx_i)$, with $\rho(\bx) = \| \bx \|_1$ over orthonormal sets $\{\bv_1,\dots,\bv_d\}$, one arrives at precisely the formulation in~\eqref{eq:l1max}.
Therefore, like projection pursuit, the formulation in~\eqref{eq:l1max} addresses the RSR problem. It thus has different characteristics than the formulation in~\eqref{eq:l1min}, which is tied to the RPCA problem.
We remark that there is no straightforward maximum likelihood interpretation of~\eqref{eq:l1max}, unlike~\eqref{eq:l1min}.

Notice that the formulation in~\eqref{eq:l1max} is not rotation invariant with respect to choice of basis, unlike the formulation in~\eqref{eq:l1min}. Indeed, if $\bR \in O(d,d)$, then unlike the Euclidean norm, $\| \bU^T \bx_i \|_1 \neq \| \bR^T \bU^T \bx_i \|_1$ in general. Thus, this formulation is not truly over $G(D,d)$. If instead we wish to formulate~\eqref{eq:l1max} over $G(D,d)$, we should try to solve
\begin{equation}\label{eq:l1max2}
\bU_{L_1-\max}' \in \argmax_{\bU \in O(D,d)} \sum_{i=1}^N \| \bU \bU^T \bx_i \|_1.
\end{equation}
Indeed, since $\| \bU \bU^T \bx_i \|_1 = \| \bU \bR \bR^T \bU^T \bx_i \|_1$, we have rotation invariance with respect to choice of basis. We are not aware of work focusing on the maximization in~\eqref{eq:l1max2}.

For both of the maximization problems in~\eqref{eq:l1max} and~\eqref{eq:l1max2}, one could come up with IRLS formulations as was done for~\eqref{eq:l1min} in~\eqref{eq:l1irls}. However, the same issues arise as before since it is not an easy task to solve the least squares portion of the algorithm.

For large $N$ and $D$, the maximization problem in~\eqref{eq:l1max} is NP-hard~\citep{mccoy2011two}. Nevertheless, \citet{kwak08} first developed an algorithm that sequentially outputs local maxima of the one-dimensional version of~\eqref{eq:l1max}. Later, exact algorithms were developed by~\citet{markopoulos2014optimal} for sufficiently small $N$ and $D$. An approximate polynomial-time solution of \eqref{eq:l1max}, within a large absolute factor, was suggested in \cite{naor2013efficient}. Their work improves over an earlier $O(\log(N))$ approximation factor in \cite{mccoy2011two}. A review of algorithms and methods for the $L_1$-maximization problem in~\eqref{eq:l1max} appears in \citep{markopoulos2018outlier}.

\subsection{Robust Covariances}
\label{subsec:robcov}

Another line of thought has considered robustly estimating the { underlying} covariance matrix of a dataset~\citep{maronna1976robust,stahel1981breakdown,donoho1982breakdown,tyler_dist_free87,dumbgen1998tyler,marden1999some,locantore1999robust,visuri2000sign,maronna2006robust,zhang2014novel,nordhausen2015cautionary,zhang2016robust}, which can then be used to locate underlying subspaces. The simplest setting assumes that the population mean, $\bmu$, is $\bzero$. After calculating the robust covariance estimator, one can find the robust principal subspace from its top eigenvectors. The direct synthesis of these ideas with the problem of subspace recovery can be seen in~\citep{zhang2014novel,zhang2016robust}.

One example is the Maronna M-estimator~\citep{maronna1976robust}. It minimizes a certain robust energy that is a maximum likelihood covariance estimator under an elliptical distribution with heavy tails. More precisely, it is the minimizer of
\begin{equation}\label{eq:maronnaM}
\frac{1}{N} \sum_{i=1}^N \rho(\bx_i^T \bSigma^{-1} \bx_i) + \frac{1}{2} \log \det(\bSigma)
\end{equation}
over all positive definite $\bSigma$, where $\rho$ is a function that satisfies certain conditions. Similarly, the Tyler M-estimator (TME)~\citep{tyler_dist_free87} minimizes the energy
\begin{equation}\label{eq:tme}
    \frac{1}{N} \sum_{i=1}^N \log(\bx_i^T \bSigma^{-1} \bx_i) + \frac{1}{D} \log \det(\bSigma),
\end{equation}
among all $\bSigma$ positive definite with trace 1. A more in depth discussion of these energy functions and their robustness is given in Appendix~\ref{app:robcov}.

The advantage of~\eqref{eq:maronnaM} and~\eqref{eq:tme} is that their formulations are geodesically convex~\cite{auderset2005angular,wiesel_geodesic_convexity12,zhang2016robust, teng_ami_maria_geo_convex}.
Both estimators can be iteratively computed by an IRLS procedure.
When $D>N$, these estimators are undefined~\citep{maronna1976robust,tyler_dist_free87}, and even when $D \leq N$ they may be ill-conditioned. It is thus common to regularize them \citep{sun2014regularized,ollila2014regularized}.

Perhaps the simplest robust covariance estimator is the spherical sample covariance~\citep{locantore1999robust,maronna2006robust,lerman2017fast}, which can be estimated as
\begin{equation}\label{eq:spherecov}
\hat{\bSigma} = \frac{1}{N} \sum_{i=1}^N  \frac{\bx_i \bx_i^T}{\|\bx_i \|^2}  .
\end{equation}
Spherical PCA (SPCA) computes the principal subspace of this estimator, which is the PCA subspace of the normalized dataset $\{\bx_i/\|\bx_i\|\}_{i=1}^N$ \citep{locantore1999robust}.

In a more general setting, both the mean, $\bmu$, and the covariance, $\bSigma$, are unknown. If one only cares about estimating the covariance, then one can calculate the estimators above on the set of differences between data points, $\bx_i - \bx_j$ for $i \neq j$, $i,j = 1,\dots,N$. For example, the spatial Kendall's tau matrix~\citep{visuri2000sign} estimates the spherical covariance by
\begin{equation}\label{eq:kendtau}
\hat{\bSigma} = \frac{2}{N(N-1)} \sum_{i \neq j} \frac{(\bx_i - \bx_j) (\bx_i - \bx_j)^T}{\| \bx_i - \bx_j \|^2} .
\end{equation}
Similarly, D\"umbgen's M-estimator~\citep{dumbgen1998tyler} computes TME on the set of differences between points, and \citet{nordhausen2015cautionary} apply this procedure, which they refer to as symmetrization, to other robust covariance estimators.
These estimators can address RSR in the affine setting. Indeed, an affine subspace can be decomposed into a linear subspace plus an offset. The estimated linear subspace is the principal subspace of the ``symmetrized'' robust covariance estimator, which is expected to approximate the underlying linear component. On the other hand, the offset could be well-approximated by a robust point estimator, such as the geometric median. A benefit of symmetrization is that it avoids estimating the offset first and centering the data at this offset. With the latter procedure, small approximation error of the offset may result in large approximation error of the linear subspace component.

\subsection{Other Energy Minimizers}
\label{subsec:otheren}
The methods reviewed so far were formulated by energy minimization or by maximization of a utility function. Another example is given by~\citet{Xu1995}, who tried to minimize a trimmed version of the PCA energy given by
\begin{equation}
	\min_{L \in G(D,d)} \sum_{i=1}^N \begin{cases}
	\| \bx_i - \bP_L \bx_i \|^2, &\| \bx_i - \bP_L \bx_i \|^2 < \eta, \\
	\eta, &\| \bx_i - \bP_L \bx_i \|^2 \geq \eta.
	\end{cases}
\end{equation}
The motivating idea is that trimming the energy would give robustness to outliers, while maintaining some desirable characteristics of PCA. An additional example of a method that aims to maximize a utility function appears below in \eqref{eq:hrpca_utility}.

\subsection{Filtering Outliers}
\label{subsec:filter}

One way of attempting RSR is to first filter outliers and then fit a subspace to the data by using PCA. A simple filtering idea is to use affinities that express presence in an underlying subspace (or multiple underlying subspaces) to screen and remove outliers. The first recipe was suggested by~\citet{chen2009spectral} (see, in particular, \S3.1). They form a symmetric weight matrix that aims to express the likelihood that pairs of points lie on an ``underlying $d$-dimensional subspace'', that is, a subspace that many other data points lie on. The degrees of the data points are then computed from this weight matrix, where a degree of a data point is the sum of weights in the corresponding row of the matrix. The outliers are identified as points with low degree, or in other words, points with low likelihood of being contained in a $d$-subspace. This idea can be used in the setting of robust subspace recovery and also in the setting of robust subspace clustering. In the latter setting, inliers lie on a union of subspaces and the goal is to recover these subspaces in the presence of outliers. A similar idea is suggested in \cite{ariascastro2011spectral} for the more general setting of robust manifold clustering, where inliers lie on a union of manifolds and the goal is to recover these manifolds in the presence of outliers.

\citet{soltanolkotabi2012}, whose ideas build on those in \cite{ssc_elhamifar13}, also identify outliers according to low degrees of a weight matrix. Their expression for the degree of a data point $\bx_j \in \reals^D$ is the value of the following program:
\begin{equation}\label{eq:ssc}
    \min_{\br_j  \in \reals^N} \|\br_j \|_1, \ \text{s.t.} \ \bX_{-j} \br_j  = \bx_j,
\end{equation}
where $\bX_{-j}$ is the $D \times N$ data matrix with the $j$th column zeroed out.
To relate this idea to the framework of~\citep{chen2009spectral}, one can form the asymmetric $N\times N$ weight matrix $\bR$, whose $j$th row is the vector $\br_j$ minimizing~\eqref{eq:ssc}. Clearly, the $j$th degree of $\bR$ (i.e., the sum of the weights in row $j$) is the minimal value in~\eqref{eq:ssc}.
\citet{YouRV17} use a similar matrix $\bR$, which is formed by elastic net minimization instead of pure $\ell_1$-norm minimization, to create a random walk over the nodes of the graph. They iterate $\bR$ in an interesting way to obtain a limiting vector that aims to be supported on the inliers of the robust subspace clustering problem. They refer to this method as Self-Representation Outlier Detection (SRO).

The recent Coherence Pursuit (CP) method by~\citet{rahmani2017coherence} follows the initial framework in~\citep{chen2009spectral} of identifying outliers according to low degrees in a certain $N \times N$ symmetric weight matrix. The weight matrix is denoted by $\bW$, where the weight $\bW_{ij}$ is the absolute value of the dot product or squared dot product of the normalized vectors $\bx_i / {\|\bx_i\|}$ and $\bx_j / {\|\bx_j\|}$. Among all the above methods that fall into the same framework (with possibly asymmetric weight matrices), this is the fastest to compute. However, it is somewhat simplistic when considering various outlier regimes. To speed up the algorithm, the authors mention using sketching to reduce computational complexity. In noisy settings, this algorithm struggles since it only takes the span of the top $d$ points. Thus,~\citet{rahmani2017coherence} propose a column sampling procedure, which iteratively projects the dataset and takes the most coherent point in an alternating fashion. This strategy is repeated until one recovers a sizeable set of points, and the underlying subspace is estimated  from the recovered set of points using PCA. However, this method requires setting extra user-specified parameters, and in particular, requires an estimate of the noise level, which is not known in practice.

\citet{xu2013outlier} developed the method of high-dimensional robust PCA (HR-PCA), which adaptively trims points to obtain a robust estimator. This method tries to maximize a robust variance estimator to capture subspace structures. Given a bound on the number of inliers, $\hat{t}$, the trimmed variance maximization is defined as
\begin{equation}
\label{eq:hrpca_utility}
    \hat{\bU} \in \argmax_{\bU \in O(D,d)} \max_{\substack{I \subset \{1,2,\dots,N\} \\ \#(I) = \hat{t}}} \sum_{i \in I} \| \bU^T \bx_i \|^2.
\end{equation}
The authors develop a randomized algorithm, where at each iteration, a point is removed with probability proportional to its variance in the current direction. The process is continued until one removes a prespecified number of points. The robust subspace can then be calculated from the remaining points. A deterministic version of this algorithm, called DHRPCA, was later developed in~\citep{Feng:12}. While the method can remove outliers with high influence on the PCA subspace, it is unintuitive as to why it should work in general settings with other more subtle types of outliers. Also, the algorithm requires the user to input the percentage of outliers, which is unknown in practice.

The idea of filtering outliers is also present in the work on adaptive compressive sampling (ACOS)~\citep{li2015identifying}. Here, the authors subsample points and coordinates of the dataset, run outlier-pursuit or some other robust method, and filter outliers from the subsampled data. A subspace for the whole dataset can then be fit from the unfiltered points in full dimension.

The TORP algorithm~\citep{cherapanamjeri2017thresholding}, as discussed earlier in \S\ref{subsec:lad}, can also be thought of as an outlier filtering method.

\subsection{Exhaustive Subspace Search Methods}
\label{subsec:exhaust}

Another classical and simple way of robustly finding a subspace is to use RANSAC. In the celebrated paper,~\citet{fischler1981random} propose a general method where a subsample and estimator are iteratively improved over a dataset.
Since this is such a common procedure, we review a RANSAC variant for RSR in more detail. The basic idea is to randomly sample $O(d)$ points and fit a $d$-subspace to them by using PCA. Then, one calculates the distances between all points and this subspace and labels inliers as those with distances less than an input consensus threshold.
If the set of inliers labelled in this way is sufficiently large (determined by comparison with an input consensus parameter), the algorithm returns this subspace. Otherwise, after a predetermined number of iterations, the algorithm outputs the model with highest consensus number.

\citet{hardt2013algorithms} proposed the RandomizedFind algorithm (RF), which is an exhaustive search method that is faster than RANSAC.
For noiseless subspace recovery of a dataset $\cX \subset \reals^D$ with $N>D$
and where the inliers and outliers are in some general position as described in \S\ref{sec:theory},
they take random subsets, $\tilde{\cX}$, of size $D$ from $\cX$ until one is found with $\rank(\tilde{\cX}) < D$.
Then this subset must contain at least $d+1$ inliers and the indices of these inliers can be found by the non-zero elements of a vector in the kernel of $\tilde{\cX}$.
In order to deal with some noise, they propose replacing the condition $\rank(\tilde{\cX}) < D$ with $\det(\tilde{\bX}^T \tilde{\bX} ) < \delta$, where $\delta>0$ is some small constant and $\tilde{\bX}$ is the data matrix corresponding to $\tilde{\cX}$.
Finally, they also derive DeRandomizedFind (DRF), a deterministic polynomial time version of the RandomizedFind algorithm.
Inspired by RF,~\citet{ariascastro2017ransac} studied a variant of RANSAC that subsamples $(d+1)$-subsets of points until a linearly dependent subset is found.

The scan statistic~\citep{glaz1999scan,Glaz01} can also be used to exhaustively
search for the underlying subspace in a structured way. This statistic measures
the maximal number of occurrences in a sliding window of a fixed length.
\citet{Arias-Castro05connect} { proposed using the scan statistic in a multi-scale, multi-orientation fashion  for the more general problem of
robust manifold recovery. In this problem, inliers are uniformly sampled from a sufficiently smooth surface in $[0,1]^D$, outliers are uniformly distributed in $[0,1]^D$ and one needs to recover the underlying manifold.}

\subsection{Parallel Works}
\label{sec:parallel}

Here we discuss some different but related works to the RSR problem. Some of them have contributed to the development of RSR algorithms, while others have solved similar yet different problems.

One cannot consider RSR without acknowledging work done on robust orthogonal regression and its subsequent extension to RSR~\citep{osborne_watson85,spath_watson1987,Nyquist_l1_88,watson2001some,maronna2005principal}. In this problem, one fits a $(D-1)$-dimensional subspace in $\reals^D$, that is, an element of $G(D,D-1)$, using orthogonal distance as an error metric. The methods in this line of work use least absolute deviations to obtain robustness to corrupted data points.

Another body of related work, which was mentioned earlier is the RPCA problem~\citep{candes2011robust,chandrasekaran2011rank,vaswani2018static}. A large variety of works have contributed to the study of this problem, such as robust energy minimization~\citep{Torre:03,WSL13}, works on convex optimization~\citep{candes2011robust,chandrasekaran2011rank}, online versions~\citep{qiu2010real,he2011online,feng2013robpca}, nonconvex optimization~\citep{WSL13,netrapalli2014non,yi2016fast}, RANSAC methods~\citep{pimentel2017random}, and many others~\citep{vaswani2018static}. Developments in RPCA and RSR seem to be somewhat complementary, and similar emergent themes can be seen in both.

Some other related problems, such as subspace clustering, synchronization, camera location estimation, and sparse vector estimation are discussed later in \S\ref{sec:rsr_influence}.

\section{Theoretical Recovery Guarantees}
\label{sec:theory}

The theory behind algorithms for RSR has come in many forms, and it is hard to make sense of what the theory indicates about these algorithms. While there are many heuristic justifications for the methods discussed in the previous section, it is important to compare and contrast the various guarantees in order to gain an understanding of the most competitive methods. In this section, we attempt to distill the current recovery guarantees given for the RSR strategies. As a result, we hope to shed some light on where the field can go next. We leave the other important theoretical aspect of estimating the computational complexity of the algorithms, and in particular, rate of convergence of iterative schemes, to \S\ref{sec:comp}.

In the following, we discuss exact recovery guarantees and near recovery guarantees. Exact recovery refers to a method's ability to exactly estimate the underlying subspace of a given noiseless inlier-outlier dataset. On the other hand, with noisy inliers-outliers datasets, one cannot hope to exactly estimate the underlying subspace. Instead, guarantees in the noisy case focus on near recovery, which means that the method finds a good approximation to the underlying subspace.
Error of approximation in near recovery is typically bounded by a function of the noise level.

We explain the primary assumptions that seem to be shared among all works on RSR and the common RSR models in~\S\ref{subsec:statmod}.
We explain the theoretical work on RSR in~\S\ref{sec:sequential}-\S\ref{subsubsec:exhaust}  following the categories given in \S\ref{sec:basics}. We remark that \S\ref{sec:sequential} also includes a discussion of the limitation of sequential methods. We conclude in \S\ref{subsec:haystack} with a comparison of the guarantees of these various methods for a specific statistical model of data.

\subsection{Assumption on and Models of Data}
\label{subsec:statmod}

The primary assumption for RSR is that inliers lie on or near a fixed underlying subspace, $L_*$, while outliers lie in the ambient space. For simplicity, we assume in most of the theoretical discussion the noiseless case, where the inliers lie exactly on the subspace. At times, we also comment on extensions to some noisy settings. We also assume in most of this paper that the dimension of $L_*$, $d$, is known. That is, we assume a noiseless (or sometimes noisy) RSR inlier-outlier dataset with known $d$, where one needs to recover (or nearly recover) the underlying $d$-subspace, $L_*$.

Here, we broadly describe the underlying statistical and combinatorial models involved in subspace recovery. An understanding of these models is essential in understanding the development of the field. We first describe several artificial examples in which the RSR problem is not well-defined and use them to motivate two basic principles for theoretical inlier-outlier datasets. These principles have to be followed in order to formulate well-defined theoretical data models.

In \S\ref{subsec:theoryperm}, we lay out a principle for inlier distributions in the RSR problem. Then, \S\ref{subsec:theoryalign} gives a corresponding principle for outlier distributions. In \S\ref{subsec:theorystab}, we briefly mention the combination of these two principles to ensure well-defined models. Finally, in \S\ref{subsec:statmod_quantify}, we carefully review specific theoretical data models that have been used for RSR in the context of these two principles.

\subsubsection{Restrictions on the Inliers and a First Principle}
\label{subsec:theoryperm}

\begin{figure*}
	\centering
	\begin{subfigure}{.47\textwidth}
		\includegraphics[width=\textwidth]{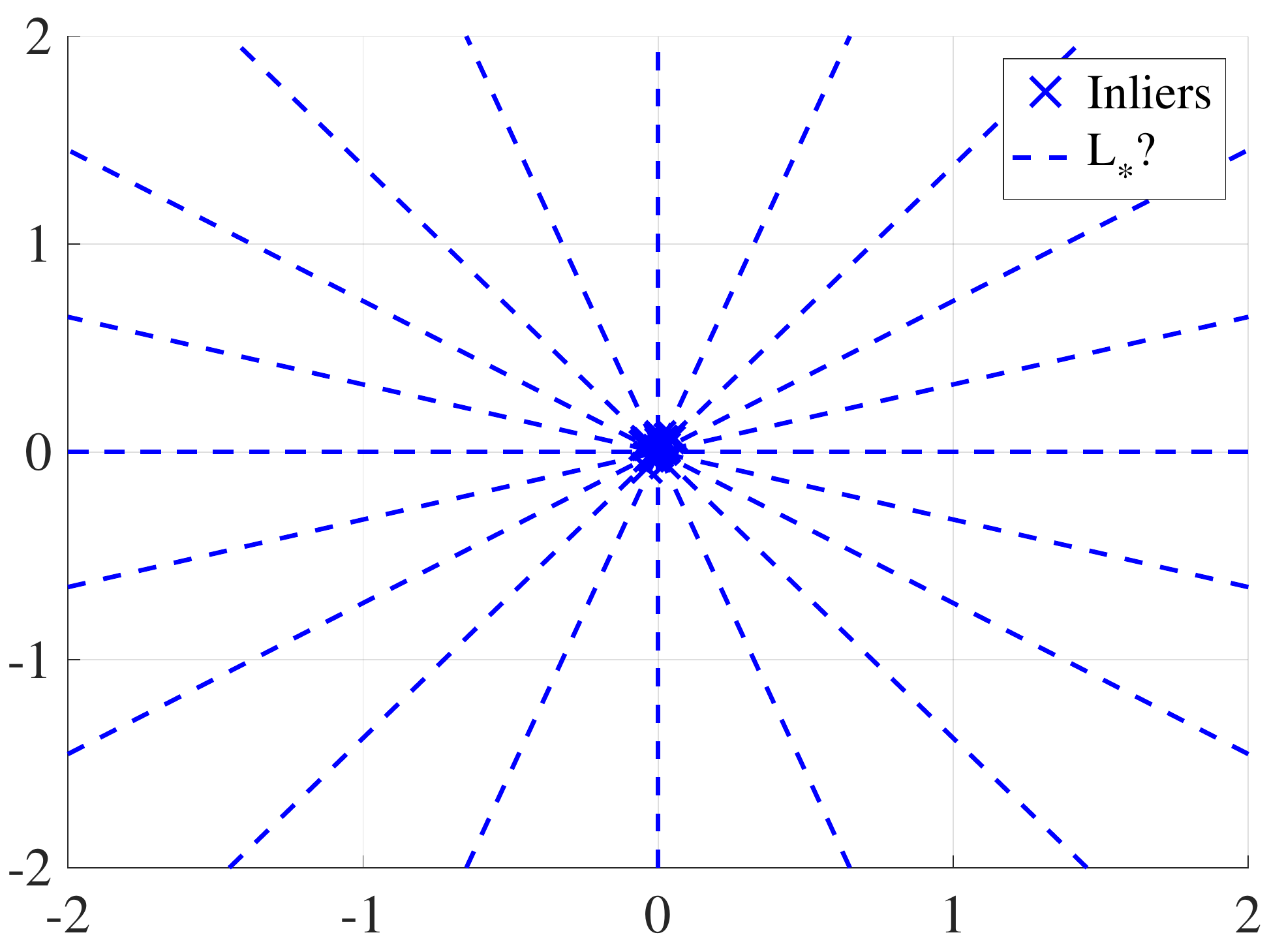}
		\caption{\label{subfig:inorigin} The artificial data is composed of only inliers lying at the origin. Any line through the origin could be the underlying subspace.}
	\end{subfigure}
	\hfill
	\begin{subfigure}{.47\textwidth}
		\includegraphics[width=\textwidth]{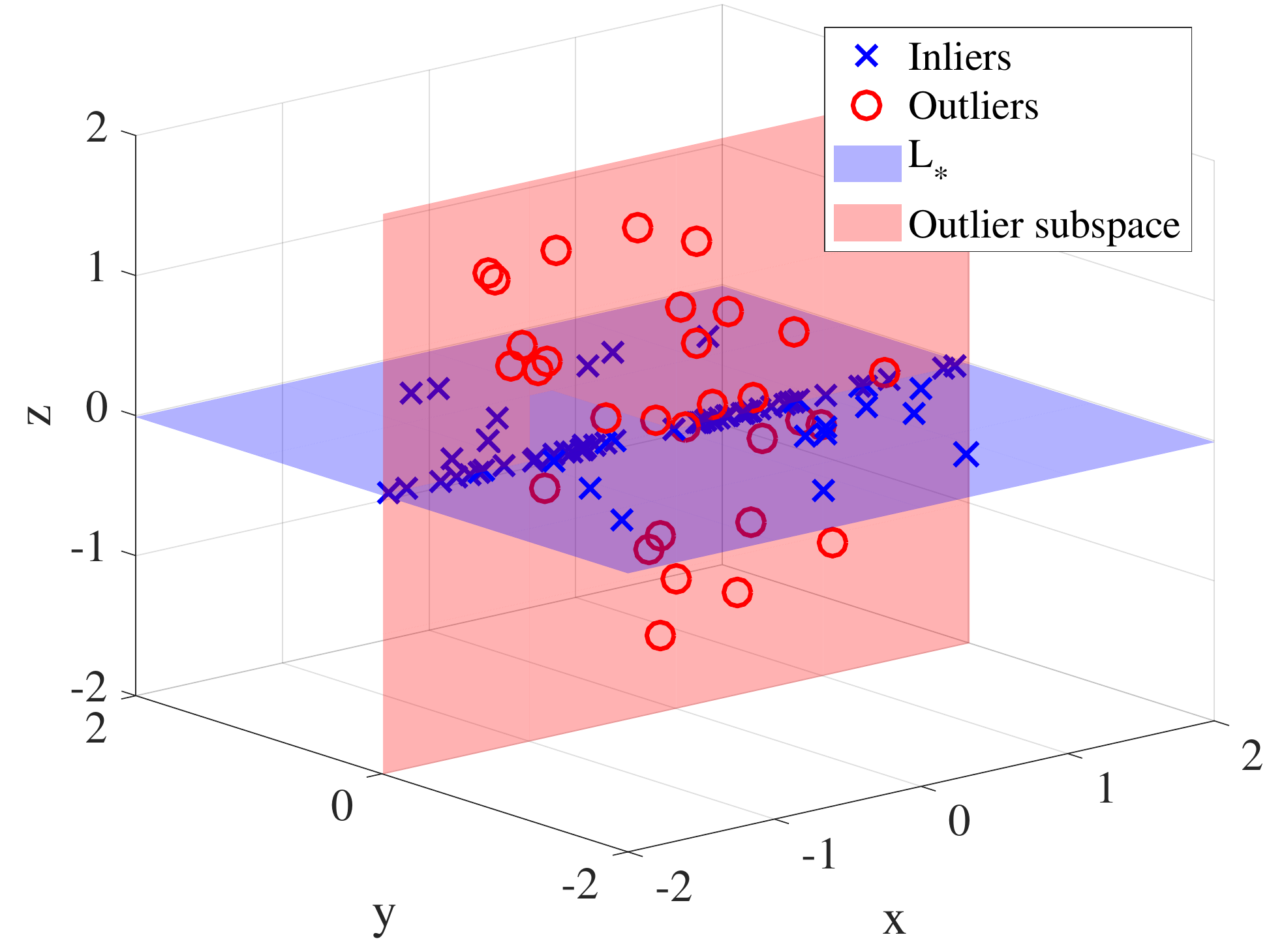}
		\caption{$L_*$ is the $xy$ plane and the inliers concentrate around a line (the $x$ axis). The outliers lie in the $xz$ plane, and this subspace seems to capture more of the data points than $L_*$. \label{subfig:lowdimin}}
	\end{subfigure}
	\begin{subfigure}{.47\textwidth}
		\includegraphics[width=\textwidth]{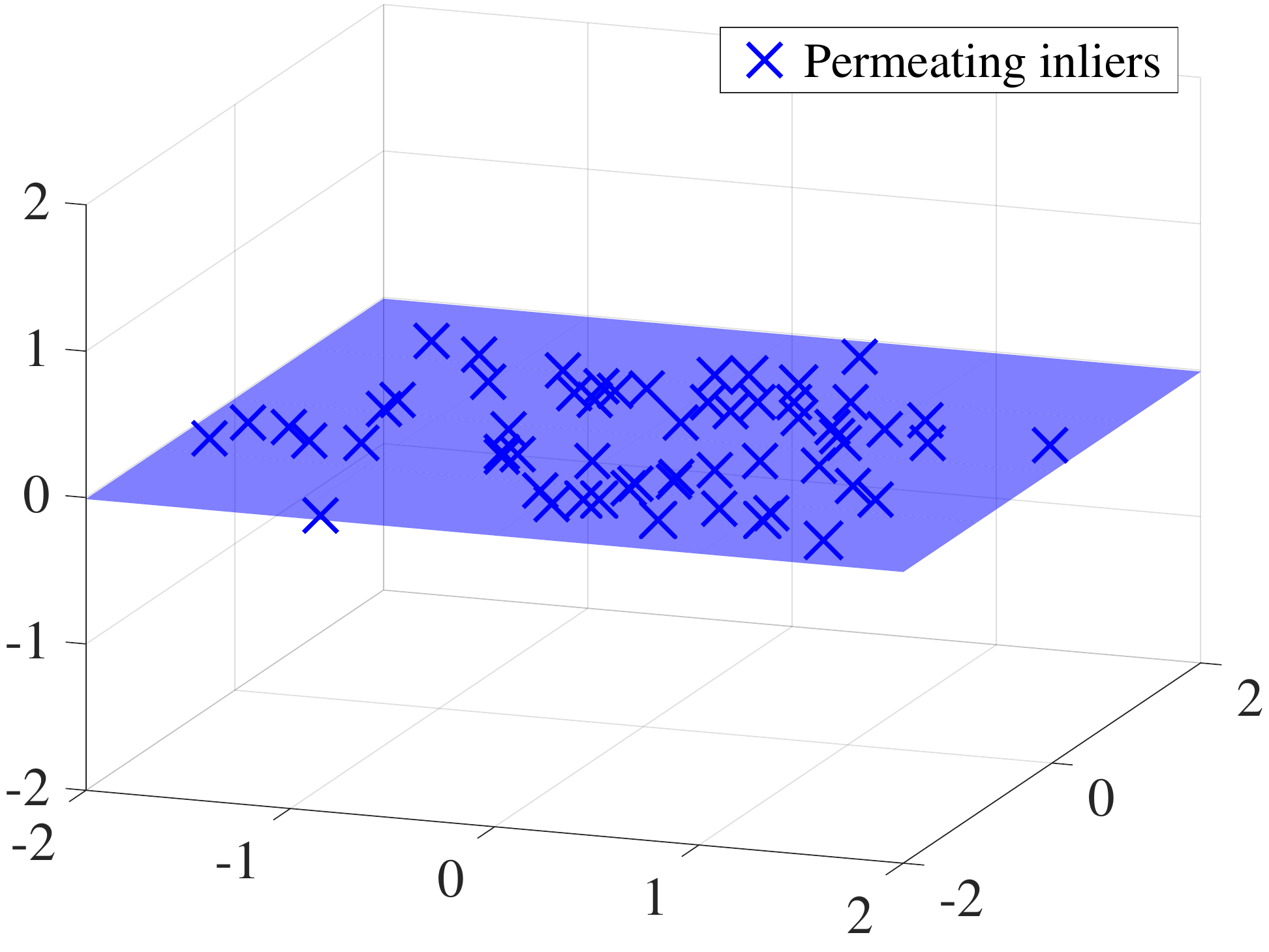}
		\caption{\label{subfig:permin} A cartoon of permeating inliers.}
	\end{subfigure}
	\hfill
	\begin{subfigure}{.47\textwidth}
		\includegraphics[width=\textwidth]{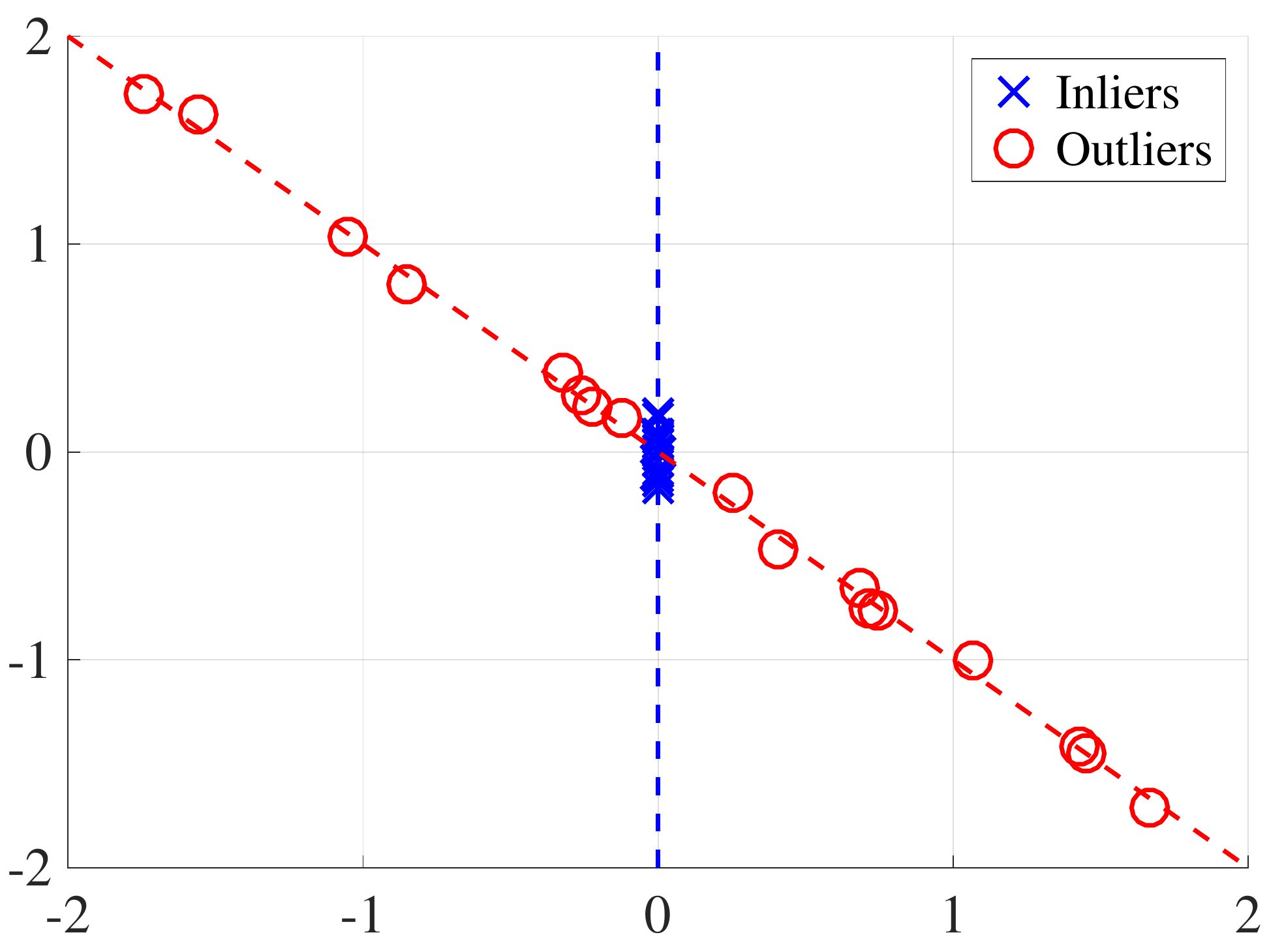}
		\caption{\label{subfig:smallinoutline}  An example where the outliers lie near a line that may describe the whole dataset better than the inliers.}
	\end{subfigure}
	\begin{subfigure}{.47\textwidth}
		\includegraphics[width=\textwidth]{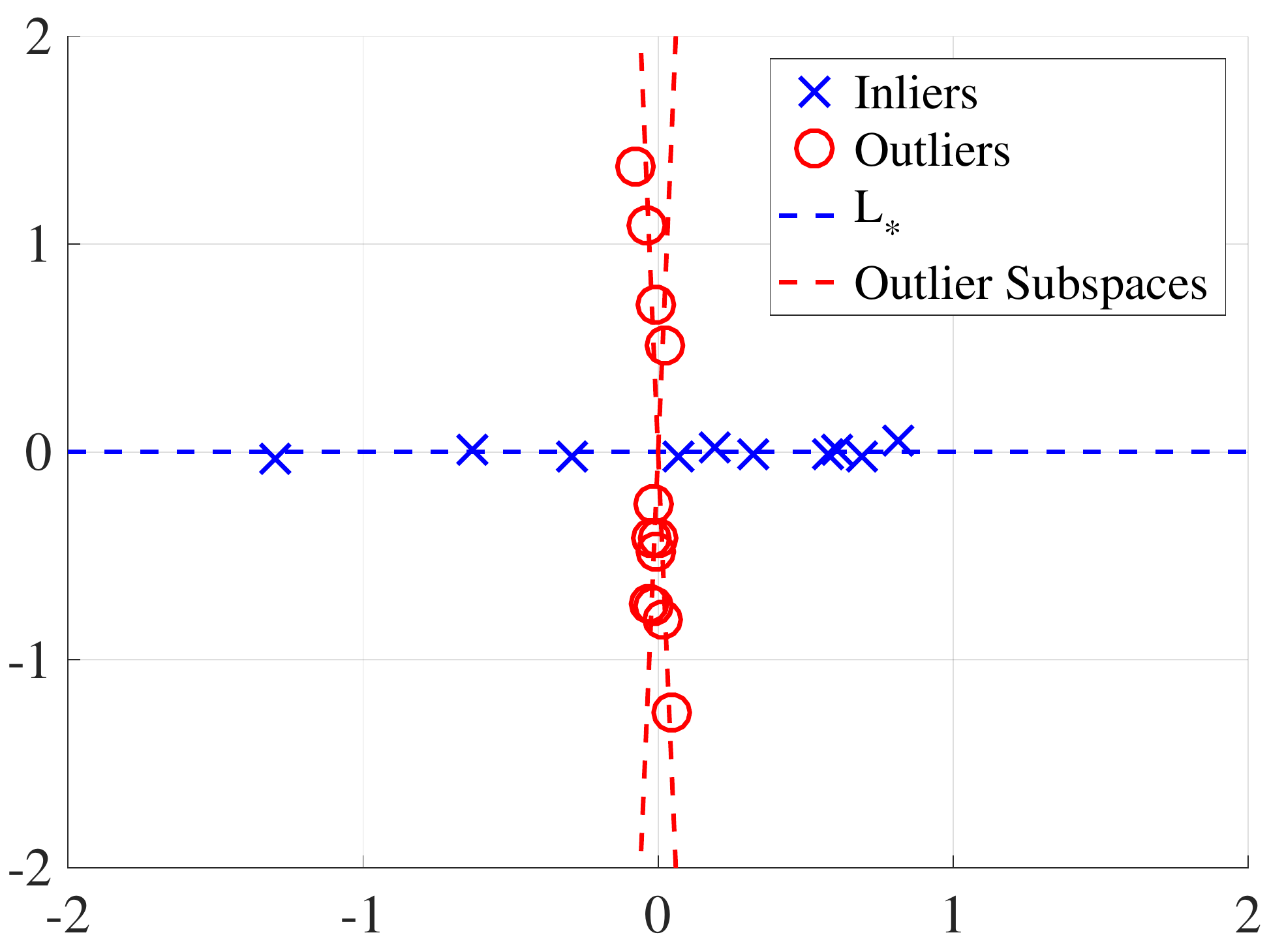}
		\caption{\label{subfig:mssissue} A demonstration of 10 inliers around a line and two lines containing 6 outliers each. For near recovery, the line in between the two outlier lines may better represent the whole dataset.}
	\end{subfigure}
	\begin{subfigure}{.47\textwidth}
		\includegraphics[width=\textwidth]{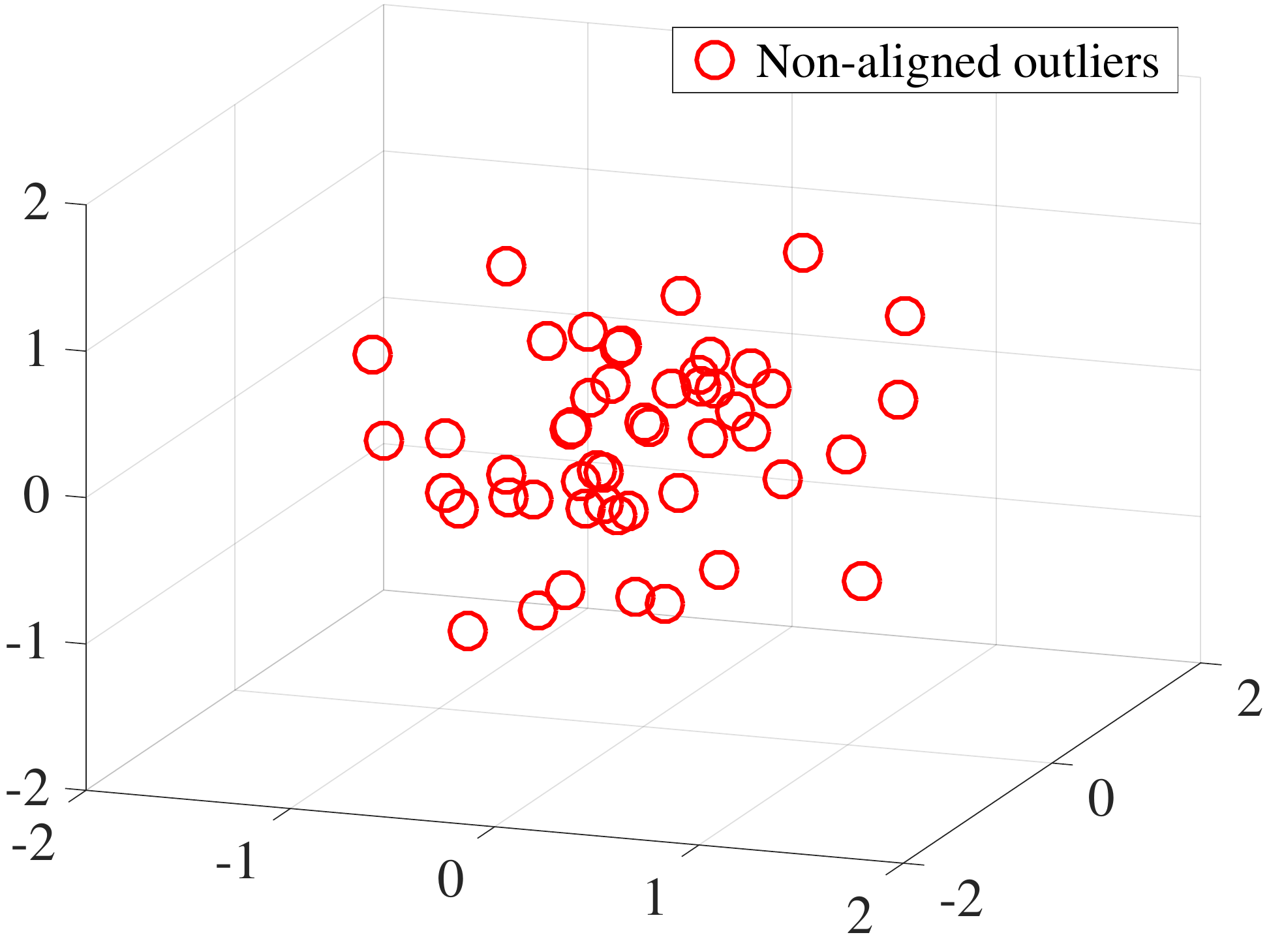}
		\caption{A cartoon of outliers with restricted alignment. As clarified in this section, there is some flexibility in the notion of restricted alignment, and the strong restriction shown here is needed only in some regimes.}\label{subfig:noalignout}
	\end{subfigure}
	\caption{Examples clarifying the two principles that ensure well-defined models.}
\end{figure*}

This section will develop a principle for inlier distributions that ensures the RSR problem is mathematically well-defined.
We start with a somewhat extreme case, where the noiseless RSR  problem is ill-defined. We assume no outliers and inliers lying at the origin, which is demonstrated in Figure~\ref{subfig:inorigin}. In this case, any linear subspace contains the inliers, and it becomes impossible to designate any one subspace as ``underlying''.

Figure~\ref{subfig:lowdimin} illustrates another example where the inliers lie in a lower-dimensional subspace of $L_*$ and the problem is ill-defined. In this example, $L_*$ is a 2-subspace in $\reals^3$, the inliers concentrate on a 1-subspace of $L_*$ and the outliers concentrate on a 2-subspace that intersects $L_*$ at this 1-subspace. The issue here is that the outlier subspace seems more natural for describing the data than the ``underlying'' subspace $L_*$. Indeed, more data points lie in this subspace than in $L_*$. There are two key points that one should take away from this artificial example. First, our setting assumes a fixed parameter $d$, which we have designated as $d=2$ in this example. If instead $d$ was unknown, one could argue that the underlying subspace is the 1-subspace at the intersection of the two 1-subspaces. Second, the issue in this example, and also in some following examples, could be resolved by exchanging the labels of inliers and outliers. However, this avoids the main issue we are trying to illustrate here. We are interested in outlining a well-defined mathematical setting with restrictions on the sets labeled as inliers and outliers. In particular, this example illustrates that some restrictions must be placed on the inlier dataset.

Restrictions on the distribution of outliers in Figure~\ref{subfig:lowdimin} could also make it well-defined. Instead, this section focuses on restrictions on the inliers that make the problem well-defined. We also comment that the notion of a subspace that describes the whole dataset better than $L_*$ is not completely well-defined yet but is somewhat conveyed by this figure. We will discuss this issue more carefully when describing how to restrict outliers in \S\ref{subsec:theoryalign}.

From the previous examples, we see that the inliers cannot be too concentrated around lower dimensional subspaces of $L_*$ and must instead fill out $L_*$ in order to have a mathematically well-defined setting. We refer to this as the principle of \emph{permeance} of the inliers, since the inliers must permeate the underlying subspace. We will later demonstrate how different works formulate this principle in different ways. Figure \ref{subfig:permin} presents a cartoon of permeated inliers when $d=2$ and $D=3$. We remark that non-uniformity of sampling within $L_*$, and possibly some very low level of concentration on low-dimensional subspaces of $L_*$, can be tolerated.

\begin{figure*}
	\centering
	\begin{subfigure}{.47\textwidth}
		\includegraphics[width=\textwidth]{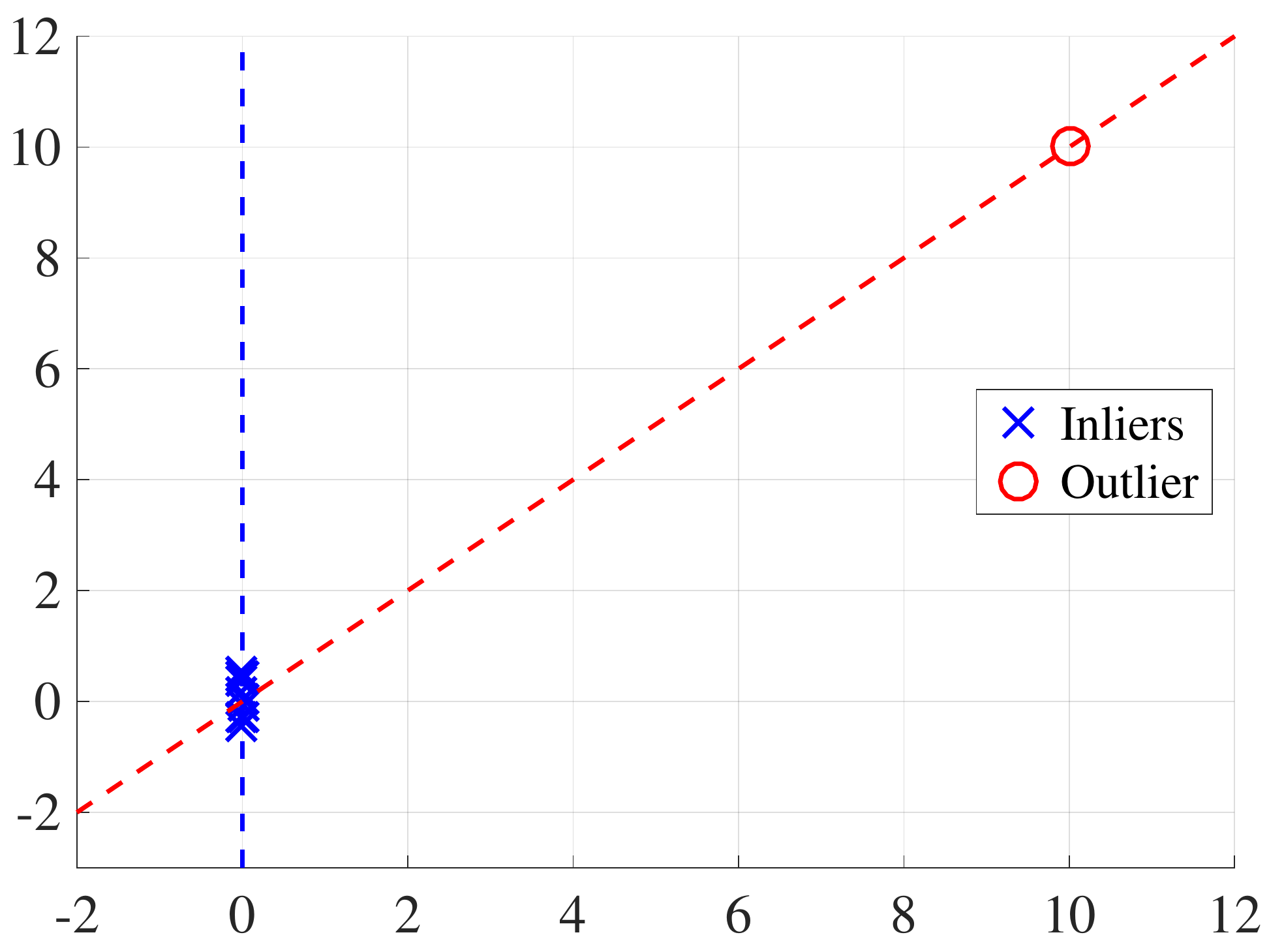}
		\caption{An example of a dataset with inliers on a line and a single outlier of very large magnitude. \label{subfig:onelargeout}}
	\end{subfigure}
	\hfill
	\begin{subfigure}{.47\textwidth}
		\includegraphics[width=\textwidth]{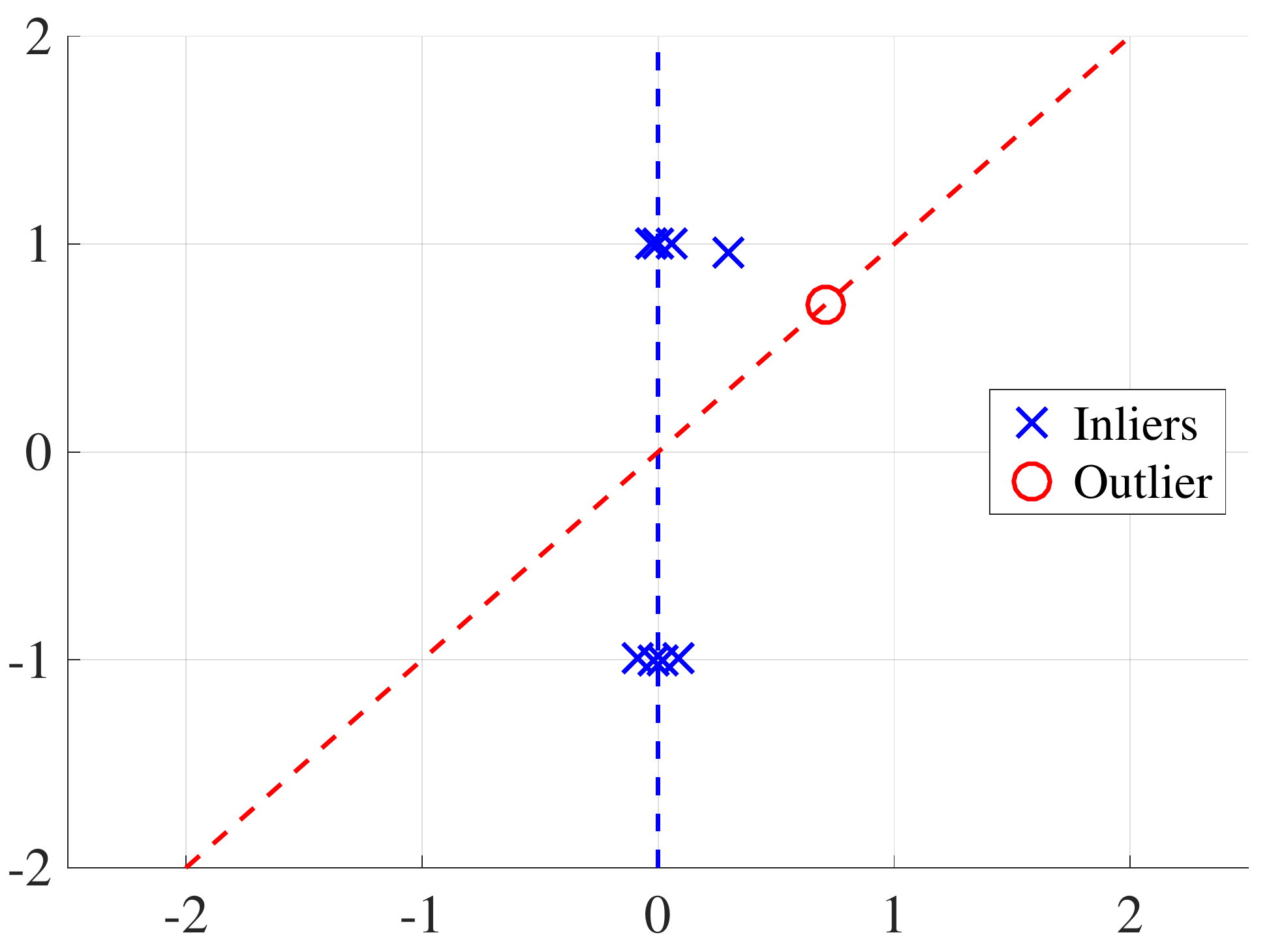}
		\caption{The dataset of Figure~\ref{subfig:onelargeout} mapped onto the unit circle by normalizing each original data point by its Euclidean norm. \label{subfig:onelargeoutspherized}}
	\end{subfigure}
	\begin{subfigure}{.47\textwidth}
		\includegraphics[width=\textwidth]{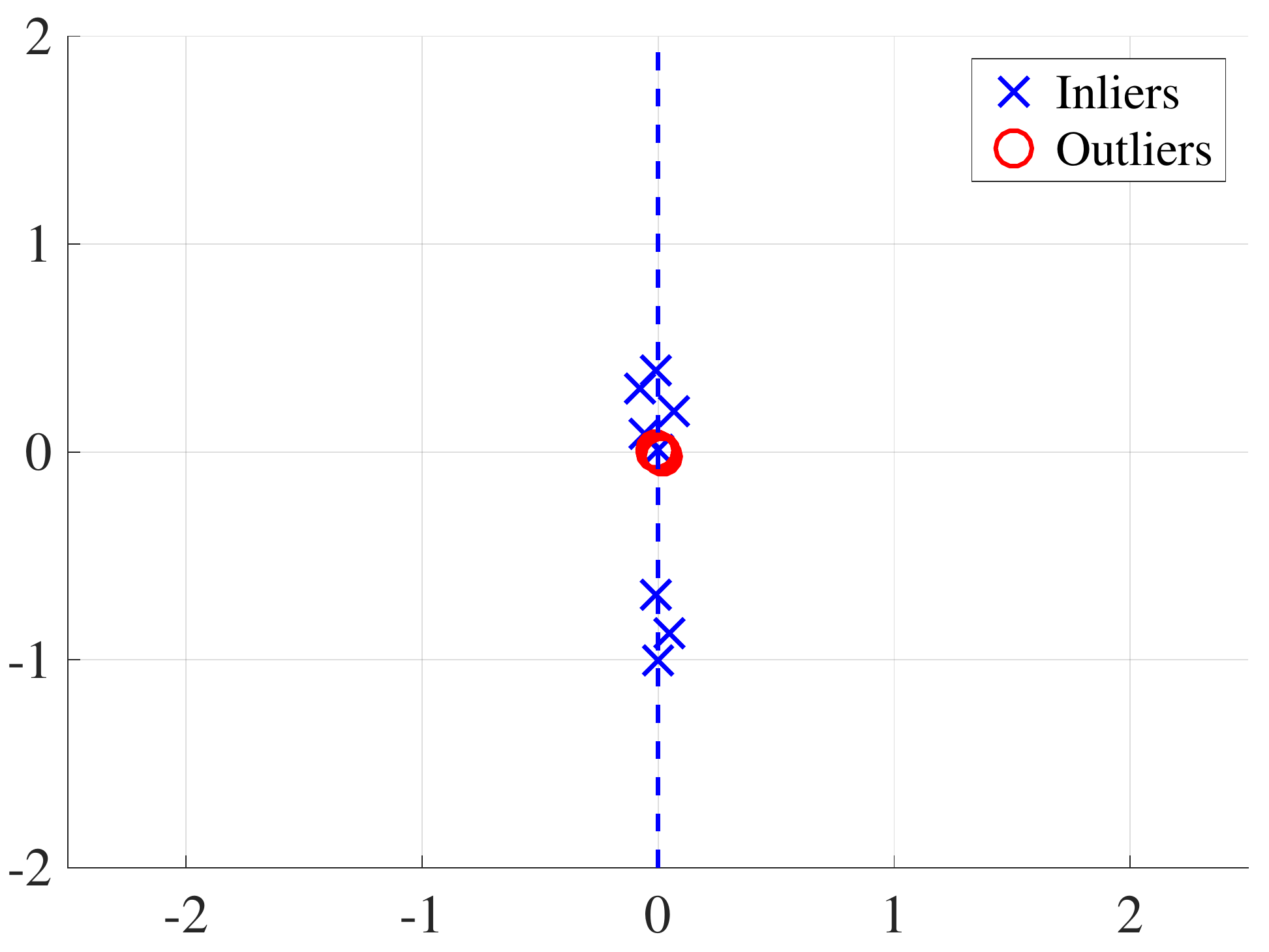}
		\caption{\label{subfig:inlierssmallout} An example of a dataset with outliers close to the origin, but near a different line than $L_*$. This line is unnoticeable since the magnitude of outliers is negligible.}
	\end{subfigure}
	\hfill
	\begin{subfigure}{.47\textwidth}
		\includegraphics[width=\textwidth]{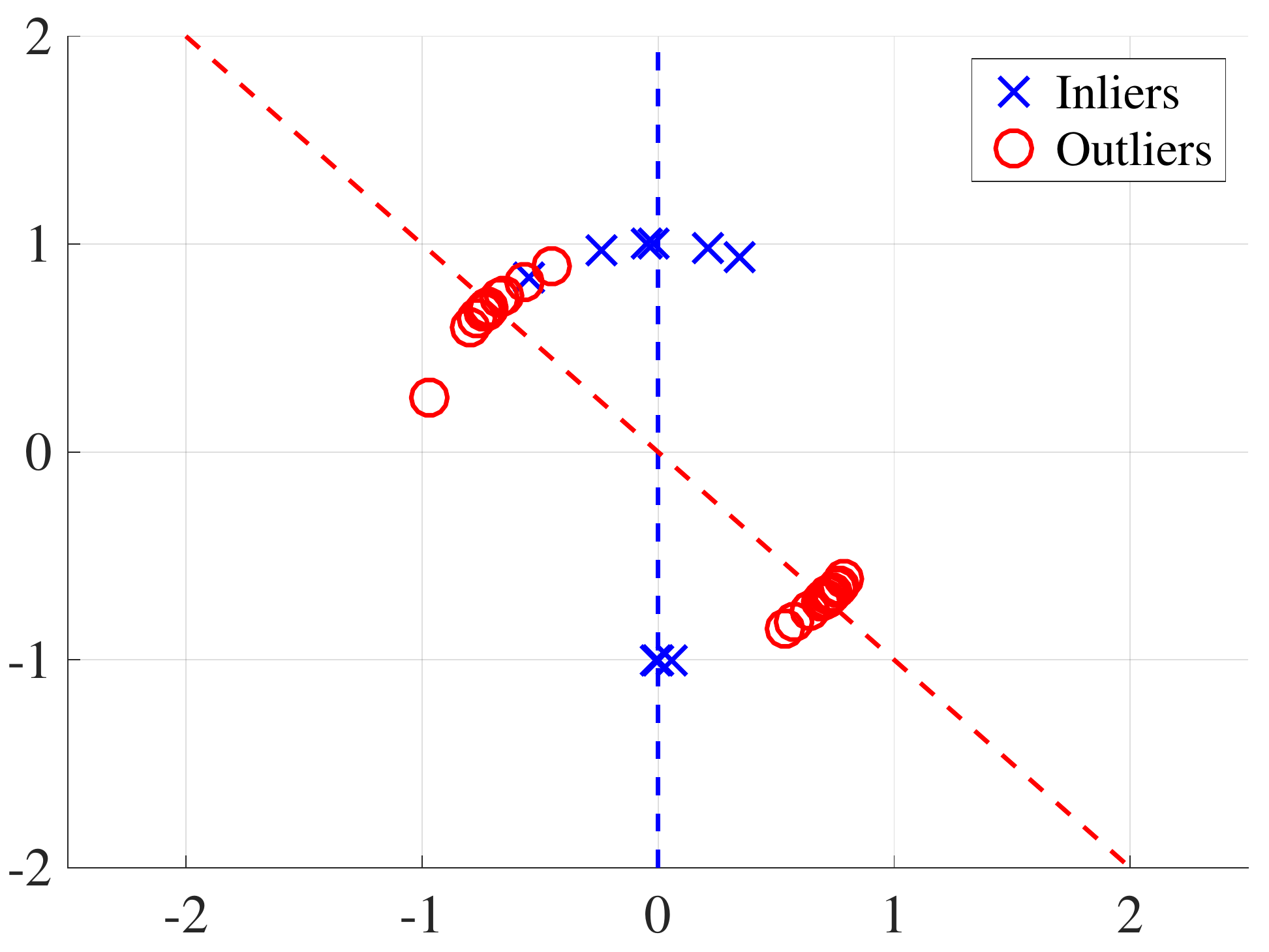}
		\caption{The dataset of Figure~\ref{subfig:inlierssmallout} mapped into the unit circle by normalizing each original data point by its Euclidean norm. \label{subfig:inlierssmalloutspherize}}
	\end{subfigure}
	\caption{{Examples before and after mapping onto the unit circle by normalizing each data point by its Euclidean norm.}}
\end{figure*}

\subsubsection{Restrictions on the Outliers and a Second Principle}
\label{subsec:theoryalign}

In a similar fashion to the previous discussion, some restrictions must also be placed on the outliers to prevent them from giving rise to a subspace that may describe the data better than the underlying subspace, $L_*$. For example, assume that the inliers permeate the underlying subspace to some degree and the outliers have a similar distribution to the inliers on another low-dimensional subspace. A special case of this more general example is demonstrated in Figure~\ref{subfig:smallinoutline}, where both subspaces are one-dimensional. One may claim that the outlier line describes the whole dataset better than the line that contains the inliers. As mentioned earlier, the notion that another subspace may fit the dataset better than the underlying subspace is not yet well-defined. First of all, for the noiseless case, the line $L_*$ may still be more significant, in the sense that it contains more points. If on the other hand, the outliers in Figure~\ref{subfig:smallinoutline} lie exactly on a line and not just near it, one could claim that the outlier line best represents the data. This debate boils down to two issues: 1) whether the number of inliers or outliers is large enough to determine which line represent better the data and 2) whether the larger relative magnitude of outliers contribute to their possible significance.

We start by focusing on the first issue, and we will discuss this second issue a bit later. We assume, in this noiseless version of the example, that the line with largest number of points best represents the data, and we will refer to this line as the ``most significant''.

The notion of most significant subspace is equivalent to the subspace satisfying \eqref{eq:l0}. However, as discussed in Section 1.1. of \cite{lp_recovery_part1_11}, this notion is problematic when the data points are even slightly noisy, where \eqref{eq:l0} needs to be replaced with \eqref{eq:l0nmin}. Figure~\ref{subfig:mssissue} demonstrates such a problem  in a simple case. Here, 10 inliers lie around the horizontal line, and 12 outliers lie around two other lines, each of which contains 6 points. Thus, while each of the outlier lines is less significant in terms of the number of points, the vertical line, which is close to the two outlier lines, has approximately 12 points near it and could be labelled as more significant. To avoid this problem, \citet{lp_recovery_part1_11}, who have a model with several underlying $d$-subspaces, refer to a subspace as ``most significant'' if it contains more points than all other $d$-subspaces combined. We remark, though, that this notion applies to a very specific model and is not well-defined in general.

Assuming that this notion of most significant subspace is well-defined, the RSR problem can also be well-defined if one follows the principle of \emph{restricted alignment} of the outliers. There are different ways of formulating this principle, which affect the nature of the subsequent recovery guarantees. The examples in Figures~\ref{subfig:lowdimin} and~\ref{subfig:mssissue} illustrate that one may need to exclude some sort of concentration of outliers around subspaces of dimensions at most $d$. This way, an outlier subspace cannot be the most significant subspace.

So far, we have ignored the effect of the relative magnitude of the outliers, although this can also influence the resulting conditions.
In some works, restriction on alignment of outliers has to include some control on the ratio between the magnitude of outliers and inliers. If outliers have much larger magnitude than the inliers, they may have undue influence over a robust subspace criterion. Consequently, this sort of magnitude differential can make the problem ill-defined. However, it is possible to use ``scale-invariant'' methods to keep the problem well-posed in cases where there are no restrictions on the relative magnitude of outliers.

We demonstrate this issue with the special case of a dataset containing a single outlier of arbitrarily large magnitude and inliers lying on a one-dimensional underlying subspace in Figure~\ref{subfig:onelargeout}. The line through the large outlier might be viewed as the line that best represents the whole dataset since the distances of all inliers to this line are negligible. On the other hand, this outlier might be perceived as an adversarial one that should be excluded, especially since the rest of data points lie on another line. In this simple case, the outlier can be easily filtered out according to its large magnitude. There are also more general scale-invariant methods that give no weight to the magnitude of the data points, and thus one arbitrarily large outlier has little contribution when applying these methods.

We say that an RSR algorithm is scale-invariant if the output of the algorithm does not change after multiplying all the data points by different non-zero factors. A simple technique that results in scale-invariant algorithms is to initially normalize the data points by their Euclidean norms so that they lie on the sphere, $S^{D-1}$, and then apply any RSR method. Application of this normalization procedure to the simple dataset of Figure~\ref{subfig:onelargeout} is demonstrated in Figure~\ref{subfig:onelargeoutspherized}. We remark that it is unclear how to do this normalization procedure when there is missing data or when the setting is affine instead of linear.

This procedure, as well as other scale-invariant algorithms, may miss some important information in the magnitude of inliers and outliers. The special example in Figures \ref{subfig:inlierssmallout} and \ref{subfig:inlierssmalloutspherize} emphasizes this issue. Here, the outliers have very small magnitudes, and so the whole dataset is well-approximated by a line. However, the small outliers actually lie around a line that is quite different than the inlier line. Normalization of the dataset then emphasizes the outlier line more than the original inlier line. Thus, Figure~\ref{subfig:inlierssmalloutspherize} demonstrates that, even when applying scale-invariant algorithms, the alignment of outliers still has to be restricted, although there is not any consideration of their magnitude.

Employing an exhaustive subspace search method to minimize~\eqref{eq:l0} is also scale-invariant. Indeed, in a well-defined setting, any such method would find the subspace containing most of the points, independently of any scaling of the data points.
Scale-invariant search algorithms can also be developed for noisy RSR by trying to minimize variants of~\eqref{eq:l0nmin}. For example, in this formulation, one can use the angles between data points and the subspace rather than the orthogonal distance, since angles are scale-invariant.

We have discussed at length the restriction of outliers since there is some flexibility in enforcing it.
Using the examples and concepts explained above, we clarify this flexibility.
In the case of some scale-invariant algorithms, bounding the percentage of outliers can be enough to restrict the alignment.
Similarly, in the case of a non-scale-invariant algorithm, it may be sufficient to bound the magnitude and percentage of the outliers.
On the other hand, when considering regimes with high percentages of outliers, outliers cannot concentrate on or around a significant $d$-subspace for any algorithm.
Notice that, following the earlier discussion in this section, this notion must also interact with the inlier permeance.
For example, the inliers in Figure~\ref{subfig:lowdimin} may require stronger assumptions on outlier alignment than the inliers in Figure~\ref{subfig:permin}.
We further discuss this interaction in the next section.
However, in general, the restriction on alignment is often formulated with respect to the outliers alone.
A case with very restricted alignment, which is needed with high percentages of outliers and is especially needed with a non-scale-invariant algorithm, is demonstrated in Figure~\ref{subfig:noalignout}.
Here, no substantial subset of outliers lies near any low-dimensional subspace, and no outliers have exceptionally large magnitude.

\subsubsection{Stability: the Combination of Permeance and Alignment}
\label{subsec:theorystab}

We refer to both the encouragement of permeance of the inliers and restriction of alignment of the outliers as the stability constraint of the model. 
An example of a stability constraint is demonstrated later in \S\ref{subsubsec:conv}. In this example, positive permeance and alignment statistics, $\cP$ and $\cA$ respectively, are formed so that higher values of $\cP$ correspond to more permeated inliers, and lower values of $\cA$ correspond to more restricted alignment of the outliers. A stability statistic is defined by a positive linear combination of $\cP$ and $-\cA$, and the stability constraint is a lower bound on the stability statistic. In the noiseless case, this bound is zero. We note that satisfying this constraint near the lower bound requires some tradeoff between inlier permeance and restricted outlier alignment. Nevertheless, each of the two quantities, $\cP$ or $\cA$, is computed with respect to only the inliers or outliers respectively, and thus the stability constraint does not fully explore the interaction between the configurations of inliers and outliers.

Some stability constraints imply an upper bound on the percentage of outliers, or equivalently, a lower bound on the percentage of inliers.
Borrowing terminology from signal processing, \citet{zhang2014novel} define the signal-to-noise ratio (SNR) of the RSR problem as the ratio of the number of inliers to the number of outliers under a given stability constraint.
For a given theoretical data model, algorithms can be compared by the lowest SNR under which they can still exactly recover the underlying subspace, or nearly recover it up to a certain error.
The next subsection reviews some of these theoretical data models.

\subsubsection{Specific Models of RSR}
\label{subsec:statmod_quantify}

In this section, we explain several models under which lowest SNRs of algorithms can be compared. The first model uses arbitrary outliers. We remark that this model only works with scale-invariant algorithms, since there is no restriction on the magnitude of the outliers, and a single outlier can make non-scale-invariant algorithms ill-posed. Here, the restriction of the alignment of outliers is only enforced by bounding their percentage, and thus the bound on SNR is relatively high. \citet{xu2012robust} claim that in this model the SNR has to be larger than $d$, and there are indeed degenerate examples where the problem is ill-defined when the SNR is $d$. If, on the other hand, one encourages permeance of inliers, then lower SNR can be obtained. More careful study of this model, including guarantees for existing and new algorithms, is needed. The authors plan to address this issue in a forthcoming paper~\cite{ML18}.

Another model is that of inliers and outliers in general position (see two similar formal definitions in \S\ref{subsec:robcovtheory} and \S\ref{subsubsec:exhaust}). As explained later, \citet{hardt2013algorithms} show that in some sense the optimal SNR in this model is $d/(D-d)$. This is much lower than the case of arbitrary outliers since the outliers exhibit no linear dependencies.
If the SNR is bounded from below by this optimal value, then \citet{hardt2013algorithms} reduce the noiseless RSR problem to finding a linearly dependent $D$-subset, which is not hard.

Only scale-invariant algorithms can have guarantees for the general position model, because again there is no restriction on the magnitudes of the inliers and outliers. However, there are three main drawbacks regarding the applicability of this model.
First, in some real datasets, such as ones involving face images under different illuminating conditions or hand-written digit images (see some relevant discussion in \S\ref{sec:exp}), subgroups of outliers may lie within low-dimensional subspaces. Therefore, the general position model may not be relevant to some real datasets.
Second, this model is well-formulated for exact recovery in the noiseless case and does not seem to easily extend to the noisy setting of near recovery. While~\citet{hardt2013algorithms} propose using the threshold $\det(\tilde{\bX}^T \tilde{\bX} ) < \delta$ in the noisy case, where $\tilde{\bX}$ is the subsampled dataset, it is not at all clear when this would work. For example, this determinant would be small if one of the points in $\tilde{\bX}$ had very small entries, even if $\tilde{\bX}$ did not contain more than $d$ inliers. It is also not clear how to set the threshold $\delta$ even for simple statistical models of noise, such as white Gaussian noise. Third, it is hard to determine how well many of the scale-invariant algorithms behave on the general position model. The only algorithms with results for this model are RF~\citep{hardt2013algorithms} and TME~\citep{zhang2016robust}.

Many times, the analysis of RSR methods lends itself to considering certain statistical models of generating data. We believe studying such statistical models is important because it gives more insight into the performance of algorithms than just the worst case scenario in theorems with arbitrary outliers. Indeed, this sort of average case analysis illuminates differences in the breakdown of algorithms in low SNR regimes. For example, the haystack model~\citep{lerman2015robust} has been used to compare the theoretical guarantees of the various algorithms. The haystack model is a simple model for RSR data, where inliers and outliers both follow Gaussian distributions. In this model, inliers are symmetrically distributed on the underlying subspace with distribution $N(\bzero,\sigma_{\mathrm{in}}^2 \bP_{L_*} / d)$, while outliers have an isotropic Gaussian distribution in the ambient space, given by $N(\bzero,\sigma_{\mathrm{out}}^2 \bI / D ))$. However, this model is limited since it captures a very particular scenario. The generalized haystack model~\cite{maunu2017well}, in which outliers have a general and possibly degenerate covariance and inliers have a general covariance restricted to the subspace, captures more diverse scenarios, but the model is still quite specialized.

Theoretical results so far have emphasized exact recovery of subspaces in the noiseless RSR setting under the models discussed above. They often discuss extension of the results to near recovery with small amount of noise.
Only a few existing works have focused on the truly noisy setting~\citep{coudron2012sample,minsker2015geometric,cherapanamjeri2017thresholding}.

\subsection{Sequential Methods and Projection Pursuit}
\label{sec:sequential}

A simple strategy for RSR is to fit one-dimensional directions sequentially. This strategy has been pursued in various lines of work, such as the projection pursuit method we discussed in~\eqref{eq:projpur} and~\eqref{eq:projpur2}.
However, there is no guarantee that a sequential method will recover a stationary point of an energy for $d$-subspace recovery. For example, for projection pursuit, such an energy is given by $\sum_{j=1}^d \rho(\bv_j^T\bX)$ over the set of orthonormal systems $\bv_1,\dots, \bv_d \in \reals^D$~\citep{mccoy2011two}. In the PCA problem formulation, one can show that joint estimation and sequential estimation of principal components result in the same subspace. However, for other energies, joint and sequential estimation do not result in the same subspace. Also, the nonconvexity of the problem has caused works to guarantee convergence to local optima in each individual subproblem (formulated in \eqref{eq:projpur} and \eqref{eq:projpur2})~\citep{kwak08} or convergence to a weak approximation of the global optimum of the joint energy $\sum_{j=1}^d \rho(\bv_j^T\bX)$~\citep{mccoy2011two,naor2013efficient}.

One shortcoming of sequential methods is the potential for compounding errors due to noise. Suppose we have a noisy data matrix $\bX$, and we find a top component $\bv_1$. Then, one can try to run the same algorithm again on the data matrix $\bQ_{\bv_1}\bX$. However, due to noise, if we expect an optimal recovery error of approximately $\epsilon$ when estimating $\bV_*$, then $\bv_1$ should be $\Omega(\epsilon)$ from the underlying subspace. After projection and running again, the next component could be, at worst, $\Omega(2\epsilon)$ from the underlying subspace, and so on.  To recover a $d$-dimensional subspace, their errors may accumulate to $\Omega(d\epsilon)$. In methods where one tries to find the orthogonal complement of the underlying subspace, such as \citep{tsakiris2015dual}, errors may even accumulate to $\Omega((D-d)\epsilon)$ if one tries to sequentially fit hyperplanes.

	Further, even in the noiseless case, the first sequential component may be far from the underlying subspace. For example, this is a feature of the least absolute deviations energy. If one has a subspace of dimension $d>1$ with points well distributed on the subspace, then one can mathematically show that the minimizer of~\eqref{eq:lad} over $G(D,1)$ will not be contained in the underlying subspace in general inlier-outlier settings.

We believe that projection pursuit methods generally suffer from the deficiencies present in sequential estimation. Overall, projection pursuit methods have lacked theoretical guarantees and have instead used heuristic arguments to justify them. We are unaware of substantial theoretical work on robust subspace recovery in this area.

\subsection{Least Absolute Deviations}

Most theoretical guarantees for RSR exist for methods aiming to minimize the least absolute deviations. We review them according to the different methods they are associated with.

\subsubsection{Guarantees for Outlier Pursuit}
\label{subsec:optheory}

\citet{xu2012robust} provide theoretical guarantee for recovery by OP, which is the program outlined in~\eqref{eq:op}. The permeance of inliers discussed in \S\ref{subsec:statmod} is quantified by the inverse of an incoherence parameter. This parameter appears in other works on nuclear norm minimization, such as RPCA and matrix completion~\citep{candes2011robust,chandrasekaran2011rank,candes2009exact,candes2010power}. The notion of incoherence and its parameter $\mu$ are defined for the low-rank inlier matrix $\bL$ as follows:
\begin{definition}
    A rank $d$ matrix $\bL$ with $(1-\alpha)N$ non-zero columns, for $\alpha \in (0,1)$, and with SVD $\bL = \bU \bS \bV^T$, is said to be $\mu$-incoherent if
	\begin{equation}
	\max_i \|\bV^T \be_i \|^2 \leq \frac{\mu d}{(1-\alpha)N}.
	\end{equation}
	Here, $\be_i$ are the unit coordinate vectors.
\end{definition}
In the special case of generating the inliers from a spherically symmetric Gaussian distribution within the underlying $d$-subspace, the incoherence parameter is $\mu = O(\max(1, \log(N)/d))$~\cite{candes2009exact}.

We note that the parameter $1-\alpha$ is the fraction of inliers, which are represented by non-zero columns in $\bL$, so $\alpha$ is the fraction of outliers and the SNR is $(1-\alpha)/\alpha$. \citet{xu2012robust} provided the following lower bound on the SNR for exact recovery by outlier pursuit:
\begin{theorem}[\citet{xu2012robust}]
    Suppose the data matrix $\bX \in \reals^{D \times N}$ can be represented as $\bX = \bL + \bC$, where $\bL$ has rank $d$ and incoherence parameter $\mu$, $\bC$ is column sparse and supported on at most $\alpha N$ columns that are not in the column space of $\bL$, and $\lambda = 3/(7\sqrt{N_{\mathrm{out}}})$. Then, if
\begin{equation}
\label{eq:snr_op}
SNR \geq \frac{121 \mu  d}{9},
\end{equation}
outlier pursuit recovers the matrices $\bL$ and $\bC$.

Suppose on the other hand that $\bX = \bL + \bC + \bN$, where $\bL$ and $\bC$ are as above, with $SNR \geq {1024 \mu d}/{9}$, and $\bN$, the noise matrix, satisfies $\|\bN\|_F \leq \epsilon$, then the output $\tilde{\bL}$ and $\tilde{\bC}$ of outlier pursuit satisfy $\|\tilde{\bL}-\bL'\|_F \leq 20 \sqrt{N} \epsilon$ and
$\|\tilde{\bC}-\bC'\|_F \leq 18 \sqrt{N} \epsilon$, {where $\bL' + \bC' = \bL + \bC$}, $\bL'$ has the same column space as $\bL$ and $\bC'$ has the same column support as $\bC$.
\end{theorem}
Nevertheless, we remark that this theory is quite weak for the following reasons.  First, the SNR for arbitrary outliers and permeated inliers is relatively weak (see~\citep{ML18}). Furthermore, it is unclear how to obtain lower SNR for other scenarios with more restriction on the alignment of outliers, where exact recovery can be obtained with significantly lower SNR (see for example Table~\ref{tab:SNR}). Finally, the bounds of near recovery for noise are relatively large.

In general, algorithms aiming to minimize~\eqref{eq:lad} are sensitive to even a single outlier with very large magnitude (without modifications such as normalization of data points to the sphere). However, since the nuclear norm is a very crude approximation of the rank, the contribution of an outlier, or more precisely, its component orthogonal to the underlying subspace, is similar to both parts of the cost function: $\| \bL \|_*$ and $\| \bC \|_{1,2}$. Since the constant $\lambda$ of the cost function is often very small, the outlier column is included in $\bC$ and not $\bL$.
Outlier pursuit is thus scale-invariant for sufficiently large SNR.

\subsubsection{Guarantees for GMS and REAPER}
\label{subsubsec:conv}

\citet{zhang2014novel} consider the development of deterministic stability conditions that ensure subspace recovery by GMS, whose estimator was defined in~\eqref{eq:gms}. They also discuss the types of outliers that can make subspace recovery hard and provide visualizations of these (see Figure 1 in~\citep{zhang2014novel}). They then show that the deterministic stability condition holds under certain sub-Gaussian inlier-outlier mixture models as well as the haystack model with overwhelming probability. By introducing a perturbation argument, they extend their results to near recovery when the inliers lie near a subspace.
Their restriction on the alignment of outliers is very strong, and, in practice, they require at least $1.5D$ outliers filling out the ambient space.
If this condition is not satisfied, then GMS does not have good accuracy. \citet{zhang2014novel} provide three solutions to this, although it is not clear how well these would perform in general. In our numerical experiments in \S\ref{sec:exp}, we test their solution of adding 1.5$D$ spherically symmetric Gaussian outliers in the ambient space.

The work of~\cite{lerman2015robust} on the REAPER algorithm, which uses the estimator given in~\eqref{eq:reaper}, also gives a deterministic recovery result when a dataset satisfies a stability criterion.
They define the permeance statistic $\cP(L_*)$ of a dataset on the underlying subspace $L_*$ as a measure of the notion of permeance of the inliers projected onto the subspace $L_*$.
Note that this definition assumes inliers possibly near the underlying subspace and that is why they project them onto the subspace. They also define the alignment statistic $\cA(L_*)$ that quantifies the restriction of the alignment of outliers. The definition of $\cP(L_*)$ and $\cA(L_*)$ appear in equation (2.1) and (2.3) of~\cite{lerman2015robust}. 
The stability statistic, $\cS$, is defined as
\begin{equation}\label{eq:stab}
\cS(L_*) = \frac{\cP(L_*)}{4 \sqrt{d}} - \cA (L_*).
\end{equation}

In the noiseless case, their theory implies that positive stability at the underlying subspace $L_*$ guarantees exact recovery of this subspace by REAPER.
Their theory also provides a probabilistic lower bound on the stability statistic under the haystack model. This implies exact recovery with overwhelming probability under the SNR indicated in Table~\ref{tab:SNR}.

In the general case of RSR, $\cS(L_*)$ needs to be larger than what they call the total inlier residual with respect to $L_*$, which is defined by
\begin{equation} \label{eqn:residual}
\cR(L_*) := \sum_{\bx_i \in \cX_{\mathrm{in}}}
	\| \bP_{L_*^\perp} \bx_i \|.
\end{equation}
When this condition is satisfied the REAPER solution approximates well the underlying subspace $L_*$ in the following way.
\begin{theorem}[\citet{lerman2015robust}] \label{thm:rrp-stable}
    Suppose $\cX$ is a general RSR dataset in $\reals^D$ with an underlying $d$-dimensional subspace $L_*$, $\tilde{\bP}$ is
    a solution to the REAPER problem~\eqref{eq:reaper}, and $\tilde{\bPi}= \bU_d \bU_d^T$, where $\bU_d \in \reals^{D \times d}$ is the matrix whose columns are the top $d$ eigenvectors of $\tilde{\bP}$.
Then,
\begin{equation} \label{eq:reapbd}
    \|{ \tilde{\bPi} - \bP_{L_*} }\|_*
    \leq \frac{4 \, \cR(L_*)}{\max( \cS(L_*) - \cR(L_*) , 0)}.
\end{equation}
\end{theorem}
Notice that the fraction in~\eqref{eq:reapbd} is only meaningful when $\cS(L_*) > \cR(L_*)$.

There is also an interesting noise-robustness analysis for the GMS and REAPER algorithms that is given in~\citep{coudron2012sample}. Here, the authors prove that the sample complexity of these algorithms is approximately the same order as that of the sample covariance for sub-Gaussian distributions. This observation implies nontrivial robustness to noise.

\subsubsection{Guarantees for Nonconvex Formulations of Least Absolute Deviations}
\label{subsubsec:nonconv}

We discuss existing theoretical guarantees or the lack thereof for the following nonconvex least absolute deviation methods according to this order: R1PCA, the pure energy minimization in~\eqref{eq:lad}, FMS, GGD, TORP, and DPCP. These methods were laid out in \S\ref{subsec:nonconvlad}.

For general datasets, convergence for all of the following algorithms is proven to a stationary point at best.
Furthermore, we do not know in general whether or not this stationary point recovers something useful.
Because of this, some works have resorted to further restrictions on the data. These restrictions are used to show when the algorithms converge to an underlying subspace and also to show the speed of convergence.

The work of~\citet{Ding+06} on R1PCA was originally claimed to be convex, but they actually optimize a nonconvex problem formulation. Thus, they do not have guarantees of global optimality for their minimization and no guarantees of subspace recovery.

\citet{lp_recovery_part1_11} prove exact subspace recovery w.o.p.~by minimization of the least absolute deviations energy~\eqref{eq:lad} under a certain probabilistic model of data.
The datasets considered involve a mixture model with i.i.d.~inliers distributed uniformly on $S^{D-1} \cap L_1^*$ and i.i.d.~outliers distributed uniformly on $S^{D-1}$ and the intersection of $S^{D-1}$ with $K-1$ subspaces $L_2^*,\dots,L_K^*$. It is further assumed that the asymptotic fraction of points on $L_1^*$ is greater than the asymptotic fraction of points on $L_2^*,\dots,L_K^*$ combined.
This work shows the least absolute deviations energy can handle any fixed fraction of i.i.d.~outliers distributed uniformly on $S^{D-1}$.
 However, this work only focuses on analysis of the pure minimization problem and not of an algorithm for minimizing it. Furthermore, its model is restrictive, and its estimates require large sample sizes.

 \citet{lerman2017fast} provide some guarantees for the FMS algorithm, although they are somewhat limited. We remind the reader that the FMS procedure tries to directly minimize~\eqref{eq:lad} using iteratively reweighted least squares. They prove that the FMS algorithm converges to a stationary point in general and is able to decrease the least absolute deviations energy monotonically from its starting point. However, they do not guarantee that this stationary point is a local minimum in general settings. They further show that the FMS algorithm can nearly recover an underlying subspace in two special settings: 1) when outliers are spherically symmetric and inliers are spherically symmetric within the underlying subspace or 2) outliers are spherically symmetric or lie on a one-dimensional less significant subspace, and inliers lie on a significant one-dimensional subspace. In the first setting, the analysis shows that FMS can nearly recover the underlying subspace for any fixed fraction of outliers (less than 1). For both settings the convergence of FMS is locally $r$-linear. Nevertheless, the estimates in \cite{lerman2017fast} require large sample sizes.

\citet{maunu2017well} formulate a deterministic stability condition that guarantees nice behavior of the energy landscape of~\eqref{eq:lad} in a local neighborhood around $L_*$ (more details are described below). They also show that under this stability condition, a geodesic gradient descent (GGD) algorithm for~\eqref{eq:lad} initialized in this neighborhood exactly recovers the underlying subspace. They further show that a similar deterministic stability condition ensures that the PCA $d$-subspace lies in this neighborhood.
Therefore, GGD initialized by PCA has an exact recovery guarantee under both stability conditions simultaneously.

The stability condition was inspired by the previous ideas of~\citep{lerman2015robust} and focuses again on a difference of two statistics: an inlier permeance and outlier alignment. For simplicity, we discuss here only the noiselss case.
The permeance and alignment statistics can be seen in (9) and (10) of~\citep{maunu2017well}. Since the condition is local, a parameter $0 < \gamma < \pi/2$ determines how large of a neighborhood is considered.
This neighborhood is defined in the following way:
    \begin{equation}
        B(L_*, \gamma) = \{ L \in G(D,d) : \theta_1(L,L_*) < \gamma \}.
    \end{equation}
    Here, $\theta_1(L_1, L_2)$ is the largest principal angle between two subspaces $L_1$ and $L_2$.
Using the bounds given in~\citep{maunu2017well}, it is easier to interpret the following lower bound on the stability statistic:
\begin{align}\label{eq:stabbd}
    \cS(\gamma,L_*) \geq \cos(\gamma) \lambda_d \left( \sum_{\bx_i \in \cX_{\mathrm{in}}} \frac{\bx_i \bx_i^T}{\|\bx_i\|} \right)  - \sqrt{N_{\mathrm{out}}} \| \bX_{\mathrm{out}} \|_2.
\end{align}
Here, $\lambda_d(\cdot)$ is the $d$th eigenvalue of the input matrix.
The first term measures how well the inliers ``fill out'' the underlying subspace, while the second term measures how aligned the outliers are in any direction.

The stability condition for the noiseless case is positivity of this statistic. The theory outlined earlier can be precisely formulated as follows.
\begin{theorem}[\citet{maunu2017well}]\label{thm:landscape}
	Suppose that an inliers-outliers dataset 
with an underlying subspace $L_*$ satisfies $\cS(\gamma,L_*) > 0$, for some $0<\gamma < \pi/2$. Then,
 all points in ${B(L_*,\gamma)}\setminus \{L_*\}$ have a directional subdifferential strictly less than $-\cS(\gamma,L_*)$, that is, it is a direction of decreasing cost. This implies that $L_*$ is the only local minimizer in ${B(L_*,\gamma)}$. Suppose further that the initial GGD iterate is $L_1 \in {B(L_*,\gamma)}$. Then, for sufficiently small $s$, GGD with step size $s/\sqrt{k}$ converges to $L_*$ with rate $\theta_1(L_k,L_*) < O(1/\sqrt{k})$.

 {
 	Under an additional ``strong gradient condition" specified in (21) of~\citep{maunu2017well}, for sufficiently small $s$ and sufficiently large $K$, GGD with step size $t^k = s / 2^{\lfloor k/K \rfloor}$ linearly converges to $L_*$.
}
\end{theorem}

Initialization in this neighborhood is guaranteed by the following lemma, which is a consequence of the Davis-Kahan $\sin \Theta$ Theorem~\citep{Davis1970}.
\begin{lemma}
Suppose that, for a noiseless inliers-outliers dataset,
\begin{equation}
    \sin(\gamma) \lambda_d(\bX_{\mathrm{in}} \bX_{\mathrm{in}}^T) > \|\bX_{\mathrm{out}}\|_2^2.
\end{equation}
Then, the PCA $d$-subspace is in ${B(L_*,\gamma)}$.
\end{lemma}
The stability condition is shown to hold with overwhelming probability under a variety of models of data, and it is also shown to be stable with small noise. In particular, GGD is shown to have recovery guarantees almost on par with the strongest convex methods on the haystack model discussed later in \S\ref{subsec:haystack}. The downside for GGD is that it requires slightly larger sample estimates: $N=O(D^2\log(D))$ versus $N=O(D)$ for convex methods like REAPER and GMS. GGD also has a guarantee of recovery for any fixed percentage of outliers under this model in the large sample limit (when one allows $N \to \infty$).

\citet{cherapanamjeri2017thresholding} give theoretical guarantees for TORP with arbitrary outliers and noise, when the fraction of outliers is known. The authors prove that the algorithm works with arbitrary corruptions up to an SNR of order $\Omega(d)$, although the constants are quite poor.
\begin{theorem}[\citet{cherapanamjeri2017thresholding}]\label{thm:TORP}
Suppose the data matrix $\bX \in \reals^{D \times N}$ can be represented as $\bX = \bL + \bC$, where $\bL$ has rank $d$ and incoherence parameter $\mu$, $\bC$ is supported on at most $\gamma N$ columns that are not in the column space of $\bL$, where $\gamma$ is an input parameter for TORP. Then, if
\begin{equation}\label{eq:snr_torp}
    SNR \equiv \frac{1-\gamma}{\gamma} \geq 128 \mu^2 d - 1,
\end{equation}
the TORP algorithm linearly converges to a point that exactly recovers the column space of $\bL$.

Suppose on the other hand that $\bX = \bL + \bC + \bN$, where $\bL$ and $\bC$ are as above and $\bN$ is added noise. Then, the TORP algorithm linearly converges to a subspace $\bU$ such that $\|(\bI - \bU \bU^T) \bL \|_F \leq 60 \sqrt{d} \| \bN \|_F$.
Under the more restrictive assumptions that $\bN$ has entries i.i.d.~$N(0,\sigma^2)$ and $SNR \geq 1024 \mu^2 d - 1$, TORP linearly converges to a subspace $\bU$ such that  $\|(\bI - \bU \bU^T) \bL \|_F \leq 4 \sqrt{\log(d)} \| \bN \|_2$ w.o.p.
\end{theorem}
Since results are only proven for arbitrary corruptions, the bounds for certain generative models of data (such as the haystack model) are weaker than those given in~\citep{lerman2015robust,maunu2017well}. We note that TORP linearly converges to the solution in all of the restricted settings in Theorem~\ref{thm:TORP}. The authors also have an analysis to noise that is similar to that in~\citep{coudron2012sample}. They show that the sample complexity is similar to that of PCA on the noisy inlier distribution.

DPCP~\citep{tsakiris2015dual}, which solves the program in~\eqref{eq:dpcp}, is able to prove recovery of subspace structures under some deterministic conditions by finding a sequence of nested hyperplanes. However, the conditions are quite hard to interpret, especially when one is finding nested structures. It is even hard to calculate what the conditions mean for a given statistical model of data, such as the haystack model.

\subsection{$L_1$-PCA}

We are currently not aware of any recovery or robustness guarantees for $L_1$-PCA, which was outlined in~\eqref{eq:l1min} and~\eqref{eq:l1max}.
Recovery guarantees for the RPCA problem in \eqref{eq:rpca1}, which is similar to the $L_1$-PCA problem of~\eqref{eq:l1min}, are reviewed in \cite{vaswani2018static}.

\subsection{Robust Covariance Estimation}
\label{subsec:robcovtheory}

For quantification of the robustness of covariance estimators, the study of breakdown points has been important~\citep{lopuha_rousseeuw_robust91}. Essentially, the robust covariances are consistent estimators of covariance matrices for elliptical distributions with nontrivial breakdown points. This means they can tolerate some percentage of arbitrary outliers and still estimate the underlying elliptical covariance well, which, in turn, means they could be able to estimate an underlying principal subspace well. However, the study of this principal subspace for RSR is only analyzed in~\citep{zhang2016robust}.

These sorts of breakdown points hold for the estimation of
covariances since the space of these matrices is non-compact and there is a notion of a covariance matrix with arbitrarily large magnitude.
On the other hand, a similar
definition of a breakdown point does not hold for subspace recovery since the Grassmannian is compact.  The notion of lowest SNR allowing exact subspace recovery or sufficiently near recovery is clearly weaker.

\citet{zhang2016robust} demonstrated that TME can also be used for subspace recovery. The stability condition in~\citep{zhang2016robust} requires a lower bound on the SNR as well as general positions of both inliers and outliers.
We say that the inliers are in general position with respect to $L_*$ if,
any $d$ of them are linearly independent. Similarly, we say that the outliers are in
general position with respect to $L_*^\perp$ if, after projecting them onto $L_*$ any $D-d$ of them are linearly independent.
Using this definition, the theorem is formulated as follows:
\begin{theorem}[\citet{zhang2016robust}]
    Assume that $\cX \subset \reals^D$ is a noiseless inliers-outliers dataset in $\reals^D$ with an underlying $d$-subspace $L_*$. If the inliers are in general position with respect to $L_*$, outliers are in general position with respect to $L_*^\perp$, and $SNR > d/(D-d)$, then TME exactly recovers $L_*$.
\end{theorem}
The theorem that extends subspace recovery by TME to noisy datasets is quite weak and hard to interpret, and so we do not state it here (see Theorem 3.1 in~\citep{zhang2016robust}).
We remark that a clear advantage of TME is that it is scale-invariant. Indeed, it is obvious from~\eqref{eq:tme} that scaling any data point by an arbitrary non-zero constant will not affect the estimator. Spherical PCA is also scale-invariant, as can be seen from~\eqref{eq:spherecov}, but it is not able to exactly recover subspaces like TME.

\subsection{Other Energy Minimization and Filtering Outliers}
\label{subsubsec:outremove}

The outlier removal energy of~\citet{Xu1995} does not have any associated guarantees for subspace recovery.

The bounds for the performance of HR-PCA and DHR-PCA are hard to interpret for arbitrary datasets~\citep{xu2013outlier,Feng:12}. The authors chose to focus on a quantity called expressed variance (EV). Suppose one would like to measure the quality of an orthogonal basis matrix $\bU \in O(D,d)$ against the optimal subspace, represented by $\bU_* \in O(D,d)$. Then, the expressed variance is given by
\begin{equation}
EV(\bU) = \frac{ \|\bU^T \bX_{\mathrm{in}}\|_F^2 }{\|\bU_{*}^T \bX_{\mathrm{in}}\|_F^2}.
\label{}
\end{equation}
The expressed variance takes values between 0 and 1, and it measures the proportion of underlying variance captured by the basis.  \citet{xu2013outlier} prove lower bounds on the expressed variance, although these bounds are quite weak. For example, in the case of spherically symmetric Gaussian inliers on a subspace and spherically symmetric Gaussian outliers, their lower bound on EV is 0.09~\citep{lerman2017fast} (while an EV of 1 amounts to exact recovery).

\citet{ariascastro2011spectral} guaranteed their method for removing outliers in the setting of robust manifold clustering. In the case of robust recovery of a single manifold, their theorem implies exact identification of outliers when the inliers are uniformly sampled from a $\tau$-neighborhood in $[0,1]^D$ of a certain $C^2$ submanifold of $[0,1]^D$, the outliers are uniformly sampled from the complement of that $\tau$-neighborhood in $[0,1]^D$ and the SNR is of order $\Omega(\max(\log(N) \cdot N^{-2(D-d)/(2D-d)}, \tau^{D-d}))$. Here, the $\tau$ neighborhood contains all points that have distance less than or equal to $\tau$ with the submanifold.

\citet{soltanolkotabi2011} guaranteed their method for removing outliers, which is similar in spirit to \cite{chen2009spectral,ariascastro2011spectral}, in the setting of noiseless robust subspace clustering. In the case of a single subspace recovery, their theorem implies exact removal of outliers when the outliers are uniform in $S^{D-1}$, the $N_{\text{in}}$ inliers are uniform on the intersection of $S^{D-1}$ with a uniformly random $d$-subspace and the SNR is of order $\Omega(\frac{d}{D}\cdot((\frac{N_{\text{in}}-1}{d})^ {\frac{cD}{d}-1}-1)^{-1})$, as long as $N<e^{c\sqrt{D}}/D$. In this method and the below work by~\citet{rahmani2017coherence}, the authors need to assume that the inlier subspace is unformly random, which is a stronger assumption than other methods make.

\citet{YouRV17} proved exact recovery of outliers in the noiseless setting of robust subspace recovery under certain conditions. They did not verify that these conditions hold under a generative model. It is interesting to note that one of the conditions, namely equation (7) in \cite{YouRV17}, is reminiscent of the stability condition of \cite{lerman2015robust} for exact recovery in the noiseless case, that is $\cS(L)>0$, where $\cS(L)$ is defined in \eqref{eq:stab}.

\citet{rahmani2017coherence} prove recovery by CP with overwhelming probability in the same setting as \citet{soltanolkotabi2011} but with a single random subspace, where they achieve SNR on the order of $\Omega( \frac{d D}{\sqrt{D}(D-d^2)}\frac{1}{\sqrt{N_{\mathrm{out}}}}  )$. The authors also prove a recovery result for the special case of inliers uniform on the intersection of $S^{D-1}$ with a uniformly random $d$-subspace and a small percentage of outliers distributed close to a random line.
They also prove that CP can recover a set of inliers with small amounts of additive Gaussian noise in these models.
However, the theory is lacking in some important regards. First, only very special models are considered, and it is hard to see how things perform in general. A further issue is that, in the noisy case, the span of a core set of recovered inliers may not represent the underlying subspace very well. Thus, while they may be able to find a subset of the inliers, they do not give bounds on subspace approximation error for their subspace identification algorithm.

\subsection{Exhaustive Subspace Search}
\label{subsubsec:exhaust}

\citet{hardt2013algorithms} show that RF and DRF, which were discussed in \S\ref{subsec:exhaust}, can recover a subspace in the noiseless case for very low SNRs if the dataset is in general position with respect to the underlying subspace $L_*$. That is, any $D$ data points are linearly independent if and only if at most $d$ of them are inliers from $L_*$. This means that $L_*$ is the only low-dimensional structure in the data. Note that this definition is similar but different than the one in \S\ref{subsec:robcovtheory}, where inliers are in general position with respect to $L_*$ and outliers are in general position with respect to $L_*^\perp$.
Their theorems for RF and DRF are formally stated as follows (with an improvement on the expected number of iterations by \citet{ariascastro2017ransac}):
\begin{theorem}[\citet{hardt2013algorithms,ariascastro2017ransac}]
Assume that $\cX \subset \reals^D$ is a noiseless inliers-outliers dataset in $\reals^D$ with an underlying $d$-subspace $L_*$.
If $\cX$ is in general position with respect to $L_*$ and $SNR > d/(D-d)$, then RandomizedFind outputs $L_*$ with expected number of iterations that is $O(1)$, and DeRandomizedFind outputs $L_*$ in polynomial time.

On the other hand, if $SNR < d/(D-d)$, the problem becomes small set expansion hard. The small set expansion problem is conjectured to be NP-hard.
\end{theorem}
There is no existing theory for RandomizedFind and DeRandomizedFind in noisy settings.

As long as the noiseless problem is well-defined, RANSAC will succeed in finding the underlying subspace. However, in low SNR regimes, the computational time becomes an issue, as we will discuss in the next section. Further, when the parameters are set correctly, one can show that near recovery is possible with RANSAC under further assumptions on the alignment of outliers~\citep{ML18}.

\citet{Arias-Castro05connect} proved that their mutli-scale, multi-orientation scan statistics may recover inliers sampled uniformly from a $d$-dimensional graph in $[0,1]^D$
of an $m$-differentiable function, when the outliers are uniform in $[0,1]^D$ and the SNR is $\Omega(N^{-m(D-d)/(d+m(D-d))})$. They also mention results for other kinds of surfaces.

\begin{table*}[!ht]
\centering
\begin{tabular}{|c|l|}\hline
\multirow{3}{*}{\textbf{GGD}} & {$N_{\mathrm{in}}/N_{\mathrm{out}} \geq \max \left( 4\sqrt{2} \frac{\sigma_{\mathrm{out}}}{\sigma_{\mathrm{in}}} \frac{d}{\sqrt{D(D-d)}} ,  2 \frac{\sigma_{\mathrm{out}}^2}{\sigma_{\mathrm{in}}^2 } \frac{ d }{D} \right)$ ($N = O(D^2)$)} \\ & { $N_{\mathrm{in}}/N_{\mathrm{out}} \gtrapprox 0 $ ($N \to \infty$)} \\ \cline{2-2}
& \multirow{1}{*}{\emph{Deterministic condition, results for a variety of data models.}} \\ \hline
\multirow{3}{*}{\textbf{FMS}} & \multicolumn{1}{l|}{$ N_{\mathrm{in}}/N_{\mathrm{out}} ``\gtrapprox'' 0 $ ($N \to \infty$)} \\ \cline{2-2}
\multirow{2}{*}{} & \emph{Approximate recovery for large samples from spherized haystack or from two one-dimensional} \\
&\emph{subspaces on the sphere.} \\ \hline
\multirow{2}{*}{\textbf{REAPER}} & \multicolumn{1}{l|}{$ N_{\mathrm{in}}/N_{\mathrm{out}} \geq
16
\frac{\sigma_{\mathrm{out}}}{\sigma_{\mathrm{in}}} \frac{d}{D}$ ($N = O(D)$, $1 \leq d \leq (D-1)/2$)} \\ \cline{2-2}
\multirow{2}{*}{} & \emph{Deterministic condition, results for haystack where $d<(D-1)/2$.} \\ \hline
\multirow{2}{*}{\textbf{GMS}} & \multicolumn{1}{l|}{$ N_{\mathrm{in}}/N_{\mathrm{out}} \geq 4 \frac{\sigma_{\mathrm{out}}}{\sigma_{\mathrm{in}}} \frac{d}{\sqrt{(D-d)D}}$ ($N = O(D)$)} \\ \cline{2-2}
& \multicolumn{1}{l|}{\emph{Deterministic condition, results for haystack that extends to elliptical outliers.}} \\ \hline
\multirow{3}{*}{\textbf{OP}} & \multicolumn{1}{l|}{$ N_{\mathrm{in}}/N_{\mathrm{out}} \geq \frac{121 d}{9} O\left( \max(1,\log(N)/d) \right) $ ($N = O(D)$)} \\ \cline{2-2}
\multirow{2}{*}{} & \emph{Deterministic condition (formulated for arbitrary outliers) with last term in the above formula} \\ &\emph{replaced by an inlier incoherence parameter $\mu$.} \\ \hline
\multirow{2}{*}{\textbf{HR-PCA}} & \multicolumn{1}{l|}{$ N_{\mathrm{in}}/N_{\mathrm{out}} \to \infty$ ($N \to \infty$)}  \\ \cline{2-2}
& \multicolumn{1}{l|}{\emph{Weak lower bound on the expressed variance, requires fraction of outliers as input.}} \\ \hline
\multirow{2}{*}{\textbf{TME/(D)RF}} & \multicolumn{1}{l|}{$ N_{\mathrm{in}}/N_{\mathrm{out}} > \frac{d}{D-d}$ ($N = O(D)$)} \\ \cline{2-2}
& \multicolumn{1}{l|}{\emph{Result for ``general-position'' data, but does not extend to noise. }}  \\ \hline
\multirow{3}{*}{\textbf{TORP}}  &  \multicolumn{1}{l|}{$N_{\mathrm{in}}/N_{\mathrm{out}} \geq 128 d \max(1, \log(N)/d)^2 $ ($N = O(D)$)}   \\ \cline{2-2}
\multirow{2}{*}{} & \emph{Deterministic condition (formulated for arbitrary outliers) with last term replaced by an inlier} \\ &\emph{incoherence parameter $\mu$, requires fraction of outliers as input.} \\ \hline
\multirow{4}{*}{\textbf{CP}} & {$N_{\mathrm{in}}/N_{\mathrm{out}} \gtrsim d/(D-d^2) $ ($N = O(D)$, $d < \sqrt{D}$)} \\ & {$N_{\mathrm{in}}/N_{\mathrm{out}} \gtrapprox 0 $ ($N \to \infty$, $d < \sqrt{D}$)}  \\ \cline{2-2}
\multirow{2}{*}{} & \emph{Exact recovery for the spherized haystack model with a random inlier subspace and $d<\sqrt{D}$,} \\ &\emph {recovery guarantees for a special model of outliers around a line.} \\ \hline
\end{tabular}
\vspace{3mm}
\caption{Comparison of lower bounds on the SNR and a summary of guarantees. The properties of each algorithm are described in two rows. The first row provides the largest lower bounds on the SNR in the haystack model for different orders of $N$. The second row briefly comments on other guarantees  under possibly different models.
\label{tab:SNR}}
\end{table*}

\subsection{Recovery with the Haystack Model}
\label{subsec:haystack}

In Table~\ref{tab:SNR}, we compare the various theoretical guarantees under a Gaussian model of data. This is one of the simplest models to compare the theoretical SNR of algorithms outside of the worst-case outliers { (a table for the latter case will be provided in \citep{ML18})}. Here, inliers are distributed i.i.d.~$N(\bzero,\sigma_{\mathrm{in}}^2 \bP_{L_*}/d)$ and outliers are distributed i.i.d.~$N(\bzero,\sigma_{\mathrm{out}}^2\bI/D)$. Under this model, we can compare the various recovery guarantees given in the works outlined throughout the whole section. The results for the haystack model are summed up well in Table 1 of~\citep{maunu2017well}, which is extended to more methods in Table~\ref{tab:SNR}. We use the earlier abbreviations from the text.

Here, we also display the sample size necessary for the probability of recovery in each result to become close to 1. Notice that for sample sizes $N=O(D)$, the optimal SNR for all $1\leq d<D$ is on the order of $d/\sqrt{D(D-d)}$. Notice that GGD achieves this optimal bound, but requires $N=O(D^2)$, and so it has guarantees that are almost on par with state-of-the-art convex ones. If we let $N \to \infty$, we see that two methods, GGD and CP, can tolerate any fixed fraction of outliers.
The FMS can also tolerate any fixed fraction of outliers but can only nearly recover the underlying subspace up to a regularization dependent precision.
Although the result for FMS dealt with the spherized haystack model, the result can also be extended to the non-spherized haystack model with minimal effort.

We remark that CP is included here even though its model assumes a uniformly random underlying subspace, which makes the analysis easier. Nonetheless, when $N \to \infty$ this assumption makes no difference. We also note that the CP theory require that $d < \sqrt{D}$, which is a major restriction compared to other methods.
The guarantees for REAPER have the weaker requirement of $d<(D-1)/2$. Other methods can tolerate any $d <D$.

\section{Computational Complexity and Memory Requirements}
\label{sec:comp}

An important tradeoff in robust subspace recovery explores the accuracy of an algorithm versus its computational complexity or memory requirement. Because of this, it is necessary to clearly state the complexity and memory requirement of the various algorithms to see how they all scale. The complexity requirements for the various RSR algorithms are given in Table~\ref{tab:comp}. For ConstApprox, $\nnz$ refers to the number of non-zero entries in the input matrix $\bX$, and the number $\epsilon$ is the desired approximation accuracy. For ACOS, the numbers $\rho_1$ and $\rho_2$ are the row and column sampling fractions, respectively.
We also examine the memory requirement for RSR algorithms in Table~\ref{tab:mem}. The parameters for ACOS are the same as those in Table~\ref{tab:comp}.

\begin{table*}[t]
	\centering
	\begin{tabular}{l|c|c}\hline
		\textbf{Method} & \textbf{Complexity} & \textbf{Convergence Rate} \\ \hline
		Maximization $L_1$-PCA~\citep{markopoulos2014optimal} & $O(N^{\rank(\bX)})$ or $O(2^N)$ & No iteration \\ \hline
		D\"umbgen's M-estimator~\citep{dumbgen1998tyler}& $T \cdot O(N^2 D^2)$ & No result\footnotemark[1]
		\\ \hline
		Spatial Kendall's tau~\citep{visuri2000sign} & $O(N^2 D^2)$ & No iteration \\ \hline
		SRO~\citep{YouRV17} & $O(N^2 D+N^3) $ & No iteration \\ \hline
		CP~\citep{rahmani2017coherence} & $O(N^2 D)$ & No iteration \\ \hline
		GMS~\citep{zhang2014novel} & $T \cdot O(ND^2 + D^3)$ & $r$-linear convergence under the 2-subspaces criterion \\
		TME~\citep{zhang2016robust} & $\dots$ & No result\footnotemark[1] \\ \hline
		RF~\citep{hardt2013algorithms}& $T \cdot O(D^3)$ & $O(1)$ w.h.p.~when SNR$\geq d/(D-d)$ in noiseless RSR \citep{ariascastro2017ransac}\\ \hline
		REAPER~\citep{lerman2015robust}& $T \cdot O(ND^2)$ & No result\footnotemark[2] \\ \hline
		OP~\citep{mccoy2011two,xu2012robust}	& $T \cdot O(ND^2)$ & $O(\epsilon^{-1/2})$ \\
		MDR~\citep{mccoy2011two} & $\dots$ & $O(\epsilon^{-1/2})$\\
		(D)HR-PCA~\citep{xu2013outlier,Feng:12} & $\dots$ & $O(1)$ \\ \hline
		RMD~\citep{goes2014robust}& $T \cdot O(D^2)$ &  No result \\ \hline
		MKF~\citep{zhang2009median} & $T \cdot O(Dd)$ & No result  \\ \hline
		R1PCA~\citep{Ding+06} & $T \cdot O(NDd)$ & No result \\
		FMS~\citep{lerman2017fast} & $\dots$ & No general result, local $r$-linear convergence for special model \\
		GGD~\citep{maunu2017well} & $\dots$ &  $O(\epsilon^{-2})$ under stability condition, $r$-linear under further condition \\
		Projection pursuit & $\dots$ & No result \\
		\ \ \ \ \citep{huber_book,Li_85,Ammann1993,choulakian06,kwak08} &  & \\
		TORP~\citep{cherapanamjeri2017thresholding} & $\dots$ & Linear convergence in the settings of Theorem~\ref{thm:TORP} \\
		RANSAC~\citep{fischler1981random} & $\dots$ & $O(1)$ w.h.p.~when SNR$ \gtrsim d $ in noiseless general position RSR \citep{ariascastro2017ransac} \\ \hline
		SPCA~\citep{locantore1999robust} & $O(NDd)$ &  No iteration \\ \hline
		ConstApprox~\citep{clarkson2015input} & ${O(\nnz(\bX) + \poly(d/\epsilon))}$ & No iteration \\ \hline
		ACOS~\citep{li2015identifying}	& $T \cdot {O(N_{\mathrm{in}}Dd+\rho_1 \rho_2 ND \max(\rho_1 D, \rho_2 N) )}$ & $O(\epsilon^{-1/2})$ \\ \hline
	\end{tabular}
	\vspace{3mm}
	\caption{Complexity of the various RSR algorithms with constant iteration count $T$.}\label{tab:comp}
\end{table*}

\begin{table}
	\centering
	\begin{tabular}{l|c}\hline
		MKF~\citep{zhang2009median} &  ${O(dD)}$ \\ \hline
		RMD~\citep{goes2014robust} & ${O(D^2)}$ \\
		RF~\citep{hardt2013algorithms} &   \\ \hline
		ConstApprox~\citep{clarkson2015input} & ${O(\nnz(\bX) + Dd)}$ \\ \hline
		ACOS~\citep{li2015identifying} & ${O(\rho_1 \rho_2 ND + \rho_1 D d)}$ \\ \hline
		R1PCA~\citep{Ding+06} &  \\
		FMS~\citep{lerman2017fast} &  \\
		GGD~\citep{maunu2017well} &  \\
		SPCA~\citep{locantore1999robust} &  \\
		RANSAC~\citep{fischler1981random} & ${O(ND)}$  \\
		Projection pursuit~\citep{huber_book,Li_85,Ammann1993,choulakian06,kwak08,mccoy2011two} &  \\
		TORP~\citep{cherapanamjeri2017thresholding} &  \\
		OP~\citep{xu2012robust} &  \\
		$L_1$-PCA~\citep{markopoulos2014optimal} &  \\ \hline
		CP~\citep{rahmani2017coherence} & $O(N^2 + ND)$  \\
		SRO~\citep{YouRV17} &   \\ \hline
		REAPER~\citep{lerman2015robust} &  \\
		GMS~\citep{zhang2014novel} &  \\
		MDR~\citep{mccoy2011two} &  \\
		TME~\citep{zhang2016robust} & ${O(ND + D^2)}$ \\
		Spatial Kendall's tau~\citep{visuri2000sign} & \\
		D\"umbgen's M-estimator~\citep{dumbgen1998tyler} & \\
	\end{tabular}
	\caption{Memory requirement of the RSR algorithms.}\label{tab:mem}
\end{table}

We first discuss at length the results presented in Table~\ref{tab:comp}.
Many algorithms are iterative and for simplicity we assume that the number of iterations is a constant, which we denote by $T$, but this is in general problematic. Indeed, for nonconvex algorithms, we expect cases of very slow convergence since the problem is NP-hard. The following algorithms are iterative: GMS, REAPER, R1PCA, TORP, MDR, OP, FMS, GGD, TME, D\"umbgen's M-estimator, RANSAC, and RF. Among these, under certain conditions, only GMS~\citep{zhang2014novel} and GGD~\citep{maunu2017well} have guarantees for $r$-linear convergence and TORP~\citep{cherapanamjeri2017thresholding} has a guarantee for linear convergence. Also, FMS~\citep{lerman2017fast} has a weak guarantee of local $r$-linear convergence in a very special case.
The conditions for GGD can be weakened at the expense of a sublinear convergence rate, and OP and MDR have sublinear convergence in general. For convergence rate, we present the number of iterations required to achieve $\epsilon$-accuracy for the given iterative algorithms. For the online algorithms, we use $T$ to denote the number of passes over the dataset, which is often very high.

The worst complexities are for the maximization
$L_1$-PCA algorithms~\cite{markopoulos2014optimal}. The exact maximization $L_1$-PCA algorithms run in $O(N^{\rank(\bX)})$ for $N\geq D$~\citep{markopoulos2014optimal} and $O(2^N)$ for $D < N$.  It is important to note that an algorithm running in $O(N^D)$ or $O(2^N)$ is not efficient at all for big datasets.

Other very slow algorithms are D\"umbgen's M-estimator~\citep{dumbgen1998tyler} and spatial Kendall's tau~\citep{visuri2000sign} that run in $T \cdot O(N^2D^2)$ and $O(N^2D^2)$ time, respectively. Calculation of the spatial Kendall's tau matrix is more efficient because there is no iteration.

\footnotetext[1]{All results on TME also apply to D\"{u}mbgen's M-estimator. For TME, \citet{kent1988maximum} proved convergence without rate guarantees in a setting that may fit near recovery in noisy RSR, and \citet{zhang2016robust} proved convergence without rate guarantees to a singular matrix in a setting for exact recovery in noiseless RSR. We note that $r$-linear convergence was proved for the similar Maronna M-estimator in a setting that may fit near recovery in noisy RSR \citep{Arslan2004}. Lemma 1 of \citep{goes2017robust} proves global linear convergence of a regularized version of TME, but the required lower bound on the regularization parameter seems impractical for RSR.}
\footnotetext[2]{\citet{lerman2015robust} proved convergence with no rate guarantee for the REAPER procedure.}

For CP, we show the complexity of calculating the full Gram matrix on all the points in $\cX$ in the full dimension $D$. The authors advocate using a random projection and column subsampling to decrease complexity, but these ideas can be extended to many of the other methods listed here, as was done in~\cite{li2015identifying}. Using these strategies also tends to decrease the accuracy of the given algorithm.

The next slowest algorithms run in $T \cdot O(N D^2)$ or $T \cdot O(D^3)$ time. For example, GMS, REAPER, and TME must calculate the full covariance, which takes $O(ND^2)$ time~\citep{zhang2014novel,lerman2015robust,zhang2016robust}. TME and GMS require matrix inversions, which require $O(D^3)$ time~\citep{zhang2014novel,zhang2016robust}. RF requires a determinant calculation/solving a system of linear equations, which takes $O(D^3)$ time~\citep{hardt2013algorithms}. Finally, solving OP or MDR using a method such as proximal gradient descent takes $T \cdot O(ND^2)$ time.

Other algorithms operate in complexity $T \cdot O(NDd)$. This is also the complexity of using the power method to compute the PCA subspace (through the top $d$ singular vectors of the data matrix). These methods include FMS~\citep{lerman2017fast}, GGD~\citep{maunu2017well}, and TORP~\citep{cherapanamjeri2017thresholding}.
Although the RANSAC variant of~\citet{ariascastro2017ransac} runs in $T \cdot O(Dd)$, we believe that the algorithm may not be as stable as classical RANSAC~\citep{fischler1981random}.
Therefore, we display the complexity of classical RANSAC in Table~\ref{tab:comp}. \citet{ariascastro2017ransac} bound the number of iterations required for their variant of RANSAC to exactly recover the underlying subspace when SNR$\geq O(d)$, the data is noiseless, and the data is in general position with respect to the underlying subspace. They also show that the number of iterations becomes exponential in $d$ for lower SNR under the same assumptions. Despite the fact that the arguments of~\citet{ariascastro2017ransac} were proven for their variant of RANSAC, one can use these arguments for the classical RANSAC paradigm as well.

ConstApprox~\citep{clarkson2015input} is able to account for sparse input matrices, and thus operates in time dependent on the number of non-zero entries in $\bX$, which is denoted by $\nnz(\bX)$. In the case of a dense matrix $\bX$, this complexity is still $O(ND)$, which is approximately $O(NDd)$ when $d$ is small. Although this method may be fast, it has no guarantee of recovering a subspace.

Beyond the $O(NDd)$ limit for exact algorithms, some have tried to pursue even faster algorithms for approximating the underlying subspace.
For example, the work of~\citep{li2015identifying} uses row and column subsampling of the matrix to reduce the $N$ and $D$ factors and speed up computational time. The outliers can then be identified resulting in a speed-up of the algorithm. However, in this case, it makes the theoretical guarantees of any algorithm used in the subsampled case somewhat weaker. Further, after filtering the outliers, one must still calculate the inlier subspace, which takes at worst $O(N_{\mathrm{in}} D d)$. Thus, depending on the number of inliers, it may not improve much over $O(NDd)$. Indeed, since ACOS is an approximation of OP, and OP is only guaranteed for large percentages of inliers, this can still take quite long.

We also include two online algorithms in our comparisons.
The Median $K$-Flats (MKF) algorithm~\citep{zhang2009median} operates in $T \cdot O(Dd)$ time, while a slower robust mirror descent (RMD) algorithm is given in~\citep{goes2014robust}, which operates in $T \cdot O(D^2)$ time. However, these algorithms must pass over the data at least once, and so there is a hidden factor of $N$ in the iteration complexity for each of these algorithms.  Further, since the sample complexity for these methods is not known, the number of iterations (or passes over the data) required for these methods can be quite large, and, in practice, can require even more time than the other $T \cdot O(ND^2)$ methods.

Next, we discuss the memory requirements presented in Table~\ref{tab:mem}. Here, the factor of $O(N^2)$ seen for CP and SRO is typical of all strategies that follow~\citep{chen2009spectral}, due to the need to store the $N \times N$ weight matrix. The $O(D^2)$ factor is typical of methods that need to store a covariance type estimator~\citep{zhang2014novel,lerman2015robust,zhang2016robust,visuri2000sign,dumbgen1998tyler,goes2014robust}, methods that use the lifting convex relaxation technique~\citep{mccoy2011two}, or methods that require a set of $D$ points~\citep{hardt2013algorithms}. The $O(ND)$ factor is typical of methods that need to store the whole data matrix in memory or calculate the SVD of a dense matrix. Online algorithms may have improved memory because they can stream the data and only need to store an estimator at each iteration, which is the case for MKF~\citep{zhang2009median}. ConstApprox~\citep{clarkson2015input} improves over other algorithms by accounting for sparse inputs. Finally, ACOS~\citep{li2015identifying} subsamples the input matrix and reduces the amount of memory needed when running OP.

\section{Numerical Experiments and Applications}
\label{sec:exp}

Numerical experimentation is very important for proper evaluation of RSR algorithms. In this section, we outline what has currently been done to evaluate RSR algorithms on both synthetic and real datasets and what remains to be done.

A fundamental issue of the RSR problem is how to measure accuracy. The use of energy-based metrics, which may use the energies described in \S\ref{subsec:nontriv}-\S\ref{subsec:filter}, is problematic since they are inherently tied to the methods that optimize them. For example, if we wanted to evaluate subspaces by their least absolute deviations energy, we would expect the least absolute deviations algorithms to give lower energy than another method not designed to optimize that energy. For synthetic experiments, where one knows the underlying subspace, an easy choice of metric is the subspace's distance from ground truth~\citep{zhang2014novel,lerman2015robust,lerman2017fast}. For real data, the metric depends on the application. For example, when using RSR for robust dimension reduction for enhanced clustering~\citep{lerman2017fast}, the actual metric by which RSR algorithms are compared is clustering accuracy. This demonstrates that accuracy should be determined by application, not by some general energy.

In the following subsections, we will discuss the application of RSR algorithms on data examples. First, \S\ref{subsec:realdata} will discuss what experiments have been done with real data and how to evaluate them. Then, \S\ref{subsec:synth} will discuss experiments on datasets that reflect both synthetic settings of theoretical interest and stylized applications as a way to compare RSR algorithms.

\subsection{Experiments with RSR on Real Datasets}
\label{subsec:realdata}


Experimentation with RSR methods on real data is somewhat lacking due to the fact that it is a general purpose tool rather than a solver for any specific application. We can compare this to the more classical subspace modelling tool of PCA. PCA is a natural and ubiquitous data processing method, due to the fact that it can reduce the dimension of a dataset and also provide an orthogonal set of descriptive directions within the data. As such, PCA is not suited to completely solve any one problem, even though it can give insight through its descriptive factors and can act as a valuable dimension-reduction submethod.
However, as discussed earlier, PCA is not robust to outliers within a dataset.

RSR should mimic the applicability of PCA and be a general purpose tool for dimension reduction, while at the same time not being as sensitive to corrupted data.
In this way, the hope would be that an RSR algorithm would perform as well or better than PCA on most, if not all, datasets that require some form of dimension reduction.
We remark that if one wants descriptive robust orthogonal factors with reduced dimension, then one may use PCA on the projection of the dataset, or its estimated inliers, onto the subspace obtained by an RSR algorithm.

Because RSR is such a general purpose tool, it is hard to point to any one stand out application. And, as we will discuss in the coming subsections, most of what has been done in the literature is quite artifical. In what has been done, we find the dimensionality reduction and denoising aspects of RSR algorithms to be the most intriguing. An important thrust for future research should be testing RSR algorithms in more real data scenarios. In particular, it would be useful to compile a database of example datasets to further test RSR algorithms as a dimension-reduction preprocessor.

In the following two subsections, we will outline experiments that have been done with RSR algorithms on real datasets. In \S\ref{sec:realdimred}, we discuss the use of RSR algorithms for dimension reduction and data preprocessing. Then, in \S\ref{sec:imagedat}, we discuss the application of RSR algorithms to image datasets.

\subsubsection{Dimensionality Reduction for Data Preprocessing}
\label{sec:realdimred}

One intriguing property of PCA is its ability to reduce dimensionality of data while simultaneously reducing noise~\citep{vempala_kannan_spectr_alg,hopcroft_kannan_foundations}. Analogously, it has also been found that robust subspaces can have great descriptive power in the presence of noise and outliers. For example, the potential application of dimension reduction by RSR algorithms in astrophysics data was first explored in~\citep{Budavari_astro_09}, and then later considered again in~\citep{lerman2017fast}. \citet{lerman2017fast} also demonstrate the descriptive power of RSR on clustering activity time series. More examples of robust dimension reduction for classification and regression can be seen in~\citep{huroyan2017distributed}.

\citet{mccoy2011two} test their low-leverage decomposition (which is the same as OP) on Fisher's iris data. They show that a low-dimensional, robust subspace can describe the observations from one of the flower varieties quite well. In this experiment, they use a dataset with many observations from one flower type and ``corrupt'' the sample with observations of other flower types. The results visually show that RSR can capture more variation of the inlier flower type than PCA, although the authors do not give a quantitative measure of this.

Other work has considered using the RSR representation for visualization of genomics data~\citep{Price_etal_2006,novembre2008europe}. Here, outliers are filtered and PCA is done on the resulting datasets. In particular,~\citet{novembre2008europe} show that this combination of filtering and PCA yields insightful visualizations that compare genes and geography.

In all of these experiments, RSR is a useful off-the-shelf tool for dimension reduction, data preprocessing, and visualization. Some quantification of the success of RSR methods appears in~\citep{lerman2017fast} and \citep{huroyan2017distributed}, although this is only done for a few datasets. More extensive experimentation with a large database is needed to study the effectiveness of RSR as such a tool.

\subsubsection{Image Data}
\label{sec:imagedat}

A popular task in machine learning is recognition of handwritten digits. Inspired by this,~\citet{xu2012robust} considered a stylized experiment to show the capability of an RSR method to find a descriptive subspace to recognize differences between 1's and 7's. However, this experiment is only visual and does not have any quantitative measures of performance.

Many researchers have also tried to apply RSR algorithms to video surveillance data~\citep{zhang2014novel}. However, we argue that this is not a proper application of RSR algorithms, and it seems that RPCA, which addresses sparse-corruptions, models this application better. And indeed, RPCA works have shown impressive results on video surveillance data~\citep{he2011online,he2012incremental}.

Other works have studied the use of RSR algorithms on datasets of face images~\citep{lerman2015robust,zhang2016robust,lerman2017fast}. Such experiments are usually synthetic in some sense, and so we leave their discussion to \S\ref{subsec:synth}. The datasets are generated based on the observation that images of the same face under changing illumination approximately lie on a linear subspace~\citep{basri2003lambertian}.

\subsection{Experiments with RSR on Synthetic and Stylized Datasets}
\label{subsec:synth}

With the lack of real experimentation pointed out in the last section, we will resort to looking at synthetic experiments and stylized applications in this section.
We first discuss previous experiments on synthetic data in \S\ref{subsec:synthrev}.
We will then include a baseline simulation with the haystack model in \S\ref{subsec:haystacksim}. Next, we supplement this with a stylized application of face subspace recovery in \S\ref{subsec:blurryfacesim}.

\subsubsection{Review of Experiments on Synthetic Data}
\label{subsec:synthrev}

In most of the works we have reviewed, experiments were run on synthetic data to show the usefulness of the developed methods. It is hard to find a complete comparison of the various subspace recovery methods, and, to our knowledge,~\citet{lerman2017fast} provide the most comprehensive comparison of RSR algorithms on Gaussian generative models.

The extensive experiments in~\citep{lerman2017fast} compare all of the various algorithms on synthetic data drawn from the haystack model of~\citep{lerman2015robust} in various regimes. For most cases, the authors found FMS and TME to be the most robust to high percentages of outliers in these models. However, TME had a higher runtime, which matches the larger computational complexity $O(T \cdot \max (ND^2,D^3))$ versus $O(T \cdot NDd)$ for FMS. Algorithms that were not sufficiently accurate were MKF~\citep{MKF_workshop09}, REAPER~\citep{lerman2015robust}, $R1$-PCA~\citep{Ding+06}, GMS~\citep{zhang2014novel}, RMD~\citep{goes2014robust}, RPCA~\citep{candes2011robust, Lin_Chen_Ma_2010, NIPS2011_4434}, HR-PCA and DHR-PCA~\citep{xu2013outlier,Feng:12}, LLD and MDR~\citep{mccoy2011two}, and OP~\citep{xu2012robust}.

Another interesting experiment can be seen in~\citep{zhang2014novel}, where the authors test robustness of various algorithms with respect to asymmetric outliers. Here, outliers are distributed i.i.d.~from the uniform distribution on $[0,1]^D$, and inliers follow a Gaussian distribution on a random subspace in $\reals^D$. In this model, the outliers are highly asymmetric with respect to the underlying subspace.

Other works have used stylized applications to test RSR algorithms on datasets with some real characteristics. For example, the inliers in a common real data example are images of a single person's face with constant pose and varying illuminations. In this case, the face images are known to approximately lie on a 9-dimensional linear subspace~\citep{basri2003lambertian}.
The ``Faces in a Crowd'' experiment is one stylized example of identifying a face subspace in a dataset with outliers~\citep{lerman2015robust,zhang2016robust,lerman2017fast}.
Here, the outliers are taken to be other natural images, and the goal is to recover the underlying face subspace.
Such a dataset is obviously stylized, since it arises nowhere in
practice


\subsubsection{Haystack Model Simulation}
\label{subsec:haystacksim}

While there is a great need for new statistical models of data, we believe that comparison of performance under the haystack model has value. The main deficiencies of the haystack model (and to a certain extent, the generalized haystack model), are: 1) when not normalized to the sphere some simple statistics may distinguish inliers from outliers, 2) recovery under the haystack model can be easy for some algorithms. Although this may raise concerns, we believe that some algorithms successful on the haystack model will succeed on many other models and settings. To this end, we also include additional tests in \S\ref{subsec:blurryfacesim}.

\begin{figure*}[!htbp]
	\centering
	\includegraphics[width=.5\textwidth]{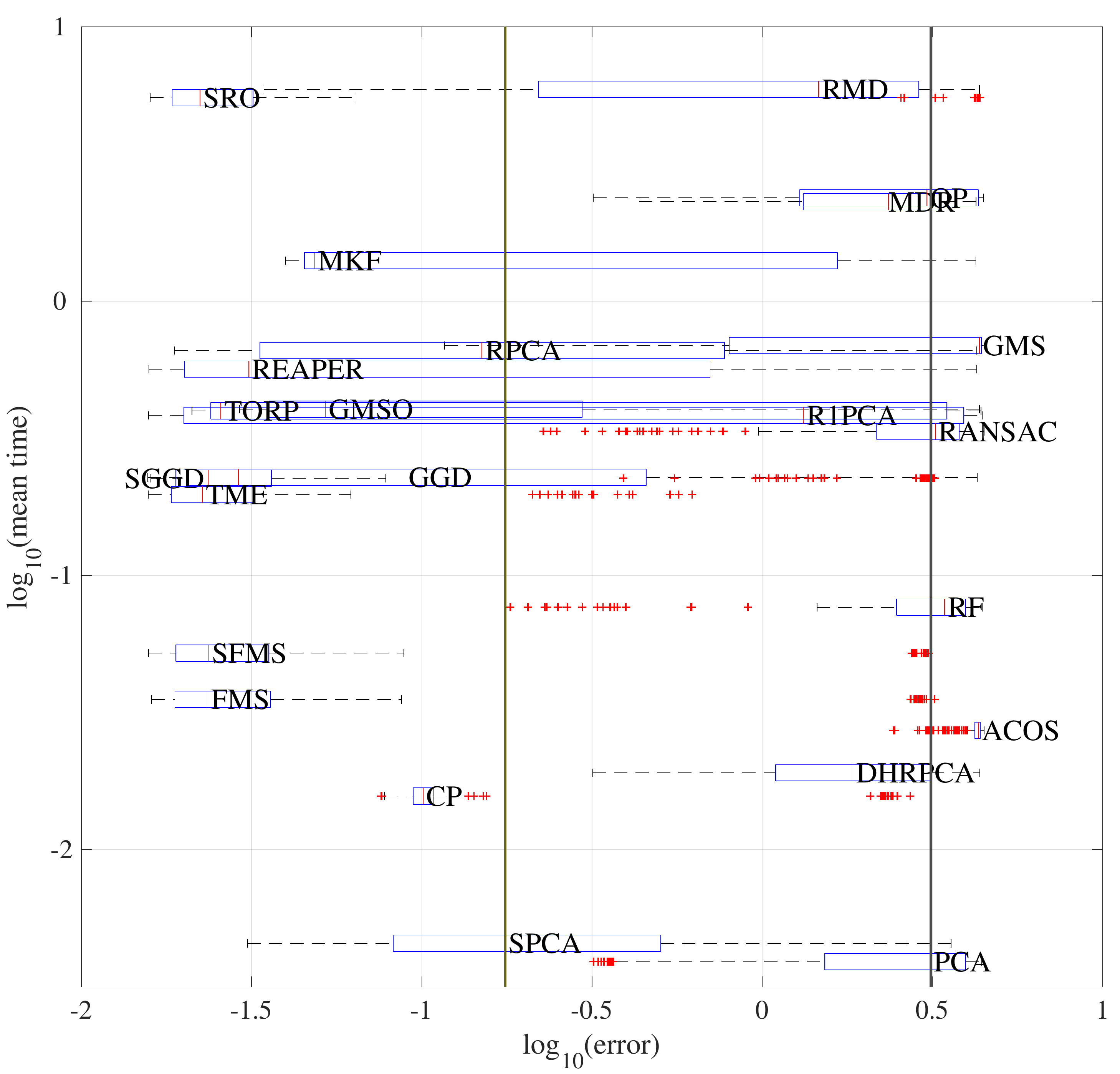}
	\caption{Accuracy-time comparison for various RSR algorithms under the haystack model. Here, we generate inliers i.i.d.~$N(\bzero,\bP_{L_*}/d)$ and outliers $N(\bzero,\bI/D)$, with $N=400$, $D=200$, and $d=10$. We also generate added noise distributed i.i.d.~$N(\bzero,10^{-4} \bI / D)$. Twenty datasets are generated at each percentage of outliers 5\%, 10\%, \dots, 95\%, and the RSR algorithms are run to calculate a robust subspace. Error is calculated as distance to ground truth (square root of sum of squared principal angles), and the runtime is recorded. The $y$-value of the boxplot represents the log-mean runtime and the $x$-value represents the log-error.}
	\label{fig:synthexp}
\end{figure*}

Our summary experiment on the haystack model is given in Figure~\ref{fig:synthexp}. We run an analogous experiment to that in~\citet{lerman2017fast}, which includes as many algorithms as possible. Here, we fix the parameters $N=400$, $D=200$, and $d=10$, and we generate inliers i.i.d.~$N(\bzero,\bP_{L_*}/d)$ and outliers i.i.d.~$N(\bzero,\bI/D)$. We perturb all points by additional noise distributed i.i.d.~$N(\bzero,10^{-4}\bI)$. We generate 20 datasets at each fixed outlier percentages $5\%,10\%,\dots,95\%$, resulting in 400 errors and times for each algorithm. These are summarized in box plots, whose $x$-values are the log-errors and $y$-values are the log-mean times for each algorithm. The edges of the boxes represent the 25th and 75th percentiles of the log-errors, and the red line represents the median log-error. The extreme ends of the whiskers represent a 99.3\% coverage interval under the assumption that the log-errors are Gaussian. The red points are errors that lie outside of this interval. The further down and left an algorithm is, the better it performs.

Acronyms or names for the algorithms are as follows: TME (Tyler's M-estimator~\citep{zhang2016robust}), (S)FMS ((Spherized) Fast Median Subspace~\citep{lerman2017fast}), (S)GGD ((Spherized) Geodesic Gradient Descent~\citep{maunu2017well}), REAPER~\citep{lerman2015robust}, GMS (Geometric Median Subspace~\citep{zhang2014novel}),
 GMSO (Geometric Median Subspace with $1.5D$ added spherically symmetric Gaussian outliers~\citep{zhang2014novel}),
 OP (Outlier Pursuit~\citep{xu2012robust,mccoy2011two}), MDR (Maximum Mean Absolute Deviation Rounding~\citep{mccoy2011two}), DHRPCA (Deterministic High-Dimensional Robust PCA~\citep{Feng:12}), R1PCA (Rotational Invariant $L_1$-norm PCA~\citep{Ding+06}), (S)PCA ((Spherized) Principal Component Analysis), MKF (Median $K$-Flats~\citep{MKF_workshop09}), SRO (Self-Representation Outlier Detection~\citep{YouRV17}), RPCA (Robust PCA, for which principal component pursuit was used)~\citep{rpca_code09}, RMD (Robust Online Mirror-Descent PCA~\citep{goes2014robust}), ACOS (Adaptive Compressive Sampling~\citep{li2015identifying}), TORP (Thresholding Based Outlier-Robust PCA~\citep{cherapanamjeri2017thresholding}), CP (Coherence Pursuit~\citep{rahmani2017coherence}), and RANSAC~\citep{fischler1981random,ariascastro2017ransac}. All algorithms are run with default parameters using code produced by the authors when available. For RF, we choose the determinant threshold $\delta$ to be $10^{-3}$.
 OP uses $\lambda = 0.8 \sqrt{D/N}$, as was used in~\citep{zhang2014novel}, and we found this choice to perform much better than the recommended $3/(7\sqrt{N_{\mathrm{out}}})$. ACOS uses this same $\lambda$ when it calls OP and also uses a subsampling rate of 1/5. For TORP and DHRPCA, we set the percentage of outliers to be $\alpha=0.5$ because there is no easy procedure to estimate this parameter in general.  MKF passes over the data ten times and RMD passes over the data twice. For RANSAC, we use the RSR variant described in \S\ref{subsec:exhaust}. We run 500 iterations and return the subspace with the best consensus number out of these iterations. We set the consensus threshold to be $10^{-3}$.
 For CP, we implemented Algorithm 2 in~\citep{rahmani2017coherence} with recommended parameters. This procedure, which is advocated by the authors of~\citep{rahmani2017coherence} for dealing with noise, was not implemented in their code, and direct implementation of their original code for noiseless RSR was not satisfying. We set the threshold parameter to be the standard deviation of the noise, which is unknown to the user, and thus running CP in this way is somewhat unrealistic. We also set the projection dimension to be $2 \cdot d$, and the algorithm is run 5 times. PCA is used to find the underlying subspace on the set of all inliers identified in the 5 runs put together.

In these tests, we do not compare with DPCP~\citep{tsakiris2015dual}, since the code provided online is really just an iterative application of a slower version of the FMS algorithm, and generally DPCP is meant for the setting of large $d$.

As we can see from this plot, the most accurate algorithms are FMS, SFMS, TME, SGGD, and SRO. Out of these algorithms, FMS is the fastest. TORP also performs well on this data when the correct percentage of outliers is used, but we cannot assume that this is known in practice. Even so, TORP is not as accurate and fast as FMS. DHRPCA does not perform well even if the true percentage of outliers is used. CP, despite having higher complexity than many other algorithms, is faster due to the fact that it is non-iterative (although it will not scale as well to large datasets).

\subsubsection{The Blurryface Model and Simulations}
\label{subsec:blurryfacesim}

We propose the \emph{blurryface model} for statistically generating data in a stylized application of recognizing the most significant face in a dataset with many face images. We simulate such data and test all implemented RSR algorithms.
The motivation here is to generate data with statistics that mimic real data.
As was mentioned earlier, images of a single person's face under varying illumination and constant pose approximately lie on a subspace of dimension 9~\citep{basri2003lambertian}. This experiment tries to recover this 9-dimensional subspace in a dataset with outliers.

We take images from the Extended Yale B face database~\citep{KCLee05} and center each subject's subset. Since there are only 64 images per person in the database, we develop the following procedure for generating low-dimensional inlier faces. First, we take all centered images of the first subject's face and calculate the sample covariance, $\hat{\bSigma}$, along with its eigenvalue decomposition. We keep only the top 9 eigenvectors and eigenvalues and store them in $\bU_*$ and $\bS_*$, respectively, so that $\hat{\bSigma} \approx \bU_* \bS_* \bU_*^T$. Synthetic inlier faces are generated i.i.d.~$N(\bzero,c_1\bU_* \bU_*^T /d)$ or i.i.d.~$N(\bzero,c_2 \bU_* \bS_* \bU_*^T)$. The first model is the \emph{spherically symmetric} inlier model and the second model is the \emph{elliptical} inlier model.  The constant $c_1$ is the average squared norm of all centered faces in the database and $c_2 = c_1/(\Tr (\bS_{*}))$. These constants are designed to give inliers comparable magnitude to the original centered faces. In both experiments, outliers are sampled without replacement from the other faces in the database. In both experiments, we also perturb all points by small Gaussian noise sampled i.i.d.~$N(\bzero,10^{-4} \bI / D)$ (which gives rise to ``blurry'' faces).

We note that the distribution of the inlier faces is quite different than the natural face images, which lie in a cone. This experiment is just meant to approximate this distribution to some reasonable degree and allow for easy generation of samples. Outliers all lie in a cone and are asymmetric, which makes subspace recovery more challenging than in the haystack model.

\begin{figure*}[!htbp]
	\centering
	\includegraphics[width=.5\textwidth]{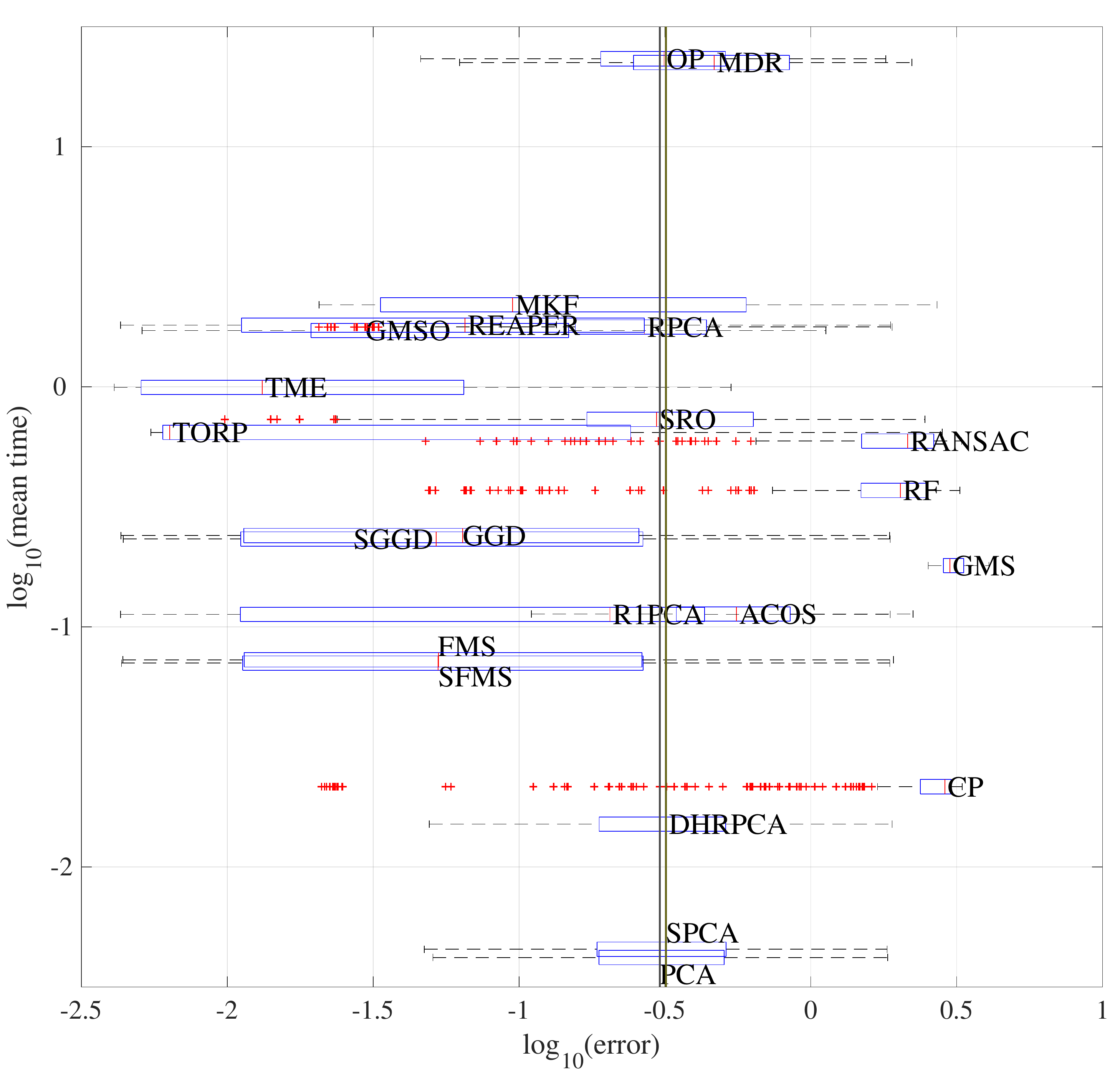}
    \includegraphics[width=.5\textwidth]{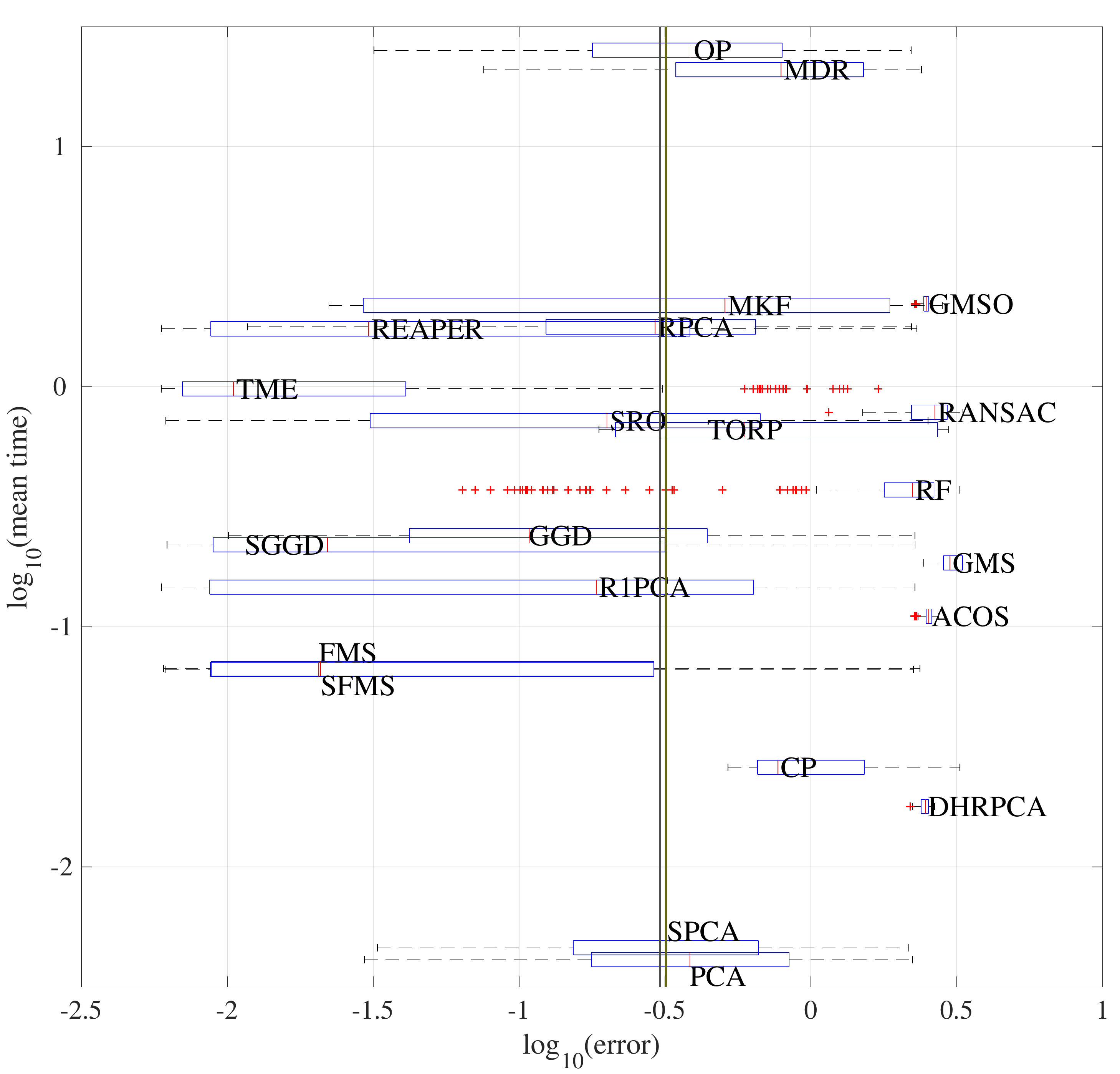}
    \caption{Accuracy-time comparison for various RSR algorithms on the blurryface data.
  Here, we generate inliers on the 9-dimensional face subspace estimated from the data, and outliers are random images of other faces.
    On the left, we generate inliers i.i.d.~$N(\bzero,c_1\bU_* \bU_*^T/9)$, and on the right we generate inliers i.i.d.~$N(\bzero,c_2 \bU_* \bS_* \bU_*^T)$. We fix $N=500$ and $D=400$, and we also generate added noise distributed i.i.d.~$N(\bzero,10^{-4} \bI / D)$. Twenty datasets are generated at each percentage of outliers 5\%, 10\%, \dots, 95\%, and the RSR algorithms are run to calculate a robust subspace. Error is calculated as distance to ground truth (square root of sum of squared principal angles). The $y$-value of the boxplot represents the log-mean runtime and the $x$-value represents the log-error.}
	\label{fig:synthexpfaceunif}
\end{figure*}

Figure~\ref{fig:synthexpfaceunif} illustrates comparisons between the implemented RSR algorithms using data generated from the two instances of the blurryface model. We generate 20 datasets at each fixed
outlier percentage ($5\%,10\%,\dots,95\%$) resulting in 400 errors
and times for each algorithm and inlier model (spherically symmetric and elliptical). These are summarized in box plots, whose $x$-values are the log-errors and $y$-values are the log-mean times for each algorithm. An explanation of the boxes, whiskers, and points were given in \S\ref{subsec:haystacksim}. The results for the spherically symmetric inlier model are displayed on the left in Figure~\ref{fig:synthexpfaceunif}, and the results for the elliptical inlier model are given on the right. Algorithm settings are as before. We modify the thresholds for RF and RANSAC to now be $10^{-3} c$, where $c$ is the mean norm of data points in the given dataset. This threshold is designed to be almost on the order of the smallest deviation of inlier faces from the underlying subspace.

For the spherically symmetric face model, TORP and TME appear to give the best accuracy. The next most accurate algorithms are FMS/SFMS, GGD/SGGD, GMSO, and REAPER. Out of all of these accurate algorithms, the fastest is FMS/SFMS.
For the elliptical face model, the best accuracy is given by TME, followed by FMS/SFMS, SGGD, and REAPER. Out of these algorithms, FMS and SFMS are the fastest, while TME is much slower. Again, CP is fast due to its non-iterative nature and the small size of the dataset.

We note that, while GMSO has some success with the spherically symmetric inlier models, it struggles with the elliptical inliers.
TORP performs very well on the spherically symmetric blurryface data with the correct percentage of outliers, but again we cannot assume this is known in practice. Even with the correct percentage, TORP fails on the elliptical blurryface model.
Again, DHRPCA does not perform well on either example even if the true percentage of outliers is used.

Following the discussion on minimization $L_1$-PCA in \S\ref{subsubsec:l1pca}, one may suggest testing RSR on models with heavy tailed elementwise noise. For example, one may try Laplacian noise as mentioned in~\citep{Baccini96}. However, this is not the RSR problem formulated in this survey. We briefly tested such settings and noticed that most RSR algorithms are comparable on it, while RPCA~\citep{rpca_code09} performs somewhat better. This result is not surprising as the RPCA problem was discussed as a loose relaxation of the
version of L1-PCA in~\eqref{eq:l1min}.

\section{Influence of RSR on Other Methods}
\label{sec:rsr_influence}

The study of RSR can influence the development of theory and algorithms for other challenging problems. In this section we discuss the important relationship of RSR with other mathematical problems.

One problem mentioned earlier is robust subspace clustering. A mathematical formulation of this problem assumes inliers sampled from (or around) a union of subspaces and outliers sampled from a different, though somewhat restricted, distribution. The goal is to identify the different underlying subspaces. This problem was addressed in~\citep{chen2009spectral,zhang2009median,ariascastro2011spectral,lp_recovery_part1_11,lp_recovery_part2_11,soltanolkotabi2012,YouRV17} by initially applying methods for filtering outliers that are the same as those in RSR. Indeed, the idea of identifying outliers by affinities that express presence in an underlying subspace (see \S\ref{subsec:filter}) works equally well for multiple underlying subspaces.

RSR methods may improve robust subspace clustering in addition to filtering outliers. A naive approach for solving robust subspace clustering is to sequentially fit a robust subspace or a single robust direction or sequentially remove a robust direction. Some previous works have already applied such sequential RSR strategies to solve this problem. For example, one of the methods in \cite{RobustGPCA} sequentially fits a single subspace by using RANSAC. Furthermore, the method of \cite{rahmani17+_icml} can be explained as sequentially searching for a least significant orthogonal direction $\bb$ that aims to minimize \eqref{eq:dpcp}. Equivalently, it can be described by sequentially searching for a least absolute deviation hyperplane, minimizing \eqref{eq:lad} (see clarification right after \eqref{eq:dpcp}). \citet{tsakiris2015dual} sequentially use the same RSR formulation in \eqref{eq:dpcp} to solve the problem of hyperplane clustering, that is, when the dimensions of all underlying subspaces are $D-1$.
However, there are many geometric obstacles to any sequential RSR approach for general robust subspace clustering, unless one assumes a very restrictive setting.

Another possibility is to use RSR within a $K$-subspace algorithm, which generalizes $K$-means to subspaces (see, e.g., \cite{zhang2009median, lp_recovery_part2_11}). However, theoretical guarantees are not developed yet for such an algorithmic approach (\citet{lp_recovery_part2_11} provide guarantees for the oracle minimization of such an approach, but not for an algorithm minimizing it). Furthermore, this method requires knowledge of the intrinsic dimensions of the subspaces, unlike~\cite{elhamifar2011robust}. Application of the $K$-subspace strategy with RPCA instead of RSR was suggested by \cite{WSL13} to address some problems in image denoising (with nonstandard noise) and blind inpainting. In these problems, the subspaces are used as approximate models and there is some flexibility in choosing the dimensions of the subspaces.
Another possible application of RSR to robust subspace clustering is the use of robust energies in the framework proposed in~\citep{LBF_journal12}. Furthemore, the local best-fit flats in this strategy can be the output of an RSR algorithm.

RSR might be extended to the more general problem of robust manifold clustering, where the inliers are sampled from (or around) a union of manifolds and the goal is to recover the underlying manifolds~\citep{ariascastro2011spectral}. Indeed, this might be possible by restricting RSR methods to local neighborhoods. Similarly, such a strategy can apply to the problem of robust recovery of a single manifold.

Developments within RSR can be beneficial for other kinds of modeling problems. For example, convex algorithms for RSR that rely on IRLS procedures \cite{zhang2014novel, lerman2015robust} inspired the development of methods for two problems in computer vision: robust recovery of camera locations from corrupted pairwise directions  \cite{ozyesil2015robust} and robust synchronization \cite{wang2013exact}, that is, robustly estimating unknown rotations (in particular, camera orientations) from a set of corrupted pairwise rotations \cite{wang2013exact}. Furthermore, the proof of the main theorem in \cite{wang2013exact} (Theorem 4.1) borrows and adapts ideas from \cite{zhang2014novel, lerman2015robust}. Similarly, ideas of filtering outliers, which are weakly reminiscent of \cite{chen2009spectral}, were used in \cite{Shi_Lerman2018_cvpr} to enhance solutions of the camera location problem. However, the latter problem is more challenging. In this setting, the outliers are associated with pairwise directions between points and not the points themselves. Moreover, any 3 uncorrupted pairwise directions lie on a two-dimensional subspace, but the subspaces defined in this way have no direct relationships. Thus, the set of uncorrupted pairwise directions does not have a simple geometric model, such as a subspace. Other methods of filtering outlier pairwise directions have to be developed to take advantage of the more complicated geometry here.

It is likely that the recent theoretical work of \citep{maunu2017well} can be generalized to other NP-hard recovery problems that can be formulated via optimization over continuous, nonconvex sets. One immediate candidate is robust synchronization over the special orthogonal group, and we have already mentioned the influence of RSR methodology there~\citep{wang2013exact}. In these optimization problems, one may possibly extend the deterministic conditions in \citep{maunu2017well} by considering appropriate notions of permeance and alignment in the new setting. One may further guarantee recovery by a gradient descent algorithm under these extended conditions.

Another problem that requires RSR is that of finding the sparsest vector within a subspace. Indeed, \citet{QuSW16} formulated this problem by using \eqref{eq:dpcp} with $\bX \bb$ instead of $\bX^T \bb$ (note the dimension of $\bb$ changes as well). They pointed out its connection with sparse dictionary learning and sparse PCA.

Robust fundamental or essential matrix estimation in computer vision can also be cast as an RSR problem. One way to calculate such matrices is to use PCA on a set of data points, although in these settings there are frequently many outliers. Further, inlier and outlier points tend to exhibit very asymmetric distributions. Specially tailored variants of RSR methods may be able to outperform existing methods for robust fundamental matrix estimation~\citep{torr1993outlier,torr1997development,sengupta2017new}.

\section{Future Work}
\label{sec:conclusion}

One option for future work is to better understand large-sample and high-dimensional limits for RSR. Some online algorithms have been proposed for robust subspace estimation~\citep{zhang2009median,goes2014robust}, but their theoretical guarantees are not satisfying and their performance is disappointing. It is also not known how many samples are needed for these methods to converge. \citet{huroyan2017distributed} have considered distributed models for RSR. Here, the authors assume that a dataset is distributed across many nodes and communication is limited by the network structure. They show under certain conditions that it is still possible to optimize some previously proposed problem formulations~\citep{lerman2015robust,zhang2014novel,lerman2017fast} in this setting.

Affine subspace estimation is not well studied, but a potential important extension of current work. One can consider estimation on the affine Grassmannian~\citep{lim2016statistical}, although the estimation considered in~\citep{lim2016statistical} may not be tight enough. A simple idea can be seen in the IRLS procedures of GMS, REAPER, and FMS~\citep{zhang2014novel,lerman2015robust,lerman2017fast}, which have a trivial extension to affine subspaces. However, we have not seen a practical advantage to including this extension for real data, and it remains an open question to see if considering affine subspaces can add real value over centering data by the geometric median. Another possibility comes from considering D\"umbgen's M-estimator~\cite{dumbgen1998tyler}, spatial Kendall's tau~\cite{visuri2000sign}, or any symmetrized version of a robust covariance estimator~\citep{nordhausen2015cautionary}. For example, both D\"umbgen's M-estimator and spatial Kendall's tau have been considered for independent component analysis by~\citet{oja2016scatter}. As mentioned earlier, it is not immediately obvious how to estimate the offset for the affine subspace with these estimators, though.

Another potential realm that is not well understood is estimation of the subspace dimension for the RSR problem. Some work has gone into dimension estimation for PCA~\citep{kritchman2008determining,dobriban2017factor}, but there are no analogous works for RSR. And, indeed, the fastest RSR algorithms require knowledge of the subspace dimension $d$ a priori. One potential issue of direct application of these methods is that, unlike PCA, RSR methods do not give nested subspaces. This makes it harder to compare subspaces across dimensions and makes heuristic strategies, such as the elbow method, hard to motivate and costly to compute. One must also determine a good metric to compare across dimensions, for which there is no easy or obvious choice.
Thus, the development of methods for this problem would be an interesting direction for future work.

Robustness to noise in the PCA problem is also a relatively unstudied problem. One possible path is to pursue ideas similar to those in~\citep{coudron2012sample}. The work of~\cite{cherapanamjeri2017thresholding} also has a nontrivial result to noise, where they achieve similar rates as PCA to sub-Gaussian noise, even in the presence of outliers.
However,~\citep{cherapanamjeri2017thresholding} requires knowledge of the fraction of outliers, which makes the setting of the robustness to outliers easier.
But, perhaps the future of noise analysis lies in looking at heavy-tailed distributions and limits in the various models of inliers and outliers. One intriguing idea for heavy-tailed noise is given by \citep{minsker2015geometric}, where the author uses the idea of median of means to construct a robust covariance estimator. This estimator can then be used to find a PCA subspace that exhibits asymptotic sub-Gaussian estimation bounds, even in the presence of heavier tailed data. The determination of optimal rates and consideration of other noise regimes remain open problems.

One question is where the recovery theory of RSR should go next. Indeed, theoretical guarantees of recovery under special models are not the primary goal of RSR. Instead, we wish to have methods that are useful in practice. Recent work on robustness has considered how well an algorithm can perform in the presence of adversarial corruption~\citep{diakonikolas2017being,steinhardt2018resilience}. Although adversarial outliers have been considered in the context of some RSR algorithms, such as~\citep{xu2012robust}, current results are weak, and it seems that better algorithms and guarantees can be developed for these cases~\cite{ML18}.

So far, analysis of the inliers and outliers in RSR has been separated. This has led to the separate notions of permeance and restricted alignment we discussed in \S\ref{subsec:statmod}, which are each formulated with respect to the inliers or outliers alone. These independent formulations are then combined to form a stability constraint on the model. The work of~\citet{hardt2013algorithms} does not separate between two conditions of inliers and outliers, but it has a very simplistic and well-defined setting.
It would be interesting to find out if there are more refined stability conditions that involve both inliers and outliers together. In contrast to the work of~\citet{hardt2013algorithms}, these conditions would need to be more general and allow both inliers and outliers to lie on lower dimensional subspaces to capture a wider range of examples.

In terms of RSR, one can raise the question of whether or not the study of high-dimensions is really needed. Using ideas from the Johnson-Lindenstrauss lemma~\citep{johnson1984extensions}, one may think that a few random projections will maintain most of the important statistics of the data, including the low-dimensional subspace structure. However, short simulations have shown that while this can work for low percentages of outliers, it becomes harder in low SNR regimes. This should not be surprising, because the low SNR regimes experience the hardness threshold of $d/(D-d)$~\citep{hardt2013algorithms}. Indeed, the SNR threshold increases as $D$ decreases. Quantifying how well random projections work for RSR is an interesting avenue for future work.

Finally, as was mentioned in \S\ref{subsec:realdata}, there is a need for more experimentation on RSR methods. One thing we advocate is the development of a database of examples to test RSR methods on. Datasets within this database must also have properly defined measures of success that are tied to the specific application. A possible start to this database could involve wider experimentation on robust dimensionality reduction for a variety of tasks.  Another option is to develop more stylized applications to test RSR methods on datasets that mirror real data in some way.

A supplemental webpage with code and data will be provided at \url{https://twmaunu.github.io/rsr_overview/}.

\section{Acknowledgements}

This work was supported by NSF grant DMS-14-18386 and a University of Minnesota Doctoral Dissertation Fellowship. The authors would also like to thank all who contributed code used in this survey, and in particular, Ery Arias-Castro for RF, Teng Zhang for GMS, REAPER, MKF and TME, John Goes for RMD, Xingguo Li for ACOS, Chong You for SRO, and Yeshwanth Cherapanamjeri for TORP. We would also like to thank Teng Zhang{, Nicolas Gillis, and the anonymous reviewers} for helpful comments and references.

\Urlmuskip=0mu plus 1mu\relax

\bibliographystyle{plainnat}
\bibliography{refs_2017}

\begin{thebibliography}{124}
\providecommand{\natexlab}[1]{#1}
\providecommand{\url}[1]{\texttt{#1}}
\expandafter\ifx\csname urlstyle\endcsname\relax
  \providecommand{\doi}[1]{doi: #1}\else
  \providecommand{\doi}{doi: \begingroup \urlstyle{rm}\Url}\fi

\bibitem[Ammann(1993)]{Ammann1993}
L.~P. Ammann.
\newblock Robust singular value decompositions: A new approach to projection
  pursuit.
\newblock \emph{Journal of the American Statistical Association}, 88\penalty0
  (422):\penalty0 pp. 505--514, 1993.
\newblock ISSN 01621459.

\bibitem[{Arias-Castro} and {Wang}(2017)]{ariascastro2017ransac}
E.~{Arias-Castro} and J.~{Wang}.
\newblock {RANSAC Algorithms for Subspace Recovery and Subspace Clustering}.
\newblock \emph{ArXiv e-prints}, November 2017.

\bibitem[Arias-Castro et~al.(2005)Arias-Castro, Donoho, Huo, and
  Tovey]{Arias-Castro05connect}
E.~Arias-Castro, D.~L. Donoho, X.~Huo, and C.~A. Tovey.
\newblock Connect the dots: how many random points can a regular curve pass
  through?
\newblock \emph{Adv. in Appl. Probab.}, 37\penalty0 (3):\penalty0 571--603,
  2005.

\bibitem[Arias-Castro et~al.(2011)Arias-Castro, Chen, and
  Lerman]{ariascastro2011spectral}
E.~Arias-Castro, G.~Chen, and G.~Lerman.
\newblock Spectral clustering based on local linear approximations.
\newblock \emph{Electron. J. Statist.}, 5:\penalty0 1537--1587, 2011.

\bibitem[Arslan(2004)]{Arslan2004}
O.~Arslan.
\newblock Convergence behavior of an iterative reweighting algorithm to compute
  multivariate {M}-estimates for location and scatter.
\newblock \emph{Journal of Statistical Planning and Inference}, 118\penalty0
  (1-2):\penalty0 115 -- 128, 2004.
\newblock ISSN 0378-3758.
\newblock \doi{10.1016/S0378-3758(02)00402-0}.

\bibitem[Auderset et~al.(2005)Auderset, Mazza, and Ruh]{auderset2005angular}
C.~Auderset, C.~Mazza, and E.~A. Ruh.
\newblock Angular {G}aussian and {C}auchy estimation.
\newblock \emph{Journal of multivariate analysis}, 93\penalty0 (1):\penalty0
  180--197, 2005.

\bibitem[Baccini et~al.(1996)Baccini, Besse, and de~Falguerolles]{Baccini96}
A.~Baccini, Ph. Besse, and A.~de~Falguerolles.
\newblock A ${L}_1$-norm {PCA} and a heuristic approach.
\newblock In E.~Diday, Y.~Lechevalier, and O.~Opitz, editors, \emph{Ordinal and
  symbolic data analysis}, pages 359--368, New York, 1996. Springer.

\bibitem[Basri and Jacobs(2003)]{basri2003lambertian}
R.~Basri and D.~W. Jacobs.
\newblock Lambertian reflectance and linear subspaces.
\newblock \emph{IEEE transactions on pattern analysis and machine
  intelligence}, 25\penalty0 (2):\penalty0 218--233, 2003.

\bibitem[Blum et~al.(2018)Blum, Hopcroft, and
  Kannan]{hopcroft_kannan_foundations}
A.~Blum, J.~Hopcroft, and R.~Kannan.
\newblock Foundations of data science.
\newblock Available online at \url{https://www.cs.cornell.edu/jeh/book.pdf},
  2018.

\bibitem[Brooks et~al.(2013)Brooks, Dul{\'a}, and Boone]{brooks2013pure}
J.~P. Brooks, J.~H. Dul{\'a}, and E.~L. Boone.
\newblock A pure {L1}-norm principal component analysis.
\newblock \emph{Computational statistics \& data analysis}, 61:\penalty0
  83--98, 2013.

\bibitem[Budavári et~al.(2009)Budavári, Wild, Szalay, Dobos, and
  Yip]{Budavari_astro_09}
T.~Budavári, V.~Wild, A.~S. Szalay, L.~Dobos, and C.-W. Yip.
\newblock Reliable eigenspectra for new generation surveys.
\newblock \emph{Monthly Notices of the Royal Astronomical Society},
  394\penalty0 (3):\penalty0 1496--1502, 2009.

\bibitem[Cand{\`e}s and Recht(2009)]{candes2009exact}
E.~J. Cand{\`e}s and B.~Recht.
\newblock Exact matrix completion via convex optimization.
\newblock \emph{Foundations of Computational mathematics}, 9\penalty0
  (6):\penalty0 717, 2009.

\bibitem[Cand{\`e}s and Tao(2010)]{candes2010power}
E.~J. Cand{\`e}s and T.~Tao.
\newblock The power of convex relaxation: Near-optimal matrix completion.
\newblock \emph{IEEE Transactions on Information Theory}, 56\penalty0
  (5):\penalty0 2053--2080, 2010.

\bibitem[Cand{\`e}s et~al.(2011)Cand{\`e}s, Li, Ma, and
  Wright]{candes2011robust}
E.~J. Cand{\`e}s, X.~Li, Y.~Ma, and J.~Wright.
\newblock Robust principal component analysis?
\newblock \emph{Journal of the ACM (JACM)}, 58\penalty0 (3):\penalty0 11, 2011.

\bibitem[Chandrasekaran et~al.(2011)Chandrasekaran, Sanghavi, Parrilo, and
  Willsky]{chandrasekaran2011rank}
V.~Chandrasekaran, S.~Sanghavi, P.~A. Parrilo, and A.~S. Willsky.
\newblock Rank-sparsity incoherence for matrix decomposition.
\newblock \emph{SIAM Journal on Optimization}, 21\penalty0 (2):\penalty0
  572--596, 2011.

\bibitem[Chen and Lerman(2009)]{chen2009spectral}
G.~Chen and G.~Lerman.
\newblock Spectral curvature clustering ({SCC}).
\newblock \emph{Int. J. Comput. Vision}, 81\penalty0 (3):\penalty0 317--330,
  2009.

\bibitem[Cherapanamjeri et~al.(2017)Cherapanamjeri, Jain, and
  Netrapalli]{cherapanamjeri2017thresholding}
Y.~Cherapanamjeri, P.~Jain, and P.~Netrapalli.
\newblock Thresholding based outlier robust {PCA}.
\newblock In \emph{COLT}, pages 593--628, 2017.

\bibitem[Choulakian(2006)]{choulakian06}
V.~Choulakian.
\newblock ${L}_1$-norm projection pursuit principal component analysis.
\newblock \emph{Computational Statistics \& Data Analysis}, 50\penalty0
  (6):\penalty0 1441--1451, March 2006.

\bibitem[Clarkson and Woodruff(2015)]{clarkson2015input}
K.~L. Clarkson and D.~P. Woodruff.
\newblock Input sparsity and hardness for robust subspace approximation.
\newblock In \emph{Foundations of Computer Science (FOCS), 2015 IEEE 56th
  Annual Symposium on}, pages 310--329. IEEE, 2015.

\bibitem[Coudron and Lerman(2012)]{coudron2012sample}
G.~Coudron and G.~Lerman.
\newblock On the sample complexity of robust {PCA}.
\newblock In \emph{Advances in Neural Information Processing Systems}, pages
  3221--3229, 2012.

\bibitem[Davis and Kahan(1970)]{Davis1970}
C.~Davis and W.~M. Kahan.
\newblock The rotation of eigenvectors by a perturbation. iii.
\newblock \emph{SIAM J. on Numerical Analysis}, 7:\penalty0 1--46, 1970.

\bibitem[De~La~Torre and Black(2003)]{Torre:03}
Fernando De~La~Torre and M.~J Black.
\newblock A framework for robust subspace learning.
\newblock \emph{International Journal of Computer Vision}, 54\penalty0
  (1-3):\penalty0 117--142, 2003.

\bibitem[Diakonikolas et~al.(2017)Diakonikolas, Kamath, Kane, Li, Moitra, and
  Stewart]{diakonikolas2017being}
I.~Diakonikolas, G.~Kamath, D.~M. Kane, J.~Li, A.~Moitra, and Alistair Stewart.
\newblock Being robust (in high dimensions) can be practical.
\newblock In \emph{International Conference on Machine Learning}, pages
  999--1008, 2017.

\bibitem[Ding et~al.(2006)Ding, Zhou, He, and Zha]{Ding+06}
C.~Ding, D.~Zhou, X.~He, and H.~Zha.
\newblock {R}1-{PCA}: rotational invariant ${L}_1$-norm principal component
  analysis for robust subspace factorization.
\newblock In \emph{ICML '06: Proceedings of the 23rd international conference
  on Machine learning}, pages 281--288, New York, NY, USA, 2006. ACM.

\bibitem[Dobriban(2017)]{dobriban2017factor}
E.~Dobriban.
\newblock Factor selection by permutation.
\newblock \emph{arXiv preprint arXiv:1710.00479}, 2017.

\bibitem[Donoho and Gasko(1992)]{donoho1982breakdown}
D.~L. Donoho and M.~Gasko.
\newblock Breakdown properties of location estimates based on halfspace depth
  and projected outlyingness.
\newblock \emph{Ann. Statist.}, 20\penalty0 (4):\penalty0 1803--1827, 1992.
\newblock ISSN 0090-5364.
\newblock URL \url{https://doi-org.ezp3.lib.umn.edu/10.1214/aos/1176348890}.

\bibitem[D{\"u}mbgen(1998)]{dumbgen1998tyler}
L.~D{\"u}mbgen.
\newblock On {T}yler's {M}-functional of scatter in high dimension.
\newblock \emph{Annals of the Institute of Statistical Mathematics},
  50\penalty0 (3):\penalty0 471--491, 1998.

\bibitem[Edelman et~al.(1999)Edelman, Arias, and Smith]{Edelman98thegeometry}
A.~Edelman, T.~A. Arias, and S.~T. Smith.
\newblock The geometry of algorithms with orthogonality constraints.
\newblock \emph{SIAM J. Matrix Anal. Appl.}, 20\penalty0 (2):\penalty0 303--353
  (electronic), 1999.
\newblock ISSN 0895-4798.

\bibitem[Elhamifar and Vidal(2011)]{elhamifar2011robust}
E.~Elhamifar and R.~Vidal.
\newblock Robust classification using structured sparse representation.
\newblock In \emph{Computer Vision and Pattern Recognition (CVPR), 2011 IEEE
  Conference on}, pages 1873--1879. IEEE, 2011.

\bibitem[Elhamifar and Vidal(2013)]{ssc_elhamifar13}
E.~Elhamifar and R.~Vidal.
\newblock Sparse subspace clustering: Algorithm, theory, and applications.
\newblock \emph{IEEE Transactions on Pattern Analysis and Machine
  Intelligence}, PP\penalty0 (99):\penalty0 1--1, 2013.
\newblock ISSN 0162-8828.
\newblock \doi{10.1109/TPAMI.2013.57}.

\bibitem[Feng et~al.(2012)Feng, Xu, and Yan]{Feng:12}
J.~Feng, H.~Xu, and S.~Yan.
\newblock Robust {PCA} in high-dimension: A deterministic approach.
\newblock \emph{International conference on machine learning (ICML)}, 2012.

\bibitem[Feng et~al.(2013)Feng, Xu, and Yan]{feng2013robpca}
J.~Feng, H.~Xu, and S.~Yan.
\newblock Online robust {PCA} via stochastic optimization.
\newblock In \emph{Advances in Neural Information Processing Systems (NIPS)},
  2013.

\bibitem[Fischler and Bolles(1981)]{fischler1981random}
M.~A. Fischler and R.~C. Bolles.
\newblock Random sample consensus: a paradigm for model fitting with
  applications to image analysis and automated cartography.
\newblock \emph{Communications of the ACM}, 24\penalty0 (6):\penalty0 381--395,
  1981.

\bibitem[Friedman and Tukey(1974)]{friedman1974projection}
J.~H. Friedman and J.~W. Tukey.
\newblock A projection pursuit algorithm for exploratory data analysis.
\newblock \emph{IEEE Transactions on computers}, 100\penalty0 (9):\penalty0
  881--890, 1974.

\bibitem[Gillis and Vavasis(2015)]{gillis2015complexity}
N.~Gillis and S.~A. Vavasis.
\newblock On the complexity of robust {PCA} and {$\ell_1$}-norm low-rank matrix
  approximation.
\newblock \emph{arXiv preprint arXiv:1509.09236}, 2015.

\bibitem[Glaz and Balakrishnan(1999)]{glaz1999scan}
J.~Glaz and N.~Balakrishnan.
\newblock \emph{Scan statistics and applications}.
\newblock Springer Science \& Business Media, 1999.

\bibitem[Glaz et~al.(2001)Glaz, Naus, and Wallenstein]{Glaz01}
J.~Glaz, J.~Naus, and S.~Wallenstein.
\newblock \emph{Scan Statistics}.
\newblock Springer Series in Statistics. Springer-Verlag, New York, 2001.
\newblock ISBN 0-387-98819-X.

\bibitem[Goes et~al.(2014)Goes, Zhang, Arora, and Lerman]{goes2014robust}
J.~Goes, T.~Zhang, R.~Arora, and G.~Lerman.
\newblock Robust stochastic principal component analysis.
\newblock In \emph{Artificial Intelligence and Statistics}, pages 266--274,
  2014.

\bibitem[Goes et~al.(2017)Goes, Lerman, and Nadler]{goes2017robust}
J.~Goes, G.~Lerman, and B.~Nadler.
\newblock Robust sparse covariance estimation by thresholding {T}yler's
  {M}-estimator.
\newblock \emph{arXiv preprint arXiv:1706.08020}, 2017.

\bibitem[Golub and Van~Loan(1996)]{golub1996matrix}
G.~H. Golub and C.~F. Van~Loan.
\newblock \emph{Matrix Computations}.
\newblock Johns Hopkins Studies in the Mathematical Sciences. Johns Hopkins
  University Press, Baltimore, MD, 3rd edition, 1996.
\newblock ISBN 0-8018-5413-X; 0-8018-5414-8.

\bibitem[Hardt and Moitra(2013)]{hardt2013algorithms}
M.~Hardt and A.~Moitra.
\newblock Algorithms and hardness for robust subspace recovery.
\newblock In \emph{Conference on Learning Theory (COLT)}, pages 354--375, 2013.

\bibitem[He et~al.(2011)He, Balzano, and Lui]{he2011online}
J.~He, L.~Balzano, and J.~C.~S. Lui.
\newblock Online robust subspace tracking from partial information.
\newblock \emph{CoRR}, abs/1109.3827, 2011.

\bibitem[He et~al.(2012)He, Balzano, and Szlam]{he2012incremental}
J.~He, L.~Balzano, and A.~Szlam.
\newblock Incremental gradient on the grassmannian for online foreground and
  background separation in subsampled video.
\newblock In \emph{Computer Vision and Pattern Recognition (CVPR), 2012 IEEE
  Conference on}, pages 1568--1575. IEEE, 2012.

\bibitem[Huber and Ronchetti(2009)]{huber_book}
P.~J. Huber and E.~M. Ronchetti.
\newblock \emph{Robust Statistics}.
\newblock Wiley Series in Probability and Statistics. Wiley, Hoboken, NJ, 2nd
  edition, 2009.
\newblock ISBN 978-0-470-12990-6.
\newblock \doi{10.1002/9780470434697}.

\bibitem[Huroyan and Lerman(2017)]{huroyan2017distributed}
V.~Huroyan and G.~Lerman.
\newblock Distributed robust subspace recovery.
\newblock \emph{arXiv preprint arXiv:1705.09382}, 2017.

\bibitem[Johnson and Lindenstrauss(1984)]{johnson1984extensions}
W.~B. Johnson and J.~Lindenstrauss.
\newblock Extensions of {L}ipschitz mappings into a {H}ilbert space.
\newblock \emph{Contemporary mathematics}, 26\penalty0 (189-206):\penalty0 1,
  1984.

\bibitem[Kannan and Vempala(2009)]{vempala_kannan_spectr_alg}
R.~Kannan and S.~Vempala.
\newblock Spectral algorithms.
\newblock \emph{Foundations and Trends in Theoretical Computer Science},
  4\penalty0 (3-4):\penalty0 157--288, 2009.
\newblock \doi{10.1561/0400000025}.
\newblock URL \url{https://doi.org/10.1561/0400000025}.

\bibitem[Ke and Kanade(2005)]{ke2005robust}
Q.~Ke and T.~Kanade.
\newblock Robust {$L_1$} norm factorization in the presence of outliers and
  missing data by alternative convex programming.
\newblock In \emph{Conference on Computer Vision and Pattern Recognition},
  volume~1, pages 739--746. IEEE, 2005.

\bibitem[Kent and Tyler(1988)]{kent1988maximum}
J.~T. Kent and D.~E. Tyler.
\newblock Maximum likelihood estimation for the wrapped {C}auchy distribution.
\newblock \emph{Journal of Applied Statistics}, 15\penalty0 (2):\penalty0
  247--254, 1988.

\bibitem[Kritchman and Nadler(2008)]{kritchman2008determining}
S.~Kritchman and B.~Nadler.
\newblock Determining the number of components in a factor model from limited
  noisy data.
\newblock \emph{Chemometrics and Intelligent Laboratory Systems}, 94\penalty0
  (1):\penalty0 19--32, 2008.

\bibitem[Kwak(2008)]{kwak08}
N.~Kwak.
\newblock Principal component analysis based on ${L}_1$-norm maximization.
\newblock \emph{Pattern Analysis and Machine Intelligence, IEEE Transactions
  on}, 30\penalty0 (9):\penalty0 1672--1680, 2008.
\newblock \doi{10.1109/TPAMI.2008.114}.

\bibitem[Lee et~al.(2005)Lee, Ho, and Kriegman]{KCLee05}
K.-C. Lee, J.~Ho, and D.~Kriegman.
\newblock Acquiring linear subspaces for face recognition under variable
  lighting.
\newblock \emph{Pattern Analysis and Machine Intelligence, IEEE Transactions
  on}, 27\penalty0 (5):\penalty0 684--698, 2005.

\bibitem[Lerman and Maunu(2017)]{lerman2017fast}
G.~Lerman and T.~Maunu.
\newblock Fast, robust and non-convex subspace recovery.
\newblock \emph{Accepted for publication, Information and Inference}, 2017.

\bibitem[Lerman and Zhang(2011)]{lp_recovery_part2_11}
G.~Lerman and T.~Zhang.
\newblock Robust recovery of multiple subspaces by geometric ${{l_p}}$
  minimization.
\newblock \emph{Ann. Statist.}, 39\penalty0 (5):\penalty0 2686--2715, 2011.

\bibitem[Lerman and Zhang(2014)]{lp_recovery_part1_11}
G.~Lerman and T.~Zhang.
\newblock {$l_p$}-recovery of the most significant subspace among multiple
  subspaces with outliers.
\newblock \emph{Constructive Approximation}, 40\penalty0 (3):\penalty0
  329--385, 2014.

\bibitem[Lerman et~al.(2015)Lerman, McCoy, Tropp, and Zhang]{lerman2015robust}
G.~Lerman, M.~B. McCoy, J.~A. Tropp, and T.~Zhang.
\newblock Robust computation of linear models by convex relaxation.
\newblock \emph{Foundations of Computational Mathematics}, 15\penalty0
  (2):\penalty0 363--410, 2015.

\bibitem[Li and Chen(1985)]{Li_85}
G.~Li and Z.~Chen.
\newblock Projection-pursuit approach to robust dispersion matrices and
  principal components: Primary theory and {M}onte {C}arlo.
\newblock \emph{Journal of the American Statistical Association}, 80\penalty0
  (391):\penalty0 759--766, 1985.
\newblock ISSN 01621459.
\newblock \doi{10.2307/2288497}.

\bibitem[Li and Haupt(2015)]{li2015identifying}
X.~Li and J.~D. Haupt.
\newblock Identifying outliers in large matrices via randomized adaptive
  compressive sampling.
\newblock \emph{IEEE Trans. Signal Processing}, 63\penalty0 (7):\penalty0
  1792--1807, 2015.

\bibitem[Lim et~al.(2016)Lim, Wong, and Ye]{lim2016statistical}
L.~Lim, K.~S. Wong, and K.~Ye.
\newblock Statistical estimation and the affine {G}rassmannian.
\newblock \emph{arXiv preprint arXiv:1607.01833}, 2016.

\bibitem[Lin et~al.(2009)Lin, Ganesh, J., Wu, Chen, and Ma]{rpca_code09}
Z.~Lin, A.~Ganesh, J., L.~Wu, M.~Chen, and Y.~Ma.
\newblock Fast convex optimization algorithms for exact recovery of a corrupted
  low-rank matrix.
\newblock In \emph{In Intl. Workshop on Comp. Adv. in Multi-Sensor Adapt.
  Processing, Aruba, Dutch Antilles}, 2009.

\bibitem[{Lin} et~al.(2010){Lin}, {Chen}, and {Ma}]{Lin_Chen_Ma_2010}
Z.~{Lin}, M.~{Chen}, and Y.~{Ma}.
\newblock {The Augmented Lagrange Multiplier Method for Exact Recovery of
  Corrupted Low-Rank Matrices}.
\newblock \emph{ArXiv e-prints}, September 2010.

\bibitem[Lin et~al.(2011)Lin, Liu, and Su]{NIPS2011_4434}
Z.~Lin, R.~Liu, and Z.~Su.
\newblock Linearized alternating direction method with adaptive penalty for
  low-rank representation.
\newblock In J.~Shawe-Taylor, R.~S. Zemel, P.~L. Bartlett, F.~Pereira, and
  K.~Q. Weinberger, editors, \emph{Advances in Neural Information Processing
  Systems 24}, pages 612--620. Curran Associates, Inc., 2011.

\bibitem[Locantore et~al.(1999)Locantore, Marron, Simpson, Tripoli, Zhang, and
  Cohen]{locantore1999robust}
N.~Locantore, J.~S. Marron, D.~G. Simpson, N.~Tripoli, J.~T. Zhang, and K.~L.
  Cohen.
\newblock Robust principal component analysis for functional data.
\newblock \emph{Test}, 8\penalty0 (1):\penalty0 1--73, 1999.

\bibitem[Lopuha{\"a} and Rousseeuw(1991)]{lopuha_rousseeuw_robust91}
H.~P. Lopuha{\"a} and P.~J. Rousseeuw.
\newblock Breakdown points of affine equivariant estimators of multivariate
  location and covariance matrices.
\newblock \emph{Ann. Statist.}, 19\penalty0 (1):\penalty0 229--248, 1991.

\bibitem[Lopuhaa and Rousseeuw(1991)]{lopuhaa1991breakdown}
H.~P. Lopuhaa and P.~J. Rousseeuw.
\newblock Breakdown points of affine equivariant estimators of multivariate
  location and covariance matrices.
\newblock \emph{The Annals of Statistics}, pages 229--248, 1991.

\bibitem[Marden(1999)]{marden1999some}
J.~I. Marden.
\newblock Some robust estimates of principal components.
\newblock \emph{Statistics \& probability letters}, 43\penalty0 (4):\penalty0
  349--359, 1999.

\bibitem[Markopoulos et~al.(2014)Markopoulos, Karystinos, and
  Pados]{markopoulos2014optimal}
P.~P. Markopoulos, G.~N. Karystinos, and D.~A. Pados.
\newblock Optimal algorithms for {$L_1$}-subspace signal processing.
\newblock \emph{IEEE Transactions on Signal Processing}, 62\penalty0
  (19):\penalty0 5046--5058, 2014.

\bibitem[Markopoulos et~al.(2018)Markopoulos, Kundu, Chamadia, Tsagkarakis, and
  Pados]{markopoulos2018outlier}
P.~P. Markopoulos, S.~Kundu, S.~Chamadia, N.~Tsagkarakis, and D.~A. Pados.
\newblock Outlier-resistant data processing with {L1}-norm principal component
  analysis.
\newblock In \emph{Advances in Principal Component Analysis}, pages 121--135.
  Springer, 2018.

\bibitem[Maronna(1976)]{maronna1976robust}
R.~A. Maronna.
\newblock Robust estimation of multivariate location and scatter.
\newblock \emph{Wiley StatsRef: Statistics Reference Online}, 1976.

\bibitem[Maronna(2005)]{maronna2005principal}
R.~A. Maronna.
\newblock Principal components and orthogonal regression based on robust
  scales.
\newblock \emph{Technometrics}, 47\penalty0 (3):\penalty0 264--273, 2005.

\bibitem[Maronna et~al.(2006)Maronna, Martin, and Yohai]{maronna2006robust}
R.~A. Maronna, R.~D. Martin, and V.~J. Yohai.
\newblock \emph{Robust statistics: Theory and methods}.
\newblock Wiley Series in Probability and Statistics. John Wiley \& Sons Ltd.,
  Chichester, 2006.
\newblock ISBN 978-0-470-01092-1; 0-470-01092-4.

\bibitem[Maunu and Lerman()]{ML18}
T.~Maunu and G.~Lerman.
\newblock Robust subspace recovery with adverserial outliers.
\newblock In preparation.

\bibitem[Maunu et~al.(2017)Maunu, Zhang, and Lerman]{maunu2017well}
T.~Maunu, T.~Zhang, and G.~Lerman.
\newblock A well-tempered landscape for non-convex robust subspace recovery.
\newblock \emph{arXiv preprint arXiv:1706.03896}, 2017.

\bibitem[McCoy and Tropp(2011)]{mccoy2011two}
M.~McCoy and J.~A Tropp.
\newblock Two proposals for robust {PCA} using semidefinite programming.
\newblock \emph{Electronic Journal of Statistics}, 5:\penalty0 1123--1160,
  2011.

\bibitem[Minsker(2015)]{minsker2015geometric}
S.~Minsker.
\newblock Geometric median and robust estimation in banach spaces.
\newblock \emph{Bernoulli}, 21\penalty0 (4):\penalty0 2308--2335, 2015.

\bibitem[Naor et~al.(2013)Naor, Regev, and Vidick]{naor2013efficient}
A.~Naor, O.~Regev, and T.~Vidick.
\newblock Efficient rounding for the noncommutative {Grothendieck} inequality.
\newblock In \emph{Proceedings of the forty-fifth annual ACM symposium on
  Theory of computing}, pages 71--80. ACM, 2013.

\bibitem[Netrapalli et~al.(2014)Netrapalli, Niranjan, Sanghavi, Anandkumar, and
  Jain]{netrapalli2014non}
P.~Netrapalli, U.~N. Niranjan, S.~Sanghavi, A.~Anandkumar, and P.~Jain.
\newblock Non-convex robust {PCA}.
\newblock In \emph{Advances in Neural Information Processing Systems}, pages
  1107--1115, 2014.

\bibitem[Nordhausen and Tyler(2015)]{nordhausen2015cautionary}
K.~Nordhausen and D.~E. Tyler.
\newblock A cautionary note on robust covariance plug-in methods.
\newblock \emph{Biometrika}, 102\penalty0 (3):\penalty0 573--588, 2015.

\bibitem[Novembre et~al.(2008)Novembre, Johnson, Bryc, Kutalik, Boyko, Auton,
  Indap, King, Bergmann, Nelson, Stephens, and Bustamante]{novembre2008europe}
J.~Novembre, T.~Johnson, K.~Bryc, Z.~Kutalik, A.~R. Boyko, A.~Auton, A.~Indap,
  K.~S. King, S.~Bergmann, M.~R. Nelson, M.~Stephens, and C.~D. Bustamante.
\newblock Genes mirror geography within europe.
\newblock \emph{Nature}, 456\penalty0 (7218):\penalty0 98--101, November 2008.
\newblock \doi{10.1038/nature07331}.

\bibitem[Nyquist(1988)]{Nyquist_l1_88}
H.~Nyquist.
\newblock Least orthogonal absolute deviations.
\newblock \emph{Computational Statistics \& Data Analysis}, 6\penalty0
  (4):\penalty0 361 -- 367, 1988.

\bibitem[Oja et~al.(2006)Oja, Sirki{\"a}, and Eriksson]{oja2016scatter}
H.~Oja, S.~Sirki{\"a}, and J.~Eriksson.
\newblock Scatter matrices and independent component analysis.
\newblock \emph{Austrian Journal of Statistics}, 35\penalty0 (2\&3):\penalty0
  175--189, 2006.

\bibitem[Ollila and Tyler(2014)]{ollila2014regularized}
E.~Ollila and D.~E. Tyler.
\newblock Regularized {M}-estimators of scatter matrix.
\newblock \emph{IEEE Trans. Signal Process.}, 62\penalty0 (22):\penalty0
  6059--6070, 2014.
\newblock ISSN 1053-587X.
\newblock \doi{10.1109/TSP.2014.2360826}.
\newblock URL \url{http://dx.doi.org/10.1109/TSP.2014.2360826}.

\bibitem[Osborne and Watson(1985)]{osborne_watson85}
M.~R. Osborne and G.~A. Watson.
\newblock An analysis of the total approximation problem in separable norms,
  and an algorithm for the total $l_1 $ problem.
\newblock \emph{SIAM Journal on Scientific and Statistical Computing},
  6\penalty0 (2):\penalty0 410--424, 1985.

\bibitem[{\"O}zye{\c{s}}il and Singer(2015)]{ozyesil2015robust}
Onur {\"O}zye{\c{s}}il and Amit Singer.
\newblock Robust camera location estimation by convex programming.
\newblock In \emph{Proceedings of the IEEE Conference on Computer Vision and
  Pattern Recognition}, pages 2674--2683, 2015.

\bibitem[Pimentel-Alarc{\'o}n and Nowak(2017)]{pimentel2017random}
D.~Pimentel-Alarc{\'o}n and R.~Nowak.
\newblock Random consensus robust pca.
\newblock \emph{Electronic Journal of Statistics}, 11\penalty0 (2):\penalty0
  5232--5253, 2017.

\bibitem[Price et~al.(2006)Price, Patterson, Plenge, Weinblatt, Shadick, and
  Reich]{Price_etal_2006}
A.~L. Price, N.~J. Patterson, R.~M. Plenge, M.~E. Weinblatt, N.~A. Shadick, and
  D.~Reich.
\newblock Principal components analysis corrects for stratification in
  genome-wide association studies.
\newblock \emph{Nature Genetics}, 38\penalty0 (8):\penalty0 904--909, 2006.
\newblock URL \url{http://www.ncbi.nlm.nih.gov/pubmed/16862161}.

\bibitem[Qiu and Vaswani(2010)]{qiu2010real}
C.~Qiu and N.~Vaswani.
\newblock Real-time robust principal components' pursuit.
\newblock In \emph{Communication, Control, and Computing (Allerton), 2010 48th
  Annual Allerton Conference on}, pages 591--598. IEEE, 2010.

\bibitem[Qu et~al.(2016)Qu, Sun, and Wright]{QuSW16}
Q.~Qu, J.~Sun, and J.~Wright.
\newblock Finding a sparse vector in a subspace: Linear sparsity using
  alternating directions.
\newblock \emph{{IEEE} Trans. Information Theory}, 62\penalty0 (10):\penalty0
  5855--5880, 2016.
\newblock URL \url{https://doi.org/10.1109/TIT.2016.2601599}.

\bibitem[Rahmani and Atia(2017{\natexlab{a}})]{rahmani17+_icml}
M.~Rahmani and G.~Atia.
\newblock Innovation pursuit: A new approach to the subspace clustering
  problem.
\newblock In Doina Precup and Yee~Whye Teh, editors, \emph{Proceedings of the
  34th International Conference on Machine Learning}, volume~70 of
  \emph{Proceedings of Machine Learning Research}, pages 2874--2882,
  International Convention Centre, Sydney, Australia, 06--11 Aug
  2017{\natexlab{a}}. PMLR.
\newblock URL \url{http://proceedings.mlr.press/v70/rahmani17b.html}.

\bibitem[Rahmani and Atia(2017{\natexlab{b}})]{rahmani2017coherence}
M.~Rahmani and G.~K. Atia.
\newblock Coherence pursuit: Fast, simple, and robust principal component
  analysis.
\newblock \emph{IEEE Transactions on Signal Processing}, 65\penalty0
  (23):\penalty0 6260--6275, 2017{\natexlab{b}}.

\bibitem[Sengupta et~al.(2017)Sengupta, Amir, Galun, Goldstein, Jacobs, Singer,
  and Basri]{sengupta2017new}
S.~Sengupta, T.~Amir, M.~Galun, T.~Goldstein, D.~W. Jacobs, A.~Singer, and
  R.~Basri.
\newblock A new rank constraint on multi-view fundamental matrices, and its
  application to camera location recovery.
\newblock \emph{{IEEE} Conference on Computer Vision and Pattern Recognition,
  {CVPR} 2017, Honolulu, Hawaii, USA, June 22-25, 2017}, pages 4798--4806,
  2017.

\bibitem[{Shi} and {Lerman}(2018)]{Shi_Lerman2018_cvpr}
Y.~{Shi} and G.~{Lerman}.
\newblock Estimation of camera locations in highly corrupted scenarios: All
  about that base, no shape trouble.
\newblock 2018.
\newblock To appear in Proceedings of CVPR 18.

\bibitem[Soltanolkotabi and Cand{\`e}s(2012{\natexlab{a}})]{soltanolkotabi2011}
M.~Soltanolkotabi and E.~J. Cand{\`e}s.
\newblock {A geometric analysis of subspace clustering with outliers.}
\newblock \emph{Ann. Stat.}, 40\penalty0 (4):\penalty0 2195--2238,
  2012{\natexlab{a}}.
\newblock \doi{10.1214/12-AOS1034}.

\bibitem[Soltanolkotabi and Cand{\`e}s(2012{\natexlab{b}})]{soltanolkotabi2012}
M.~Soltanolkotabi and E.~J. Cand{\`e}s.
\newblock {A geometric analysis of subspace clustering with outliers.}
\newblock \emph{Ann. Stat.}, 40\penalty0 (4):\penalty0 2195--2238,
  2012{\natexlab{b}}.
\newblock \doi{10.1214/12-AOS1034}.

\bibitem[Song et~al.(2017)Song, Woodruff, and Zhong]{song2017low}
Z.~Song, D.~P. Woodruff, and P.~Zhong.
\newblock Low rank approximation with entrywise l 1-norm error.
\newblock In \emph{Proceedings of the 49th Annual ACM SIGACT Symposium on
  Theory of Computing}, pages 688--701. ACM, 2017.

\bibitem[Sp{\"a}th and Watson(1987)]{spath_watson1987}
H.~Sp{\"a}th and G.~A. Watson.
\newblock On orthogonal linear approximation.
\newblock \emph{Numer. Math.}, 51:\penalty0 531--543, October 1987.

\bibitem[Srebro and Jaakkola(2003)]{srebro2003weighted}
N.~Srebro and T.~Jaakkola.
\newblock Weighted low-rank approximations.
\newblock In \emph{Proceedings of the 20th International Conference on Machine
  Learning (ICML-03)}, pages 720--727, 2003.

\bibitem[Stahel(1981)]{stahel1981breakdown}
W.~A. Stahel.
\newblock \emph{Breakdown of covariance estimators}.
\newblock Fachgruppe f{\"u}r Statistik, Eidgen{\"o}ssische Techn. Hochsch.,
  1981.

\bibitem[Steinhardt et~al.(2018)Steinhardt, Charikar, and
  Valiant]{steinhardt2018resilience}
J.~Steinhardt, M.~Charikar, and G.~Valiant.
\newblock Resilience: A criterion for learning in the presence of arbitrary
  outliers.
\newblock In \emph{ITCS}, 2018.

\bibitem[Sun et~al.(2014)Sun, Babu, and Palomar]{sun2014regularized}
Y.~Sun, P.~Babu, and D.~P. Palomar.
\newblock Regularized {T}yler's scatter estimator: existence, uniqueness, and
  algorithms.
\newblock \emph{IEEE Trans. Signal Process.}, 62\penalty0 (19):\penalty0
  5143--5156, 2014.
\newblock ISSN 1053-587X.
\newblock \doi{10.1109/TSP.2014.2348944}.

\bibitem[Torr and Murray(1993)]{torr1993outlier}
P.~H.~S. Torr and D.~W. Murray.
\newblock Outlier detection and motion segmentation.
\newblock In \emph{Sensor Fusion VI}, volume 2059, pages 432--444.
  International Society for Optics and Photonics, 1993.

\bibitem[Torr and Murray(1997)]{torr1997development}
P.~H.~S. Torr and D.~W. Murray.
\newblock The development and comparison of robust methods for estimating the
  fundamental matrix.
\newblock \emph{International journal of computer vision}, 24\penalty0
  (3):\penalty0 271--300, 1997.

\bibitem[Tsakiris and Vidal(2015)]{tsakiris2015dual}
M.~C. Tsakiris and R.~Vidal.
\newblock Dual principal component pursuit.
\newblock In \emph{{ICCV} Workshops 2015}, pages 850--858, 2015.

\bibitem[Tyler(1987)]{tyler_dist_free87}
D.~E. Tyler.
\newblock A distribution-free {$M$}-estimator of multivariate scatter.
\newblock \emph{Ann. Statist.}, 15\penalty0 (1):\penalty0 234--251, 1987.
\newblock ISSN 0090-5364.
\newblock \doi{10.1214/aos/1176350263}.

\bibitem[Vaswani and Narayanamurthy(2018)]{vaswani2018static}
N.~Vaswani and P.~Narayanamurthy.
\newblock Static and dynamic robust {PCA} via low-rank+sparse matrix
  decomposition: A review.
\newblock \emph{arXiv preprint arXiv:1803.00651}, 2018.

\bibitem[Visuri et~al.(2000)Visuri, Koivunen, and Oja]{visuri2000sign}
S.~Visuri, V.~Koivunen, and H.~Oja.
\newblock Sign and rank covariance matrices.
\newblock \emph{Journal of Statistical Planning and Inference}, 91\penalty0
  (2):\penalty0 557--575, 2000.

\bibitem[Wang and Singer(2013)]{wang2013exact}
L.~Wang and A.~Singer.
\newblock Exact and stable recovery of rotations for robust synchronization.
\newblock \emph{Information and Inference}, 2013.

\bibitem[Wang et~al.(2013)Wang, Szlam, and Lerman]{WSL13}
Y.~Wang, A.~Szlam, and G.~Lerman.
\newblock Robust locally linear analysis with applications to image denoising
  and blind inpainting.
\newblock \emph{SIAM J. Imaging Sciences}, 6\penalty0 (1):\penalty0 526--562,
  2013.

\bibitem[Watson(2001)]{watson2001some}
G.~A. Watson.
\newblock \emph{Some Problems in Orthogonal Distance and Non-Orthogonal
  Distance Regression}.
\newblock Defense Technical Information Center, 2001.
\newblock URL \url{http://books.google.com/books?id=WKKWGwAACAAJ}.

\bibitem[Wiesel(2012)]{wiesel_geodesic_convexity12}
A.~Wiesel.
\newblock Geodesic convexity and covariance estimation.
\newblock \emph{{IEEE} Trans. Signal Processing}, 60\penalty0 (12):\penalty0
  6182--6189, 2012.
\newblock \doi{10.1109/TSP.2012.2218241}.

\bibitem[Xu et~al.(2012)Xu, Caramanis, and Sanghavi]{xu2012robust}
H.~Xu, C.~Caramanis, and S.~Sanghavi.
\newblock Robust {PCA} via outlier pursuit.
\newblock \emph{{IEEE} Trans. Information Theory}, 58\penalty0 (5):\penalty0
  3047--3064, 2012.
\newblock \doi{10.1109/TIT.2011.2173156}.

\bibitem[Xu et~al.(2013)Xu, Caramanis, and Mannor]{xu2013outlier}
H.~Xu, C.~Caramanis, and S.~Mannor.
\newblock Outlier-robust {PCA}: the high-dimensional case.
\newblock \emph{IEEE transactions on information theory}, 59\penalty0
  (1):\penalty0 546--572, 2013.

\bibitem[Xu and Yuille(1995)]{Xu1995}
L.~Xu and A.~L. Yuille.
\newblock Robust principal component analysis by self-organizing rules based on
  statistical physics approach.
\newblock \emph{Neural Networks, IEEE Transactions on}, 6\penalty0
  (1):\penalty0 131--143, 1995.
\newblock ISSN 1045-9227.
\newblock \doi{10.1109/72.363442}.

\bibitem[Yang et~al.(2006)Yang, Rao, and Ma]{RobustGPCA}
A.~Y. Yang, S.~R. Rao, and Y.~Ma.
\newblock Robust statistical estimation and segmentation of multiple subspaces.
\newblock In \emph{Computer Vision and Pattern Recognition Workshop, 2006.
  CVPRW '06. Conference on}, page~99, june 2006.
\newblock \doi{10.1109/CVPRW.2006.178}.

\bibitem[Yi et~al.(2016)Yi, Park, Chen, and Caramanis]{yi2016fast}
X.~Yi, D.~Park, Y.~Chen, and C.~Caramanis.
\newblock Fast algorithms for robust pca via gradient descent.
\newblock In \emph{Advances in neural information processing systems}, pages
  4152--4160, 2016.

\bibitem[You et~al.(2017)You, Robinson, and Vidal]{YouRV17}
C.~You, D.~P. Robinson, and R.~Vidal.
\newblock Provable self-representation based outlier detection in a union of
  subspaces.
\newblock In \emph{2017 {IEEE} Conference on Computer Vision and Pattern
  Recognition, {CVPR} 2017, Honolulu, HI, USA, July 21-26, 2017}, pages
  4323--4332, 2017.
\newblock \doi{10.1109/CVPR.2017.460}.

\bibitem[Yu et~al.(2012)Yu, Zhang, and Ding]{yu2012efficient}
L.~Yu, M.~Zhang, and C.~Ding.
\newblock An efficient algorithm for {$L1$}-norm principal component analysis.
\newblock In \emph{Acoustics, Speech and Signal Processing (ICASSP), 2012 IEEE
  International Conference on}, pages 1377--1380. IEEE, 2012.

\bibitem[Zhang(2016)]{zhang2016robust}
T.~Zhang.
\newblock Robust subspace recovery by {T}yler's {M}-estimator.
\newblock \emph{Information and Inference}, 5\penalty0 (1):\penalty0 1--21,
  2016.

\bibitem[Zhang and Lerman(2014)]{zhang2014novel}
T.~Zhang and G.~Lerman.
\newblock A novel {M}-estimator for robust {PCA}.
\newblock \emph{Journal of Machine Learning Research}, 15\penalty0
  (1):\penalty0 749--808, 2014.

\bibitem[Zhang et~al.(2009{\natexlab{a}})Zhang, Szlam, and
  Lerman]{MKF_workshop09}
T.~Zhang, A.~Szlam, and G.~Lerman.
\newblock Median {$K$}-flats for hybrid linear modeling with many outliers.
\newblock In \emph{International Conference on Computer Vision Workshops
  ({ICCV} Workshops)}, pages 234--241, Kyoto, Japan, 2009{\natexlab{a}}.

\bibitem[Zhang et~al.(2009{\natexlab{b}})Zhang, Szlam, and
  Lerman]{zhang2009median}
T.~Zhang, A.~Szlam, and G.~Lerman.
\newblock Median {K}-flats for hybrid linear modeling with many outliers.
\newblock In \emph{Computer Vision Workshops (ICCV Workshops), 2009 IEEE 12th
  International Conference on}, pages 234--241. IEEE, 2009{\natexlab{b}}.

\bibitem[Zhang et~al.(2012)Zhang, Szlam, Wang, and Lerman]{LBF_journal12}
T.~Zhang, A.~Szlam, Y.~Wang, and G.~Lerman.
\newblock Hybrid linear modeling via local best-fit flats.
\newblock \emph{International Journal of Computer Vision}, 100:\penalty0
  217--240, 2012.
\newblock ISSN 0920-5691.
\newblock \doi{10.1007/s11263-012-0535-6}.

\bibitem[Zhang et~al.(2013)Zhang, Wiesel, and Greco]{teng_ami_maria_geo_convex}
T.~Zhang, A.~Wiesel, and M.~S. Greco.
\newblock Multivariate generalized {G}aussian distribution: Convexity and
  graphical models.
\newblock \emph{{IEEE} Trans. Signal Processing}, 61\penalty0 (16):\penalty0
  4141--4148, 2013.
\newblock \doi{10.1109/TSP.2013.2267740}.

\bibitem[Zhou et~al.(2010)Zhou, Li, Wright, Cand\`es, and Ma]{zhou2010stable}
Z.~Zhou, X.~Li, J.~Wright, E.~Cand\`es, and Y.~Ma.
\newblock Stable principal component pursuit.
\newblock In \emph{International Symposium on Information Theory Proceedings
  (ISIT)}, pages 1518--1522. IEEE, 2010.

\end{thebibliography}

\appendix

\subsection{Intuition for the Robust Covariance Matrices}
\label{app:robcov}

To clarify the robust energies in \eqref{eq:maronnaM} and \eqref{eq:tme}, we express them as scaled versions of negative log-likelihood functions with respect to heavy-tailed elliptical distributions. We thus understand the Maronna and Tyler M-estimators as maximum likelihood estimators that are robust to heavy tails. We assume a centered elliptical distribution with density function $f$ that is everywhere positive.
That is, $f$ has the form
\begin{equation}
    f(\bx; \bSigma) = \frac{g(\bx^T \bSigma^{-1} \bx)}{\sqrt{\det(\bSigma)}}, \ \text{ where } \ g: (0, \infty) \to (0, \infty).
\end{equation}
If $\bx_1$, $\ldots$, $\bx_N$ are i.i.d.~sampled from $f$, then the likelihood function has the form
\begin{equation}
L(\bSigma | \cX) = \frac{\prod_{i=1}^N g(\bx_i^T \bSigma^{-1} \bx_i)}{{\det(\bSigma)}^{\frac{N}{2}}}.
\end{equation}
Setting $\rho(t)= -2 \log(g(t))$, the negative log-likelihood function can be expressed as follows
\begin{equation}
\label{eq:maronna_explan}
-\frac{\log(L(\bSigma | \cX))}{N} = \frac{1}{2N} \sum_{i=1}^N \rho(\bx_i^T \bSigma^{-1} \bx_i) + \frac{1}{2} \log \det(\bSigma).
\end{equation}
This is the energy in \eqref{eq:maronnaM} and its minimization is equivalent to maximization of the likelihood function.
Using basic calculus, we calculate the derivative of this function as
\begin{align}\label{eq:maronna_der}
    - &\frac{\partial}{\partial\bSigma} \frac{\log(L(\bSigma | \cX))}{N} = \\ \nonumber &\ \ \ \ \ \ \ \ \frac{1}{2N} \sum_{i=1}^N \rho'(\bx_i^T \bSigma^{-1} \bx_i) \bSigma^{-1} \bx_i \bx_i^T \bSigma^{-1}   +\frac{1}{2} \bSigma^{-1}.
\end{align}
Setting~\eqref{eq:maronna_der} equal to zero, the minimizer of \eqref{eq:maronna_explan}, or equivalently \eqref{eq:maronnaM}, can be obtained by solving the following equation for $\bSigma$, where $w(t)=\rho'(t)$:
\begin{equation}
\label{eq:robust_explan}
\frac{1}{N} \sum_{i=1}^N w(\bx_i^T \bSigma^{-1} \bx_i) \bx_i \bx_i^T = \bSigma.
\end{equation}

We first note that when $f$ is a multivariate Gaussian distribution then $g(t) = \exp(-t/2)/c(D) = \exp(-t/2)/(2 \pi)^{D/2}$. 
This implies that $w(t)=1$ and the corresponding minimizer of \eqref{eq:maronnaM}, whose formula is expressed in \eqref{eq:robust_explan}, is the sample covariance matrix. On the other hand, when $g$ has heavier tails, e.g. $g(t) = \exp(-t^p/2p)/c(D,p)$ for $0 < p < 1$, \eqref{eq:robust_explan} results in more robust estimators to heavy tails. Indeed, in this case $w(t) = t^{p-1}$ and the solution of equation \eqref{eq:robust_explan} can be interpreted as a more robust version of the covariance matrix.
In the left hand side of \eqref{eq:robust_explan}, each term $\bx_i \bx_i^T $ is weighted by $(\bx_i^T \bSigma^{-1} \bx_i)^{p-1}$.
We further note that since we want to emphasize the top $d$ eigenvectors of $\bSigma$, we may identify $\bx_i$ as an ``outlier'' whenever $\bx_i^T \bSigma^{-1} \bx_i$ is relatively large, or equivalently, when $(\bx_i^T \bSigma^{-1} \bx_i)^{p-1}$ is relatively small. Therefore, the left hand side of \eqref{eq:robust_explan} is a weighted covariance matrix, which tends to de-emphasize outliers.

Another heavy-tailed density function can be obtained by considering the $D$-variate Student's $t$-distribution with $\nu$ degrees of freedom. In this case, $g(t) = c/(t+\nu)^{(D+\nu)/2}$ for some constant $c$ and thus $w(t)=(D+\nu)/(t+\nu)$. The tails of this distribution are heaviest when $\nu$ approaches zero. Formally, in this case, $\rho(t) = D \log(t) -2 \log(c) $, $w(t) = D/t$ and the energy in \eqref{eq:tme} corresponds to the expression in \eqref{eq:maronna_explan} divided by $D$ with the non-constant part of $\rho$, that is, $\rho(t) = D \log(t)$. This energy in \eqref{eq:tme} is different than the one in \eqref{eq:maronnaM} since its solution is not unique over the set of positive definite matrices and an additional requirement, such as $\tr(\bSigma)=1$, is needed.
On the other hand, the Maronna M-estimator assumes some conditions on $\rho$ that guarantee a unique minimizer of~\eqref{eq:maronnaM}  over the set of positive definite matrices.

A typical example of the Maronna M-estimator is the one mentioned above, where $w(t) = \rho'(t) = t^{p-1}$ for $0<p<1$. Notice that the non-constant part of $\rho$ is $\rho(t) =t^p/p $.
For this and other $\rho$'s satisfying the required conditions, the Maronna M-estimator can be computed by the following iterative procedure arbitrarily initialized with any positive definite matrix $\bSigma_0$:
\begin{equation}
\label{eq:iterate_maronna}
\bSigma_{k+1} = \frac{1}{N} \sum_{i=1}^N w(\bx_i^T \bSigma^{-1}_k \bx_i) \bx_i \bx_i^T.
\end{equation}
Numerical properties of this solution and, in particular, its convergence to the fixed point in \eqref{eq:robust_explan} are discussed in \cite{Arslan2004} and \cite{maronna1976robust}.
TME can similarly be computed by substituting $w(t)=D/t$ in \eqref{eq:iterate_maronna} and
dividing the resulting  $\bSigma_{k+1}$ by $\tr(\bSigma_{k+1})$, so that it satisfies the constraint $\tr(\bSigma_{k+1})=1$.
Numerical properties of this solution for TME are discussed
in~\cite{kent1988maximum} and \cite{zhang2016robust}.

As explained in \citep{maronna1976robust}, this framework can be formally extended to the more general setting where both $\bmu$ and $\bSigma$ are unknown, and one wishes to estimate them jointly. We remark that the estimate of $\bmu$ in this procedure would be a robust point estimator. Alternatively, one can follow the symmetrization procedure explained in \S\ref{subsec:robcov} and independently estimate $\bSigma$. The mean, $\bmu$, can then be estimated separately by some robust point estimator. As alluded to in \S\ref{subsec:robcov}, the advantage of the latter procedure over the former one is that errors in estimating $\bmu$ do not propogate errors in estimating $\bSigma$.

\begin{IEEEbiography}[{\includegraphics[width=1in,height=1.25in,trim=340 0 100 0,clip,keepaspectratio]{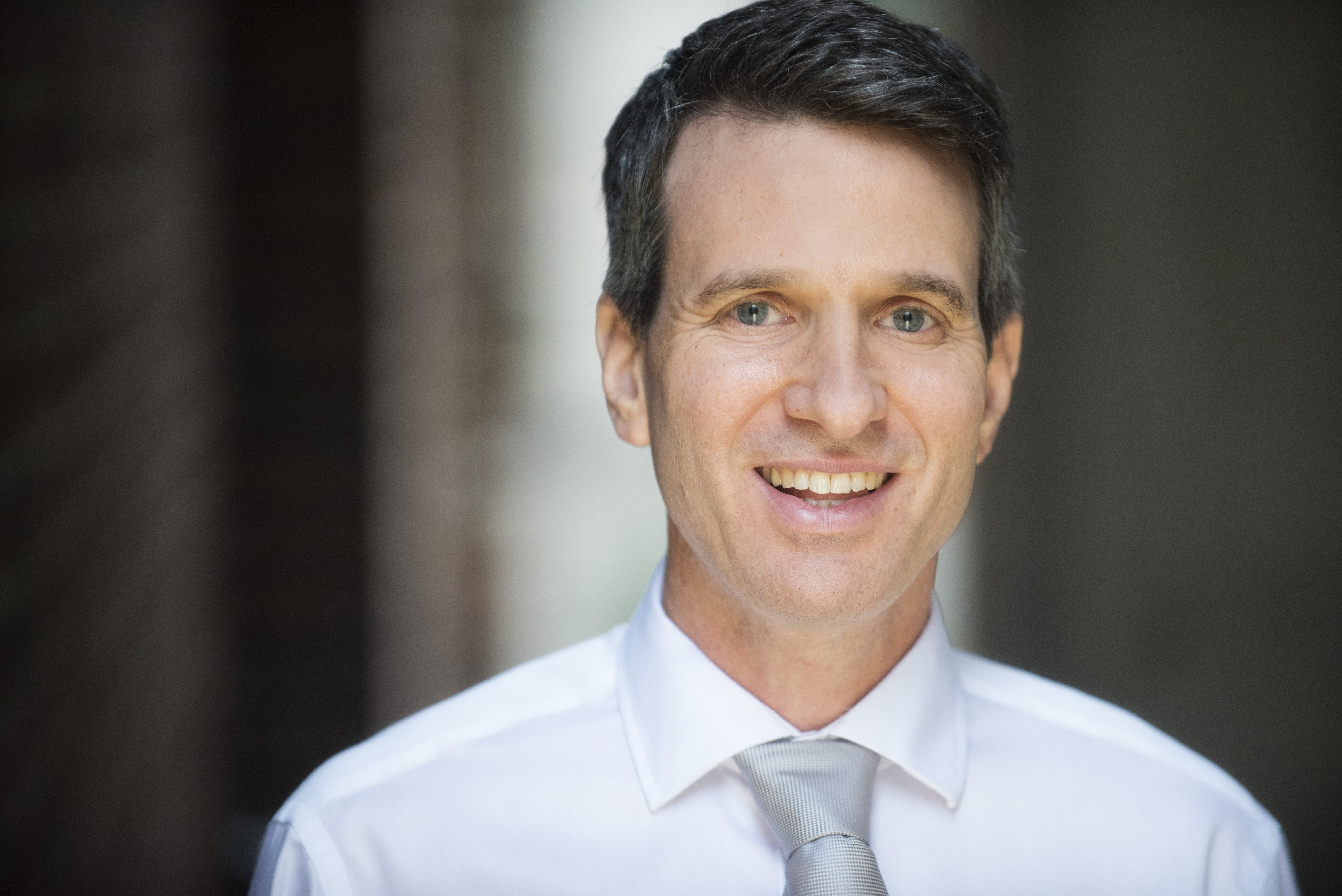}}]{Gilad Lerman}
	received the B.S. degree in Physics and Mathematics (summa cum laude) and M.S. degree in applied mathematics from Tel Aviv University in 1993 and 1995 respectively, and the Ph.D. degree in Mathematics from Yale University in 2000. After graduation, he was a Courant Instructor at the Courant Institute. Since 2004, he has been a faculty member in the School of Mathematics at the University of Minnesota, where he is currently a professor of Mathematics and director of the Minnesota Center for Industrial Mathematics and the IMA data science lab. His research interests include high-dimensional data analysis and modeling, non-convex optimization, robust statistics and machine learning. He is a recipient of an NSF CAREER award in 2010 and the Feinberg Foundation Visiting Faculty Fellowship at the Weizmann Institute in 2013.
\end{IEEEbiography}

\begin{IEEEbiography}[{\includegraphics[width=1in,height=1.25in,trim=0 100 0 100,clip,keepaspectratio]{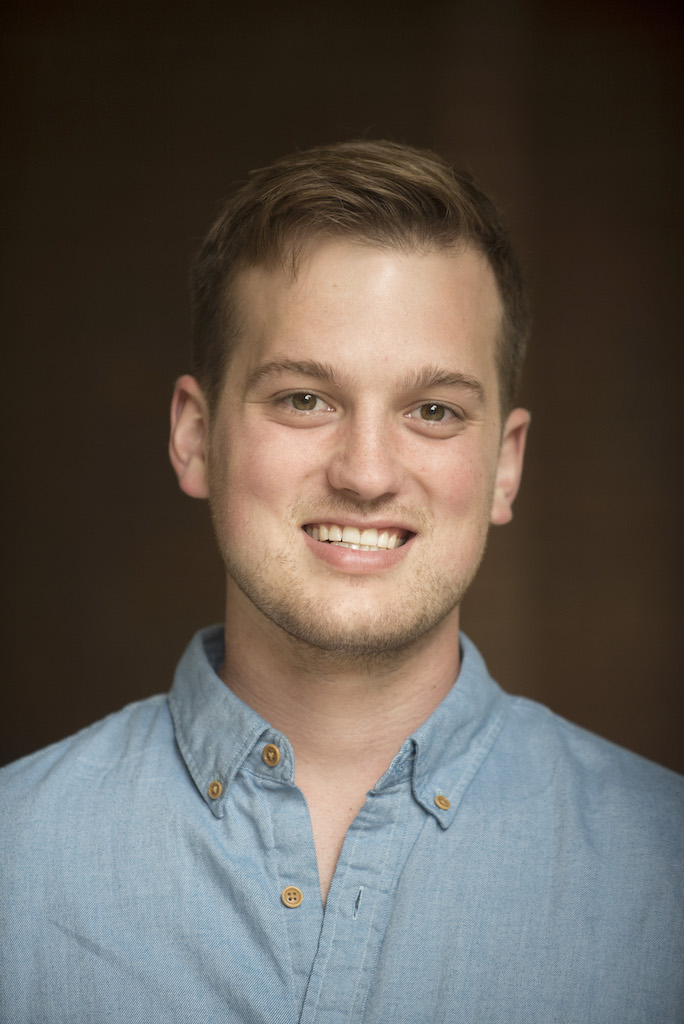}}]{Tyler Maunu}
	received the B.S., M.S., and Ph.D. degrees in Mathematics from University of Minnesota, Minneapolis, MN, USA, in 2013, 2016, and 2018 respectively. He also received the M.S. degree in Statistics at the University of Minnesota, Minneapolis, MN, USA in 2018. His research interests include machine learning, robust statistics, and non-convex optimization.
\end{IEEEbiography}

\newpage
\onecolumn

\end{document}